\documentclass[10pt,twocolumn,letterpaper]{article}

\usepackage{wacv}              
\usepackage{graphicx}
\usepackage{amsmath}
\usepackage{amssymb}
\usepackage{booktabs}
\usepackage{enumitem}
\usepackage{multirow}
\usepackage{pifont}
\usepackage{bm}
\usepackage{comment}
\usepackage{caption}
\usepackage{adjustbox}
\usepackage{float}

\newcommand{\xmark}{\ding{55}}%
\newcommand{\cmark}{\ding{51}}%
\newcommand\blfootnote[1]{
    \begingroup
    \renewcommand\thefootnote{}\footnote{#1}
    \addtocounter{footnote}{-1}
    \endgroup
}
\usepackage[pagebackref,breaklinks,colorlinks]{hyperref}

\usepackage[capitalize]{cleveref}
\crefname{section}{Sec.}{Secs.}
\Crefname{section}{Section}{Sections}
\Crefname{table}{Table}{Tables}
\crefname{table}{Tab.}{Tabs.}

\def\wacvPaperID{332} 
\def\confName{WACV}
\def\confYear{2025}
\newcommand{\name}{\texttt{STRIDE}}

\begin{document}

\title{\name: Single-video based Temporally Continuous \\ Occlusion-Robust 3D Pose Estimation} 

\author{Rohit Lal$^{\ddagger1*}$ \ Saketh Bachu$^{1*}$ \ Yash Garg$^{1}$ \ Arindam Dutta$^{1}$ \ Calvin-Khang Ta$^{1 }$ \ Hannah D. Cruz$^{1}$ \\ Dripta S. Raychaudhuri$^{\dagger1}$ \ M. Salman Asif$^{1}$ \ Amit K. Roy-Chowdhury$^{1 }$\\
$^{1}$University of California, Riverside\\
\tt\small \{rlal011,sbach008,ygarg002,adutt020,cta003,hdela004,drayc001,sasif,amitrc\}@ucr.edu}


\twocolumn[{
    \renewcommand\twocolumn[1][]{#1}
    \maketitle
    \centering
    \vspace{-2em}
    \begin{minipage}{0.95\textwidth}
        \centering
        \includegraphics[trim=000mm 000mm 000mm 000mm, clip=False, width=\linewidth]{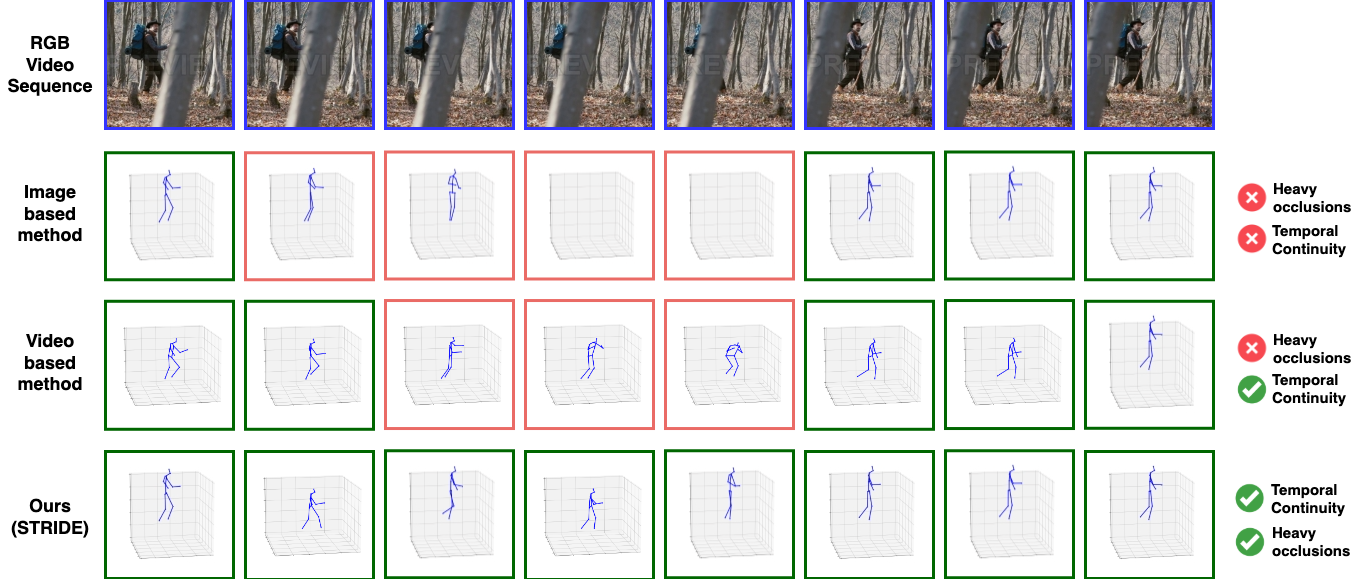}
    \end{minipage}
    \vspace{-0.5 em}
    \captionof{figure}{\textbf{Effect of occlusions on pose estimation.}
    Image-based 3D pose estimators \cite{black2023bedlam} often struggle with heavy occlusions, as illustrated in this figure. Without temporal context, predictions on highly obscured frames are inconsistent with prior poses, like the erroneous pose in the third column. Notably, even state-of-the-art video approaches \cite{shin2023wham} fail on prolonged full occlusions spanning multiple frames, as in columns 4 and 5. This highlights yet another critical limitation - models are brittle when deployed outside their training distributions. Without training examples of such long-duration occlusions, models fail to extrapolate reasonable poses. Our work addresses this through test-time training of a human motion prior. By fine-tuning on each new video, we tailor this parametric prior to handling sequence-specific occlusion patterns not observed during training. Given an initial noisy estimate, our approach refines the pose sequence into an accurate, temporally coherent output, as shown in the final row.}
    \label{fig:teaser}
    \vspace{2.2em}
}]

\blfootnote{* Equal contribution, $\ddagger$ Currently at NASA MSFC IMPACT. Work done while the author was at UCR., $\dagger$ Currently at AWS AI Labs. Work done while the author was at UCR.}

\begin{abstract}
Accurately estimating 3D human poses is crucial for fields like action recognition, gait recognition, and virtual/augmented reality. However, predicting human poses under severe occlusion remains a persistent and significant challenge. Existing image-based estimators struggle with heavy occlusions due to a lack of temporal context, resulting in inconsistent predictions, while video-based models, despite benefiting from temporal data, face limitations with prolonged occlusions over multiple frames. Additionally, existing algorithms often struggle to generalize unseen videos. Addressing these challenges, we propose \name~(\textbf{S}ingle-video based \textbf{T}empo\textbf{R}ally cont\textbf{I}nuous Occlusion-Robust 3\textbf{D} Pose \textbf{E}stimation), a novel Test-Time Training (TTT) approach to fit a human motion prior for estimating 3D human poses for each video. Our proposed approach handles occlusions not encountered during the model's training by refining a sequence of noisy initial pose estimates into accurate, temporally coherent poses at test time, effectively overcoming the limitations of existing methods. Our flexible, model-agnostic framework allows us to use any off-the-shelf 3D pose estimation method to improve robustness and temporal consistency. We validate \name's efficacy through comprehensive experiments on multiple challenging datasets where it not only outperforms existing single-image and video-based pose estimation models but also showcases superior handling of substantial occlusions, achieving fast, robust, accurate, and temporally consistent 3D pose estimates. Code is made publicly available at \url{https://github.com/take2rohit/stride}
\vspace{-7mm}
\end{abstract}


\section{Introduction}
\label{sec:intro}

Accurate 3D pose estimation \cite{10.1145/3603618} is an important problem in computer vision with a variety of real-world applications, including but not limited to action recognition \cite{kong2022human}, virtual and augmented reality \cite{9952586}, and gait recognition \cite{zhu2023sharc, dutta2024poise, dutta2024posture}. While the performance of 3D pose estimation algorithms has improved rapidly in recent years, the majority of these are image-based \cite{rhodin2018learning, nie2017monocular,raychaudhuri2023prior, ta2024multi}, estimating the pose from a single image. Consequently, these approaches still face inherent challenges in handling occluded subjects due to the limited visual information contained in individual images. To address these issues, recent efforts have explored video-based pose estimation algorithms \cite{yuan2022glamr, rempe2021humor}, leveraging temporal continuity across frames to resolve pose ambiguities from missing visual evidence.

Further, the success of both image and video-based state-of-the-art algorithms \cite{black2023bedlam, shin2023wham, yuan2022glamr,cycleadapt} relies heavily on supervised training on large datasets captured in controlled settings \cite{black2023bedlam}. This limits generalizability, as distribution shifts in uncontrolled environments can significantly degrade performance. For example, consider a scenario of an individual walking through a forest, periodically becoming fully obscured by trees, as depicted in Fig.~\ref{fig:teaser}. Image-based pose estimation methods \cite{black2023bedlam} struggle in such cases, as key spatial context is lost when the person is occluded. Without additional temporal cues, the model has insufficient visual evidence to accurately determine the 3D pose \cite{newell2016stacked,newell2017associative}. On the other hand, video-based approaches \cite{cycleadapt, yuan2022glamr, zhao2023poseformerv2} also suffer from performance degradation, despite modeling temporal information, due to such prolonged occlusions being absent in the training data \cite{cheng2019occlusion}. 

To deal with this large diversity in contexts, occlusion patterns, and imaging conditions in real-world videos, we explore the Test-Time Training (TTT) paradigm for 3D pose estimation. TTT allows for efficient on-the-fly adaptation to the specific occlusion patterns and data distribution shifts present in each test video. This facilitates better generalization, improving the model's capability to handle even prolonged occlusions. Furthermore, this reduces reliance on large annotated datasets, which are costly to collect, especially for occluded motions. 

Recent TTT approaches for 3D pose estimation \cite{cycleadapt, guan2021bilevel, guan2022out} fine-tune models using 2D cues like keypoints from test images. This approach has limitations as the 2D projection of 3D poses is ambiguous, as many 3D configurations can map to the same 2D pose. Also, 2D pose estimators can fail on unseen data distributions \cite{raychaudhuri2023prior,kim2022unified}, so fine-tuning on imperfect and ambiguous 2D poses can lead to incorrect model adaptation and degraded 3D pose predictions.

\begin{figure}[!t]
    \centering
    \includegraphics[width=\linewidth]{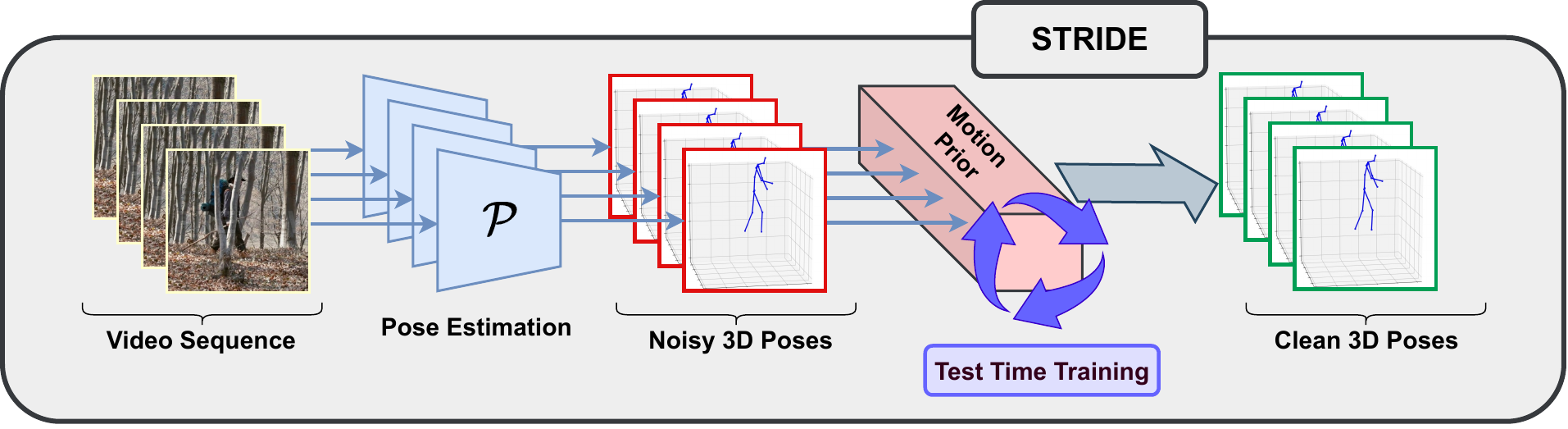}
    \caption{\textbf{Overview of our approach.} Our method enhances 3D pose estimation for occluded videos through test-time training of a motion prior model. We first extract initial 3D pose estimates from the test video using any 3D off-the-shelf pose estimator. To address occlusions and test distribution shifts, we then fine-tune the motion prior on that specific video by optimizing for smooth and continuous poses over the sequence.}
    \vspace{-5mm}
    \label{fig:TTT_teaser}
\end{figure}

To overcome the limitations of existing methods, we propose \name~(\textbf{S}ingle-video based \textbf{T}empo\textbf{R}ally cont\textbf{I}nuous Occlusion-Robust 3\textbf{D} Pose \textbf{E}stimation), a novel test-time training framework for 3D pose estimation under occlusion. The key component of our approach is a \emph{parametric motion prior} that is capable of modeling the dynamics of natural human motions and poses. This motion prior is pre-trained using a BERT-style \cite{devlin2019bert,zhu2023motionbert} approach on 3D pose sequences, learning to reconstruct temporally coherent poses when given a series of noisy estimates as input. At test time, given a sequence of noisy 3D poses from any existing pose estimation algorithm, \name~leverages this pre-trained prior to produce a clean sequence by fine-tuning it on each new video. We use 3D kinematic losses for motion smoothing via adapting the model to the video-specific motion patterns. By leveraging the motion prior's inherent knowledge of natural human movement during test-time training, \name~avoids ambiguities of 2D pose information faced by existing approaches. An overview of our approach is shown in Fig.~\ref{fig:TTT_teaser}.

A key advantage of our algorithm is that it can work alongside any off-the-shelf pose estimator to improve temporal consistency, providing model-agnostic pose enhancements. This allows \name~to not only surpass image-based pose estimators that lack contextual cues to resolve occlusions, but also outperform video-based methods. Notably, \name~can handle situations with up to 100\% occlusion of the human body over many consecutive frames. In comparison to existing test-time video based pose estimation method \cite{rempe2021humor,cycleadapt}, our approach is \emph{up to 2$\times$ faster} than previous state-of-the-art method \cite{cycleadapt} and operates without accessing any labeled training data during inference time, making it privacy \cite{schwartz2011pii} and storage-friendly.
\vskip 2pt
\noindent \textbf{Contributions.} In summary, we make the following key contributions:
\begin{enumerate}[topsep=2pt]
    \item We propose a novel test-time training algorithm (\name), for achieving temporally continuous 3D pose estimation under occlusions. 
    \item \name~is designed as a model-agnostic framework that leverages a human motion prior model to refine noisy 3D pose sequences from any off-the-shelf estimator into smooth and continuous predictions, highlighting efficiency and generalizability.
    \item \name~achieves state-of-the-art results on multiple challenging benchmarks including OCMotion \cite{huang2022object}, Occluded Human3.6M and Human3.6M \cite{h36m_pami}, thus demonstrating enhanced occlusion-robustness and temporal consistency. Additionally, \name~is computationally amicable, achieving a minimum of 2$\times$ speed-up over existing analogous algorithms.
\end{enumerate}

\section{Related Works}
\label{sec:related}

\noindent \textbf{Monocular 3D pose estimation.} Monocular 3D pose estimation is a fundamental and challenging problem in computer vision which involves the localisation of 3D spatial pose coordinates from just a single image. Recent deep learning-based methods for the problem have shown impressive performance on challenging academic datasets \cite{black2023bedlam, 10.1145/3603618}. \cite{mehta2017monocular} introduced the first CNN-based approach to regress 3D joints from a single image. Subsequent works \cite{rhodin2018learning, nie2017monocular} improved upon this by incorporating multi-view constraints and depth information. Recent methods \cite{xu2020deep, xu2021monocular} use kinematic and anatomical constraints along with data augmentation to achieve state-of-the-art results on academic datasets. However, these supervised methods often fail under distribution shifts. To address this, \cite{kundu2021non, kundu2022uncertainty} proposed self-supervised algorithms for 3D human pose estimation, which perform well on single images but struggle with occlusions and lack temporal continuity in video settings.

\noindent \textbf{Video-based 3D pose estimation.} Video-based 3D human pose estimation have shown impressive performance on challenging datasets. \cite{zhou2016sparseness} directly regresses 3D poses using consistency between 3D joints and 2D keypoints. \cite{pavllo20193d} utilized temporal convolutions for pose estimation in videos, while \cite{arnab2019exploiting} exploited SMPL pose and shape parameters for fine-tuning HMR in the wild. \cite{zhang2022mixste} proposed a mixed spatio-temporal approach alternating between spatial and temporal consistency. HuMoR \cite{rempe2021humor} maintained consistency across frames with weighted regularization using predicted contact probabilities. The state-of-the-art CycleAdapt \cite{cycleadapt} addresses domain shifts in 3D human mesh reconstruction by cyclically adapting HMRNet \cite{hmrnet} and MDNet \cite{cycleadapt} during test time. Despite the success of the above methods in maintaining temporal consistency, they are extremely slow due to an external optimization step and do not generalise well under distribution shifts. Severe occlusions often degrade the performance of these methods due to missing poses. Our work emphasizes these shortcomings and brings temporal continuity under severe occlusions by leveraging a motion-prior model that seamlessly handles missing poses.

\noindent \textbf{3D pose estimation under occlusion.} Handling occlusions poses a significant challenge in both image-based and video-based 3D pose estimation. Approaches like 3DNBF \cite{zhang2023nbf} leverage generative models to estimate poses but do not account for any temporal continuity. To alleviate the problem in a video-based setting, \cite{occaware} introduced data augmentation using occlusion labels with the Cylinder Man Model. Current methods address this by refining 3D poses for temporal consistency. Recent approaches such as GLAMR \cite{yuan2022glamr} recover human meshes globally from local motions and perform motion infilling based on visible motions. SmoothNet \cite{zeng2022smoothnet} uses a temporal refinement network to mitigate motion jitters from single image-based pose estimations. While effective for minor occlusions, these methods struggle with heavy occlusions. Also, these algorithms often fail to generalize under domain shifts. To improve on this, our approach adapts a motion prior from noisy 3D pose sequences to predict missing poses and maintain temporal consistency. 

\noindent \textbf{Test Time Optimization for 3D pose estimation.} A major shortcoming of fully supervised learning is that it can only handle test cases that are similar to the ones seen during the training process. For example, a novel type of occlusion or a human pose during testing can confuse the model and reduce its performance. \cite{iso} suggested verifying the estimated 2D poses and additionally ensuring the consistency of the lifted 3D poses by using randomly projected 2D poses to enhance the 3D human pose estimation. Further, \cite{PhysCap} introduced the idea of enforcing physical constraints on the estimated human poses to ensure they are physically plausible. Subsequently, \cite{dual_networks} proposed combining top-down and bottom-up human pose estimation approaches to take advantage of their strengths and also perform test time optimization using a re-projection loss, and bone length regularizations. Although these methods handle unseen test cases well, they are not designed for handling heavy occlusions and fail to predict poses under complete occlusions. Our work focuses on these gaps by refining and filling in missing 3D poses to maintain temporal consistency.

\section{Method} \label{sec:method}

\begin{figure*}[!h]
    \centering
    \includegraphics[width=0.81\linewidth]{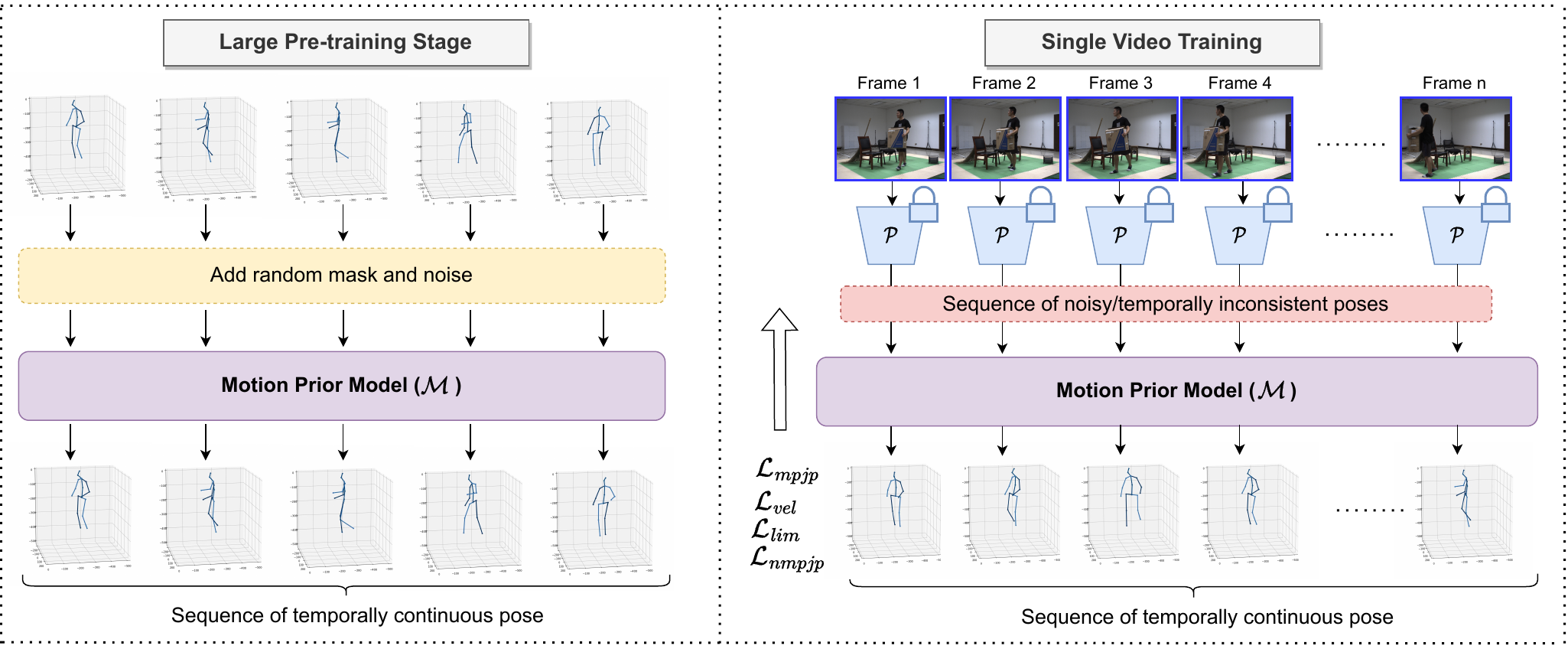}
    \caption{The presented figure illustrates the pipeline for our temporally continuous pose estimation, \name. Initially, we pre-train a motion prior model, denoted as $\mathcal{M}$, using a diverse set of 3D pose data sourced from various public datasets. The primary objective of this motion prior model is to generate a sequence of poses that exhibit temporal continuity when provided with a sequence of initially noisy poses. Moving into the single video training stage, we acquire a sequence of noisy poses using a 3D pose estimation model, $\mathcal{P}$. The weights of $\mathcal{P}$ are held constant during this phase. Subsequently, we pass this noisy pose sequence through the motion prior model $\mathcal{M}$ and retrain it using various supervised losses, as outlined in Equation~\ref{eq:final_loss}. The end result of this training process is a model capable of producing temporally continuous 3D poses for that specific video.}
    \label{fig:temp3d_arch}
    \vspace{-5mm}
\end{figure*}

We address the problem of extracting temporally continuous 3D pose estimates from a monocular video that may contain heavy occlusions. Given an off-the-shelf monocular 3D pose estimator $\mathcal{P}$ (either image or video-based) that produces temporally inconsistent poses due to occlusions or domain gaps, our goal is to output clean, temporally coherent 3D pose sequences that better match natural human motion dynamics. To achieve this, we propose a two-stage approach, as illustrated in Fig.~\ref{fig:temp3d_arch}.
\begin{enumerate}
    \item \textbf{Learning a motion prior:} We first pre-train a self-attention-based motion prior model $\mathcal{M}$ on labeled 3D pose datasets in a BERT-style  \cite{devlin2019bert,zhu2023motionbert}. During pre-training, we synthetically corrupt the 3D joint inputs with noise to simulate occlusions and other errors. $\mathcal{M}$ is then trained to denoise these inputs and reconstruct a sequence of temporally coherent 3D poses.  This allows $\mathcal{M}$ to learn strong general priors of natural human motion dynamics.
    
    \item \textbf{Test-time alignment:} For a given test video, we obtain noisy per-frame poses using $\mathcal{P}$ \cite{black2023bedlam} and adapt the motion prior model $\mathcal{M}$ in an unsupervised manner to align it to the specific motion exhibited in the video. This adaptation step allows us to obtain optimal pose estimates for the given video.   
\end{enumerate}

Section~\ref{sec:method_arch} outlines the architecture of the motion prior model $\mathcal{M}$. Section~\ref{sec:method_train} discusses the masked sequence modelling approach for pre-training $\mathcal{M}$ on synthetically corrupted pose sequences. Lastly, Section~\ref{sec:method_inference} introduces the self-supervised losses for fine-tuning $\mathcal{M}$ during test time on individual videos.

\subsection{Network Architecture} 
\label{sec:method_arch}

\noindent We base our motion prior model $\mathcal{M}$ on the DSTFormer architecture \cite{zhu2023motionbert}, originally proposed for lifting 2D poses to 3D. Here, we modify and adapt DSTFormer for the sequence-to-sequence task of denoising and smoothing noisy 3D pose sequence inputs. Specifically, the motion prior $\mathcal{M}$ takes in a sequence of 3D body poses represented as $\mathbf{X} \in \mathbb{R}^{T\times J \times 3}$, where $T$ is the number of frames, $J$ is the number of joints, and each pose consists of $J \times 3$ coordinate values. $\mathcal{M}$ denoises the input sequence to produce refined temporally coherent 3D poses $\bar{\mathbf{X}} \in \mathbb{R}^{T\times J \times 3}$. Additional details on the architecture and implementation method is provided in the Section 1 and Section 2 of supplementary material respectively.

\subsection{Learning a Motion Prior} 
\label{sec:method_train}

To build a strong prior for human motion dynamics, we draw inspiration from the success of large language models like BERT \cite{devlin2019bert} that leverage large-scale self-supervised pre-training. 
Here, we extend this paradigm to 3D human pose estimation. Specifically, given a dataset of 3D pose sequences, we synthetically mask these sequences to simulate occlusions and other errors. Similar to \cite{ci2022gfpose,zhu2023motionbert}, the prior $\mathcal{M}$ is trained to denoise these noisy inputs to reconstruct a sequence of temporally coherent 3D poses. We selected BERT-style training for \name's motion prior because it is highly effective at capturing bidirectional dependencies in human motion data and excels in scenarios where pose information may be incomplete or corrupted. BERT-style training leverages a bidirectional approach, allowing the model to consider both preceding and succeeding context simultaneously. This is particularly advantageous for understanding human motion, as it mirrors real-world situations where poses might be partially missing or noisy.

During pre-training, we apply both joint-level and frame-level masking to a 3D pose sequence $\mathbf{X}$ to obtain a corrupted sequence $\texttt{mask}(\mathbf{X})$ which mimics realistic scenarios of imperfect predictions and occlusions. The prior $\mathcal{M}$ is trained to reconstruct the complete 3D motion sequence $\mathbf{\bar{X}}$ from this corrupted input $\mathbf{{X}}$ by minimizing losses on 3D joint positions $\mathcal{L}_\text{3D}$ between the reconstruction and the ground-truth pose. Additionally, we incorporate a velocity loss following \cite{pavllo20193d, zhang2022mixste}.

\subsection{Test-Time Alignment} \label{sec:method_inference}

Given the pre-trained motion prior model $\mathcal{M}$ that takes in noisy 3D poses and outputs temporally coherent predictions, our goal is to leverage this for pose estimation on new test videos. We first obtain an initial noisy estimate of the 3D pose sequence using any off-the-shelf pose detector $\mathcal{P}$ \cite{black2023bedlam}. As these models struggle on occlusions and distribution shifts, their outputs lack temporal consistency. To address this, we pass the noisy poses through $\mathcal{M}$ to achieve a refined estimate.

Although the prior refines pose, some inconsistencies like domain shift and novel human motion may be present in the videos. Hence, we propose additional test-time training of $\mathcal{M}$ using geometric and physics-based constraints to adapt to such situations. Similar to internal learning approaches like Deep Video Prior \cite{lei2022deep}, our proposed self-supervision strategy fine-tunes the motion prior to the specifics of each test video for enhanced outputs. We use four loss regularizers targeting different aspects of human motion: (1) Limb Loss, (2) Mean Per Joint Position (MPJP) Loss, (3) Normalized MPJP (N-MPJP) Loss, and (4) Velocity Loss. Crucially, only $\mathcal{M}$ is updated during test-time training while $\mathcal{P}$ remains fixed to preserve the pose estimation capabilities of off-the-shelf models.

\noindent\textbf{Limb Loss:} Limb length consistency is an important aspect of anatomically plausible 3D human pose predictions. This loss encourages the model to produce temporally stable limb lengths, contributing to more realistic and physically plausible pose estimations. The idea is to penalize variability in limb lengths across frames. If the limb lengths exhibit large variations, it may indicate inconsistency or instability in the predicted poses. The limb loss function $\mathcal{L}_\text{lim}$ is defined as follows,
\begin{equation}
\label{eq:L_lim_var}
\mathcal{L}_\text{lim} = \frac{1}{J} \sum_{j=1}^{J-1} \underbrace{\frac{1}{T} \sum_{t=1}^{T} \left( \bm{\mathcal{J}}_{t,j} - \frac{1}{T} \sum_{t'=1}^{T} \bm{\mathcal{J}}_{t',j} \right)^2}_{\text{Variance of Joint Lengths Across Time}} \ .
\end{equation}
Here $\bm{\mathcal{J}} \in \mathbb{R}^{T \times (J-1)}$ represents a matrix of the normalised length of limb $j<(J-1)$ at any time $t < T$. By calculating the variance of limb lengths and taking the mean, the loss encourages the model to produce more consistent and stable limb lengths across the entire sequence. This can be beneficial in applications where it is crucial to maintain anatomical consistency in the predicted 3D poses.

To further regularize for the cases where the 3D pose estimation model $\mathcal{P}$ fails to detect any pose, we use linear interpolation between joints. Consider that the video consists of $N$ frames, out of which the model fails to predict anything for $q$ frames. 
The linear extrapolation and interpolation function $L: \mathbb{R}^{(N-q) \times J \times 3} \rightarrow \mathbb{R}^{N \times J \times 3}$ fills in the missing inputs. This provides pseudo-labels during training for two of our loss functions. These pseudo-labels also help to ensure temporal continuity in the predicted poses. 
\noindent \textbf{Mean Per Joint Position (MPJP) Loss:} This loss focuses on the accuracy of the pose estimation by penalizing deviations in the spatial position of individual joints. It computes the mean Euclidean distance between the predicted $\mathbf{\hat{X}}$ poses and pseudo-poses $\mathbf{\tilde{X}} = L(\mathbf{\hat{X}})$ where $\mathbf{\hat{X}}$ is the noisy sequence of poses obtained from $\mathcal{P}$. It measures the average distance between corresponding joints in the predicted and pseudo labels. It is defined as follows,
\begin{equation}
\label{eq:L_mpjp}
\mathcal{L}_\text{MPJP} = \frac{1}{T \cdot J \cdot 3} \sum_{t=1}^{T} \sum_{j=1}^{J} \sum_{d=1}^{3}\lVert \mathbf{\hat{X}}_{t,j,d} - \mathbf{\tilde{X}}_{t,j,d}\rVert_2
\end{equation}
\noindent \textbf{Normalized MPJP (N-MPJP) Loss:} This loss function introduces a normalization step to address scale variations between the predicted and target poses. It calculates the scale factor based on the norms of the predicted and target poses and then applies this scale factor to the predicted poses before computing the MPJPE. The normalization in $\mathcal{L}_\text{N-MPJP}$ aims to make the model more robust to variations in absolute pose values. It is particularly useful when the scale of the poses in the training and testing data may differ. By incorporating scale information, $\mathcal{L}_\text{N-MPJP}$ addresses scale-related issues during training, potentially improving the model's generalization to different scenarios.
\begin{align}
\label{eq:L_nmpjp}
\mathcal{L}_\text{NMPJP} = \mathcal{L}_\text{MPJP}(s\mathbf{\hat{X}},\mathbf{\tilde{X}})
\end{align}
\begin{equation}
\text{where } s = \frac{\sum_{t=1}^{T} \sum_{j=1}^{J} \sum_{d=1}^{3} \left\| \mathbf{\tilde{X}}_{t,j,d} \cdot \mathbf{\hat{X}}_{t,j,d} \right\|_2}{\sum_{t=1}^{T} \sum_{j=1}^{J} \sum_{d=1}^{3}\left\| \mathbf{\hat{X}}_{t,j,d} \right\|_2^2}
\end{equation}
In Equation \ref{eq:L_nmpjp}, $s$ represents the scale. The combination of both $\mathcal{L}_\text{NMPJP}$ and $\mathcal{L}_\text{MPJP}$ losses allows the model to simultaneously optimize for accurate joint positions ($\mathcal{L}_\text{MPJP}$) and address scale variations ($\mathcal{L}_\text{NMPJP}$). The incorporation of $\mathcal{L}_\text{NMPJP}$ allows the model to learn to handle scenarios where the pose scale may differ between training and testing data.
\noindent \textbf{Velocity Loss:} We optimize velocity loss similar to Equation~\ref{eqn:loss_3d_vel}, but instead of ground truth, we use pseudo-labels,
\begin{equation}
\mathcal{L}_\text{vel} = \frac{1}{N \cdot (J-1)} \sum_{t=1}^{T-1} \sum_{j=1}^{J} \sum_{d=1}^{3} \left\| \mathbf{\hat{V}} - \mathbf{\tilde{V}}\right\|_2
\label{eqn:loss_3d_vel}
\end{equation}
where $\mathbf{\hat{V}} = \mathbf{\hat{X}}_{t+1,j,d} - \mathbf{\hat{X}}_{t,j,d}$ and $\mathbf{\tilde{V}} = \mathbf{\tilde{X}}_{t+1,j,d} - \mathbf{\tilde{X}}_{t,j,d}$ represent velocities of predicted poses and pseudo label poses respectively. The velocity loss helps in smoothing the movement and removing unwanted jittering across frames.\\
\noindent \textbf{Overall Loss.} In summary, by combining all the above-mentioned losses into one final loss function as shown in Equation~\ref{eq:final_loss}, $\mathcal{M}$ is trained to produce accurate joint positions, maintain anatomical consistency, and handle scale variations,
\begin{equation}
\label{eq:final_loss}
\mathcal{L}_{total} = \lambda_1\mathcal{L}_\text{lim} + \lambda_2\mathcal{L}_\text{MPJP} + \lambda_3\mathcal{L}_\text{NMPJP} + \lambda_4\mathcal{L}_\text{vel}
\end{equation}
Here, $\lambda_i$, where $i \in {1,2,3,4}$, refers to loss-weighing hyper-parameters which remain constant for all evaluations.

\section{Experiments and Results}
\label{sec:exp}
In this section, our primary objective is to provide a comprehensive understanding of our approach. We elaborate on the datasets employed and conduct a thorough comparison with state-of-the-art methodologies. Furthermore, we analyze the qualitative results, pinpointing areas where existing methods may falter. As a conclusive step, we perform an ablation study to assess the impact of pre-training and different loss functions, shedding light on their contributions to our experimental framework.

We conduct evaluations on three datasets with varying levels of occlusion: Human3.6M, representing scenarios without occlusion; OCMotion, moderate occlusion; and Occluded Human3.6M, representing heavy occlusion. The metrics assessed include Procrustes-aligned mean per joint position error (PA-MPJPE), mean per joint position error (MPJPE), and acceleration error (Accel), measured as the disparity in acceleration between ground-truth and predicted 3D joints. We report the metrics in (mm). We use BEDLAM-CLIFF \cite{black2023bedlam} as the off-the-shelf pose estimation method. We compare the error rates of \name~ and the baseline methods in Tables~\ref{tab:occludedH36M}, \ref{tab:OCMotion} and \ref{tab:cleanH36M}. The best results are in \textbf{bold} and arrows indicate the percentage improvement over the best existing algorithm. Qualitative video results can be found on our shared  \href{https://github.com/take2rohit/stride}{GitHub repository}.


\subsection{Datasets}

\noindent\textbf{Human3.6M} \cite{h36m_pami}: This indoor-scene dataset is crucial for 3D human pose estimation from 2D images. We use every 1 in 5 frames in the test split and achieve comparable performance to state-of-the-art methods.

\noindent\textbf{OCMotion} \cite{huang2022object}: This video dataset extends the 3DOH50K image dataset with natural occlusions, comprising 300K images at 10 FPS. We use only the test split since our method does not require supervised training.

\noindent\textbf{Occluded Human3.6M}: We prepare a new dataset by modifying the  Human3.6M \cite{h36m_pami}. It is curated to evaluate pose estimation under significant occlusion. This dataset uses random erase occlusions covering up to 100\% of a person for 1.6 seconds within 3.2-second videos.

\noindent\textbf{BRIAR} \cite{cornett2023expanding}: Features videos of human subjects in extremely challenging conditions, recorded at varying distances and from UAVs. Additional details on datasets and implementation specifics, are provided in Section 7 of supplementary material.

\subsection{Quantitative Results}
\begin{table}[h]
    \centering
    \resizebox{0.95\linewidth}{!}{
    \setlength{\tabcolsep}{3.5pt}
    \def\arraystretch{1.35}  
    \begin{tabular}{ccccc}
    \toprule
    & Method & PA-MPJPE & MPJPE & Accel \\ \hline
    \multirow{2}{*}{\rotatebox{90}{Image}}& 
    CLIFF~\cite{li2022cliff} & 183.5 & 100.5 & 38.4 \\
    & BEDLAM~\cite{black2023bedlam} & 179.5 & 98.9 & 39.1 \\
    \midrule
    \multirow{4}{*}{\rotatebox{90}{Video}} 
    & GLAMR~\cite{yuan2022glamr} & 213.9 & 380.3 & 42.3 \\
    & PoseFormerV2~\cite{zhao2023poseformerv2} & 193.9 & 260.2 & 38.7 \\
    & CycleAdapt~\cite{cycleadapt} & 77.6 & 132.6 & 48.7 \\
    & MotionBERT~\cite{zhu2023motionbert} & 76.1 & 112.8 & 28.7 \\
    \midrule
    & {\name~(ours)} & \textbf{59.0} \scriptsize{{(57\%$\downarrow$)}} 
                      & \textbf{80.7} \scriptsize{{(18\%$\downarrow$)}} 
                      & \textbf{26.6} \scriptsize{{(7\%$\downarrow$)}} \\
    \bottomrule
    \end{tabular}}
    \caption{3D Pose estimation results on \textbf{Occluded Human3.6M}. This dataset is crucial as it is the only dataset that has significant occlusion. The results underscore that \name~ surpasses all state-of-the-art with substantial percentage improvements, affirming its robustness in handling occlusions.}
    \label{tab:occludedH36M}
\end{table}

Our method is most effective under heavy occlusions. We significantly outperform other state-of-the-art methods on the Occluded Human3.6M dataset as shown in Table~\ref{tab:occludedH36M}. Notably, \name~ performs significantly better than BEDLAM despite using pseudo-labels from BEDLAM. BEDLAM fails to produce poses under heavy occlusion; hence, the evaluation results drop significantly. However, since \name~ incorporates temporal information to address these gaps in the video, we predict reasonable poses even in case of heavy occlusions and improve the result of BEDLAM by a significant margin. It is important to note that by using \name~ we do not only outperform BEDLAM, but we also outperform all the other existing video- and image-based state-of-the-art methods. This is mainly because existing methods do not incorporate human motion prior and hence result in temporally implausible poses.


\begin{table}[h]
    \centering
    \resizebox{0.95\linewidth}{!}{     
    \setlength{\tabcolsep}{3.5pt}
    \def\arraystretch{1.35}  
    \begin{tabular}{ccccc}
    \toprule
    & Method & PA-MPJPE & Accel & Avg \\ \hline
    \multirow{3}{*}{\rotatebox{90}{Image}}& 
    OOH~\cite{zhang2020object} & 55.0 & 48.6 & 51.8\\
    & PARE~\cite{kocabas2021pare} & 52.0 & 43.6 & 47.8\\ 
    & BEDLAM~\cite{black2023bedlam} & 47.1 & 49.0 & 48.0\\
    \midrule   
    \multirow{5}{*}{\rotatebox{90}{Video}} 
    & PoseFormerV2~\cite{zhao2023poseformerv2} & 126.3 & \textbf{28.5} & 77.4\\
    & GLAMR~\cite{yuan2022glamr} & 89.9 & 51.3 & 70.6\\
    & CycleAdapt~\cite{cycleadapt} & 74.6 & 57.5 & 66.0\\
    & ROMP~\cite{sun2021monocular} & 48.1 & 57.2 & 52.6\\
    \midrule
    & {\name~(ours)} & \textbf{46.2} \scriptsize{{(2\%$\downarrow$)}} & {47.8} & \textbf{47.0} \scriptsize{{(2\%$\downarrow$)}}\\
    \bottomrule
    \end{tabular}}
    \caption{\textbf{3D pose estimation results on OCMotion \cite{huang2022object}}. \name~ outperforms other image and video-based pose estimation methods. While PoseFormerV2 has the lowest accel., it also exhibits the highest PA-MPJPE error. This is due to oversmoothing and inaccurate interpolation between poses which compromises the pose estimation accuracy.}
    \label{tab:OCMotion}
\end{table}

Since Occluded Human3.6M contains artificial occlusions, we also evaluated on the OCMotion dataset, which contains real-world, natural occlusions. Table~\ref{tab:OCMotion} shows that our approach \name~attains state-of-the-art results on the OCMotion dataset \cite{huang2022object}. Since we obtained good pseudo-labels from BEDLAM under partial occlusions, we observe the proximity of our results to BEDLAM. \emph{It is important to highlight that methods such as \cite{sun2021monocular,kocabas2021pare} are supervised and trained on the training split of OCMotion. In contrast to these algorithms, our proposed approach (\name) does not assume access to any labeled training dataset.}

\begin{table}[h]
    \centering
    \resizebox{0.95\linewidth}{!}{
    \setlength{\tabcolsep}{4pt}
    \def\arraystretch{1.35}  
    \begin{tabular}{ccccc}
    \toprule
    & Method & PA-MPJPE & MPJPE & Accel \\
    \midrule
    \multirow{3}{*}{\rotatebox{90}{Image}} 
    & CLIFF~\cite{li2022cliff} & 56.1 & 89.6 & -\\
    & BEDLAM-HMR~\cite{black2023bedlam} & 51.7 & 81.6 & - \\
    & BEDLAM-CLIFF~\cite{black2023bedlam} & 50.9 & 70.9 & 39.14 \\ \hline   
    \multirow{3}{*}{\rotatebox{90}{Video}}
    & GLAMR~\cite{yuan2022glamr} & - & - & - \\
    & CycleAdapt~\cite{cycleadapt} & 64.5 & 106.3 & 57.25 \\
    & MotionBERT$^\ast$~\cite{zhu2023motionbert} & 64.15 & 95.8 & \textbf{14.8} \\
    \midrule
    & {\name~(ours)} & \textbf{50.4} \scriptsize{{(1\%$\downarrow$)}} & \textbf{69.7} \scriptsize{{(2\%$\downarrow$)}} & 37.1 \\  
    \bottomrule
    \end{tabular}}
    \caption{3D pose estimation results on \textbf{Human3.6M}. Our evaluation demonstrates that our results are comparable to the BEDLAM-CLIFF baseline. This is due to the occlusion-free nature of the Human3.6M, which yields already refined and consistent poses with limited room for improvement.}
    \label{tab:cleanH36M}
\end{table}

Our method demonstrates minor improvement over BEDLAM-CLIFF~\cite{black2023bedlam} on the original Human3.6M dataset, as evidenced in Table~\ref{tab:cleanH36M}. The marginal enhancement is primarily due to the nature of the Human3.6M dataset, which lacks occlusions, thereby limiting the potential for improvement beyond the baseline. A thorough analysis of our findings, including the observed enhancement in temporal smoothness, is provided in the Section 3 of supplementary.

\begin{figure*}[t]
    \centering
    \begin{subfigure}{0.11\linewidth}
    \centering
        \includegraphics[width=\linewidth]{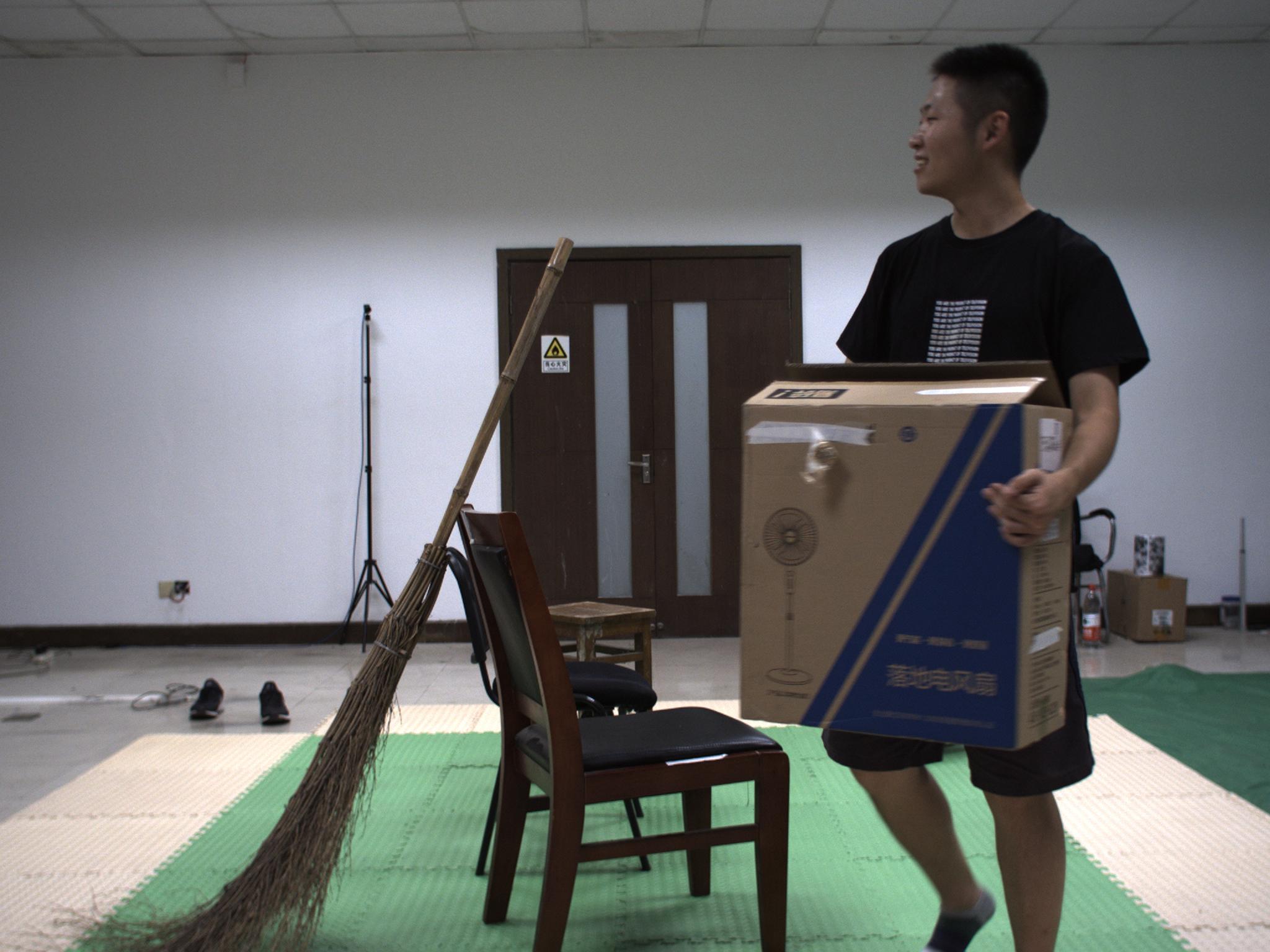}
    \end{subfigure}\hfill%
    \begin{subfigure}{0.11\linewidth}
    \centering
        \includegraphics[width=\linewidth]{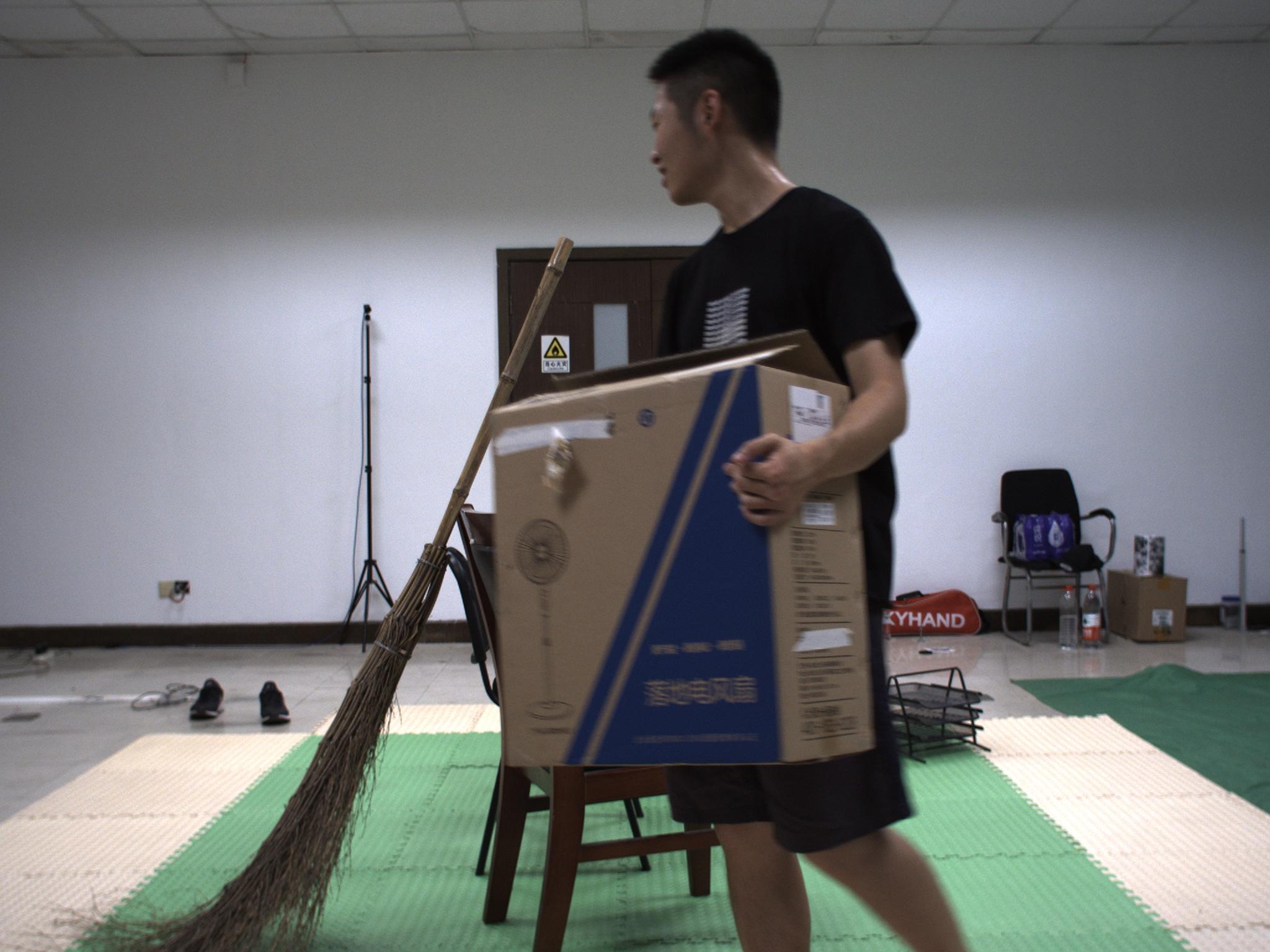}
    \end{subfigure}\hfill%
    \begin{subfigure}{0.11\linewidth}
    \centering
        \includegraphics[width=\linewidth]{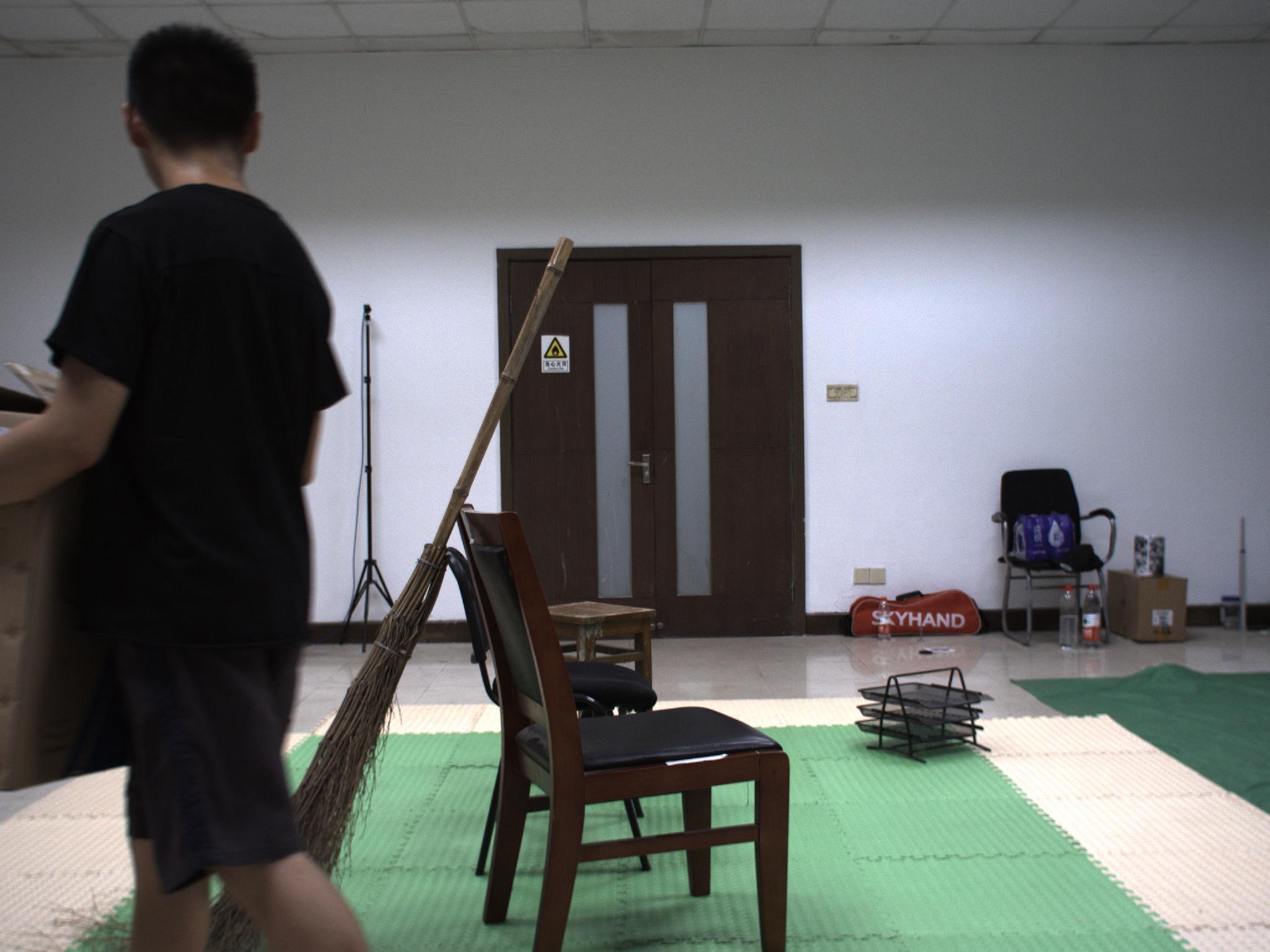}
    \end{subfigure}\hfill%
    \begin{subfigure}{0.11\linewidth}
     \centering
       \includegraphics[width=\linewidth]{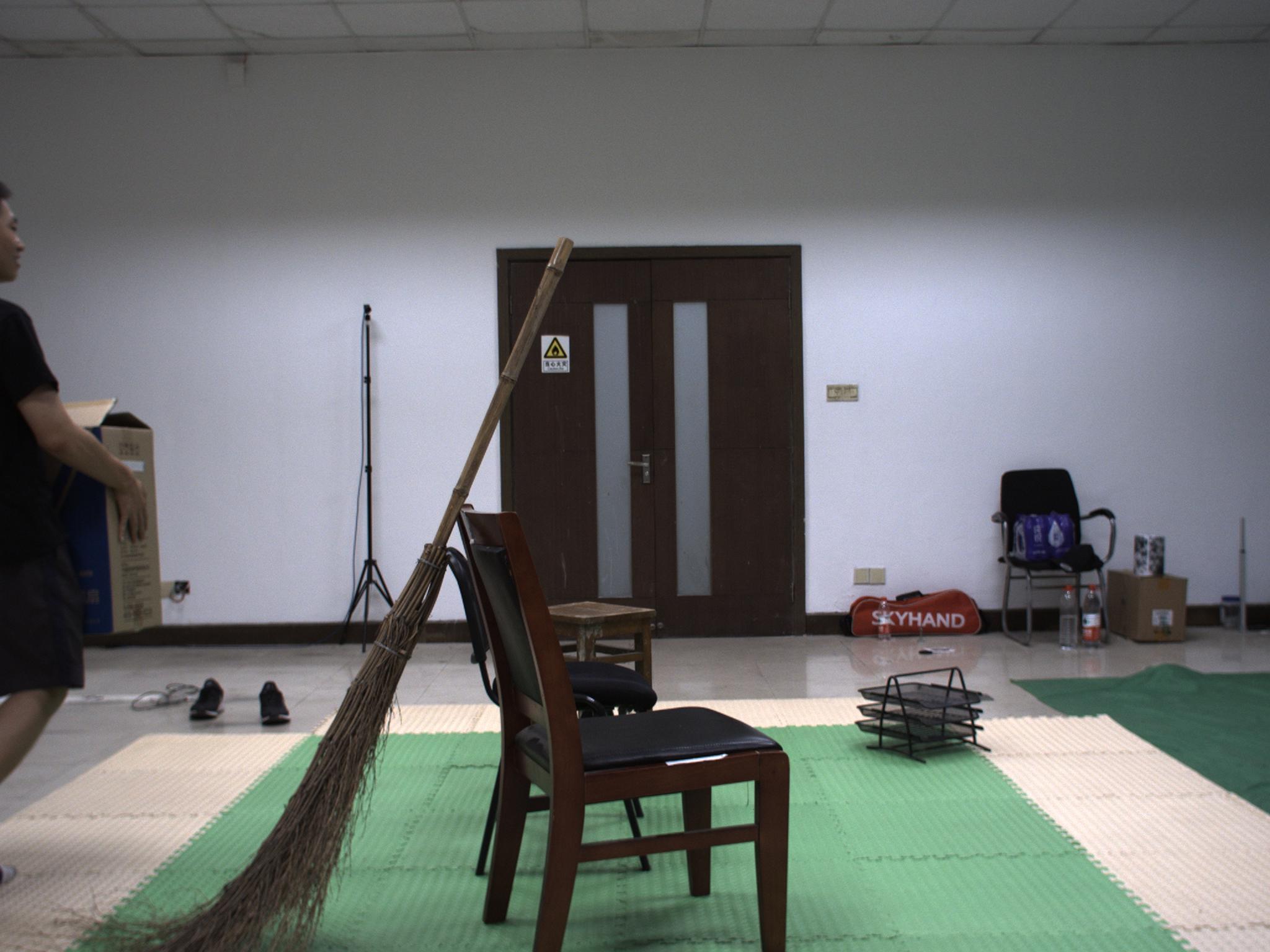}
    \end{subfigure}\hfill%
    \begin{subfigure}{0.11\linewidth}
    \centering
        \includegraphics[width=\linewidth]{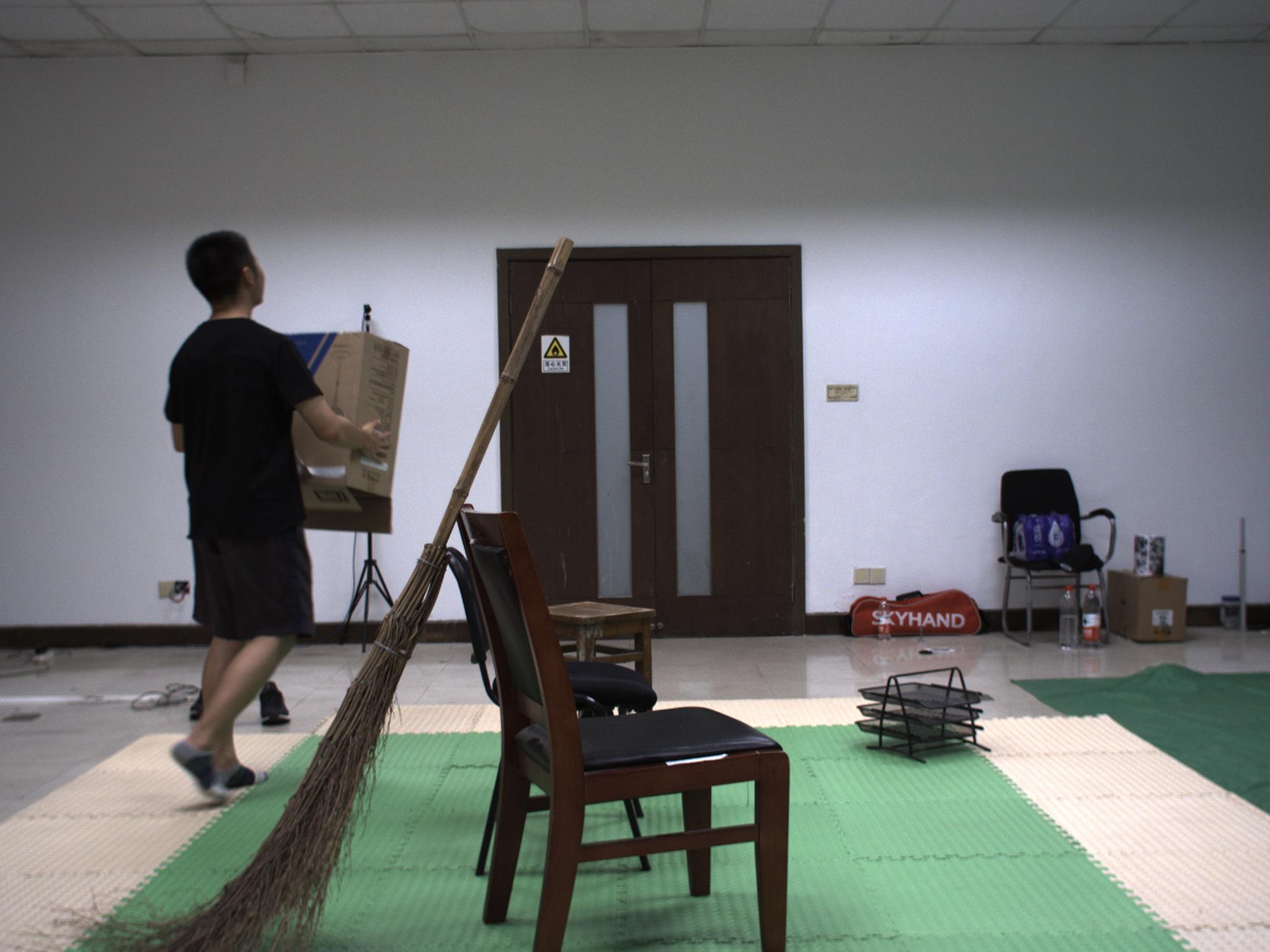}
    \end{subfigure}\hfill%
    \begin{subfigure}{0.11\linewidth}
\centering
    \includegraphics[width=\linewidth]{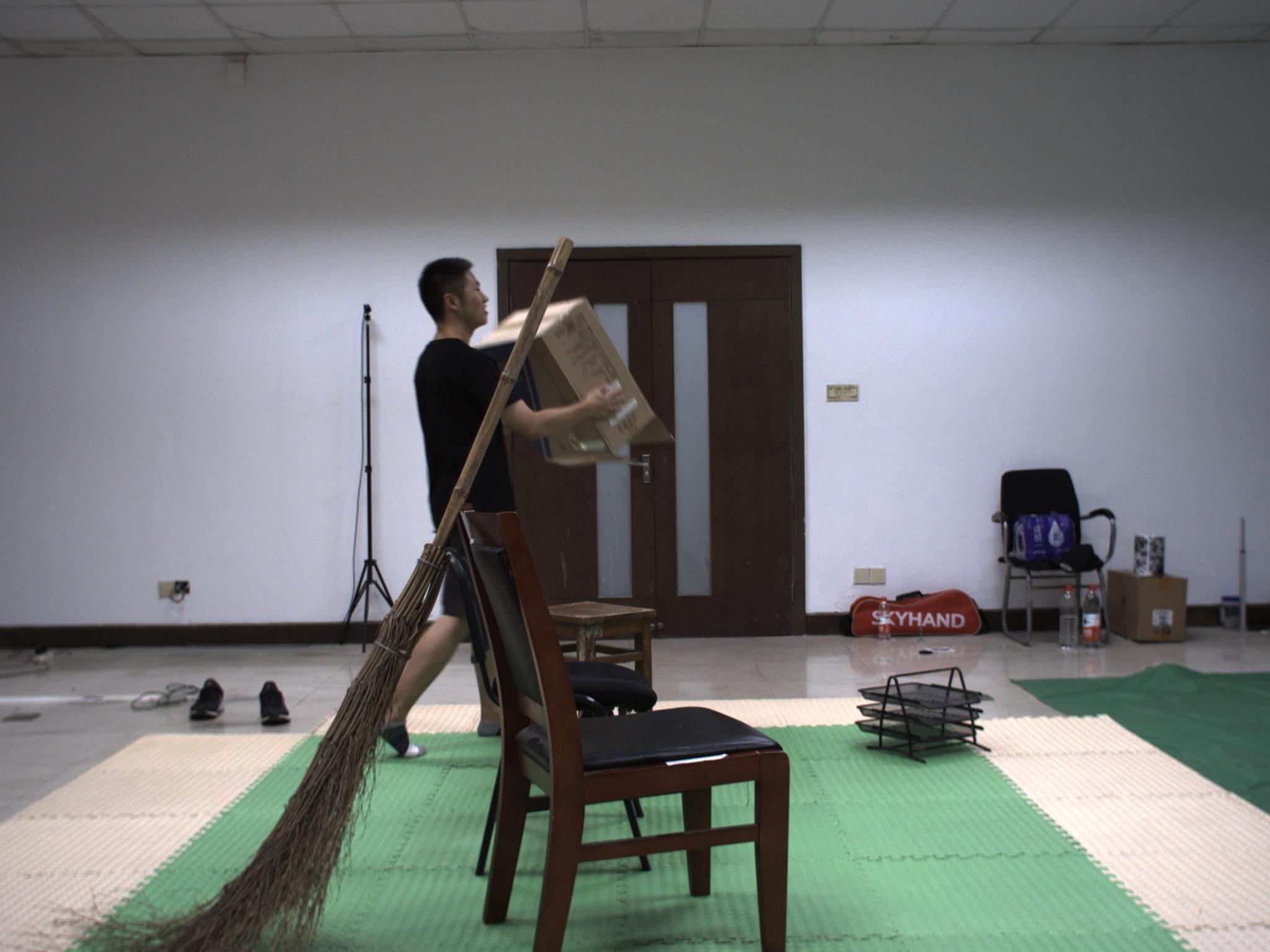}
\end{subfigure}\hfill%
    \begin{subfigure}{0.11\linewidth}
    \centering
        \includegraphics[width=\linewidth]{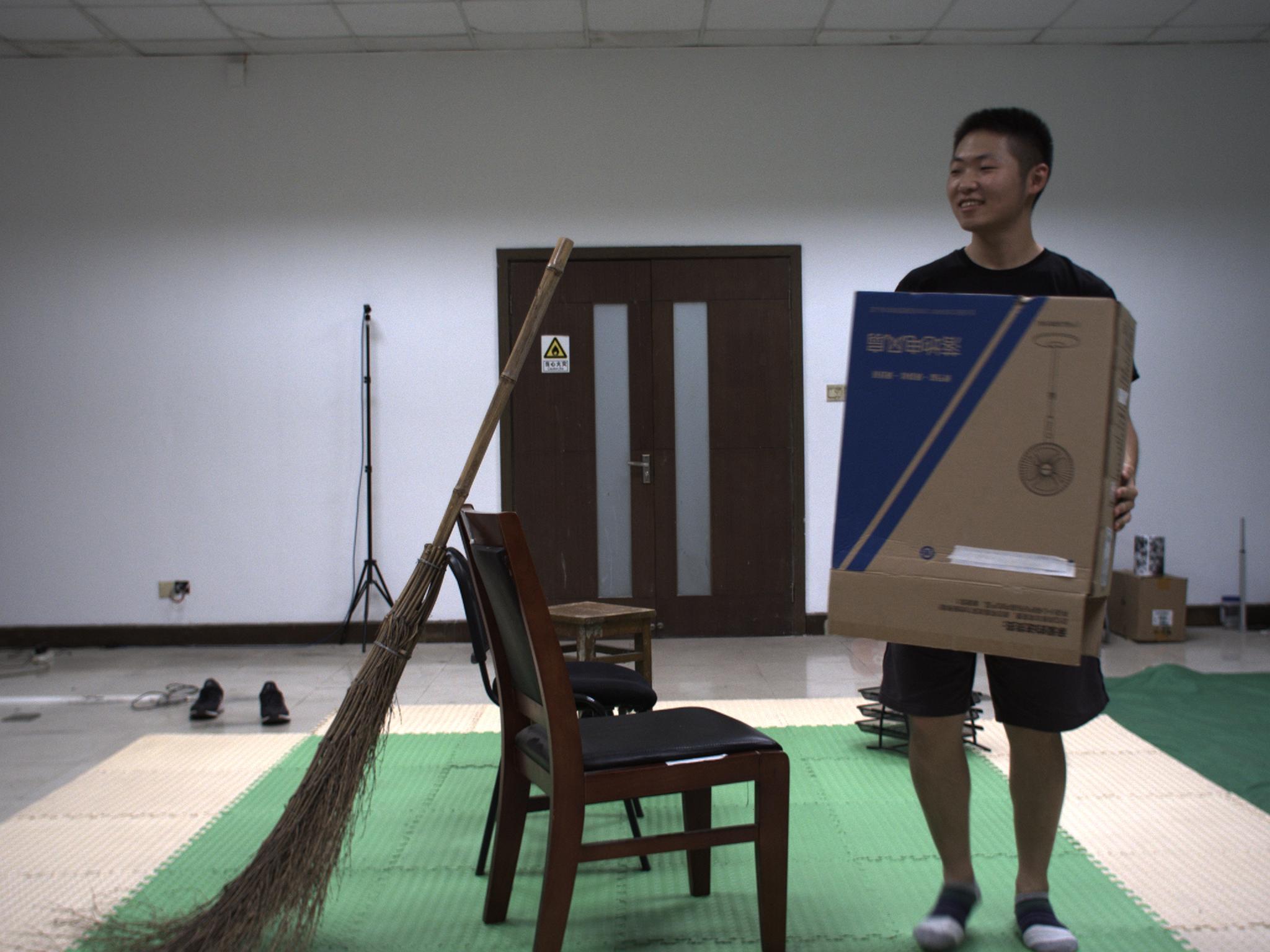}
    \end{subfigure}\hfill%
    \begin{subfigure}{0.11\linewidth}
    \centering
        \includegraphics[width=\linewidth]{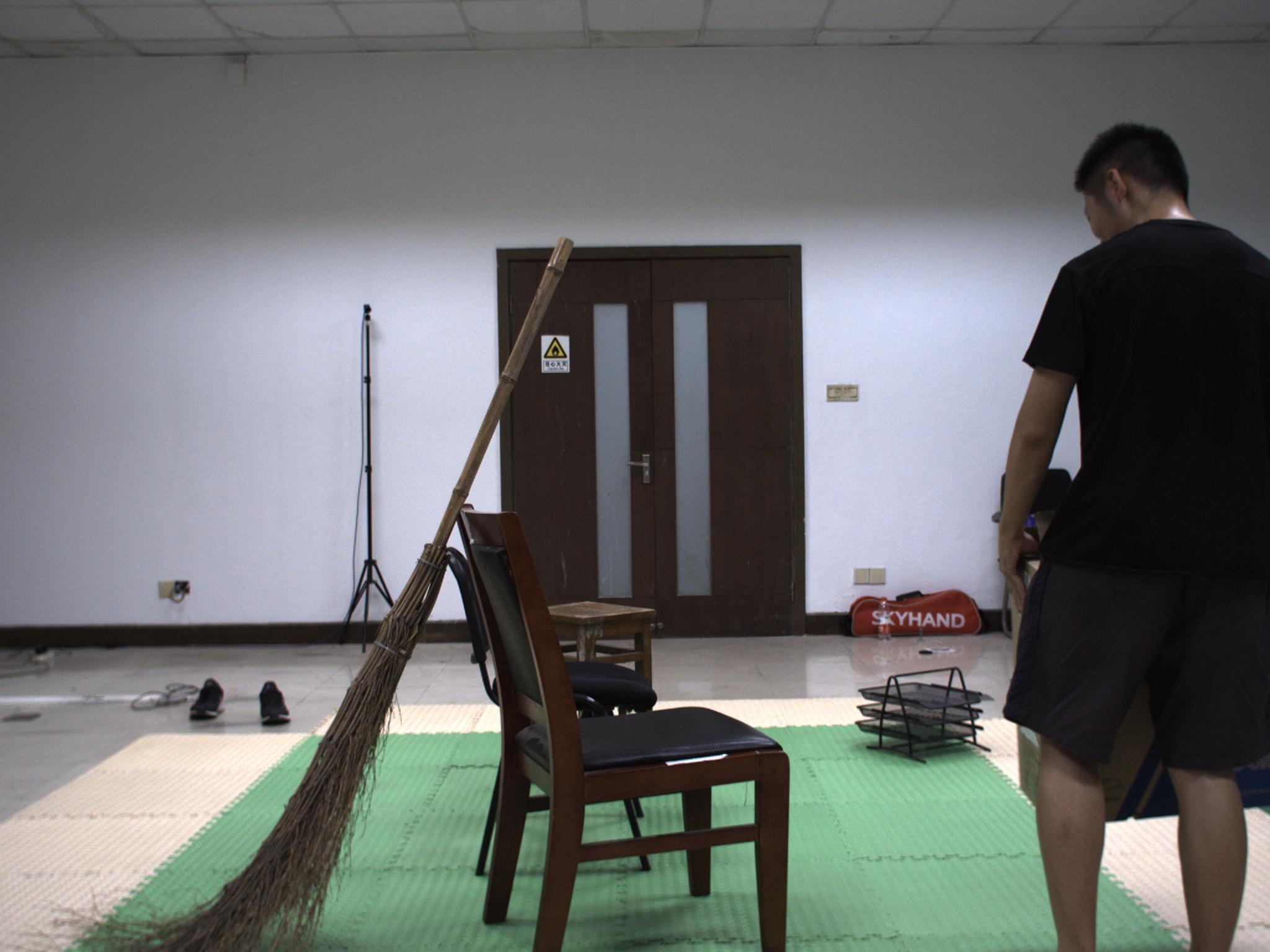}
    \end{subfigure}\hfill%
    \begin{subfigure}{0.11\linewidth}
        \centering
        \includegraphics[width=\linewidth]{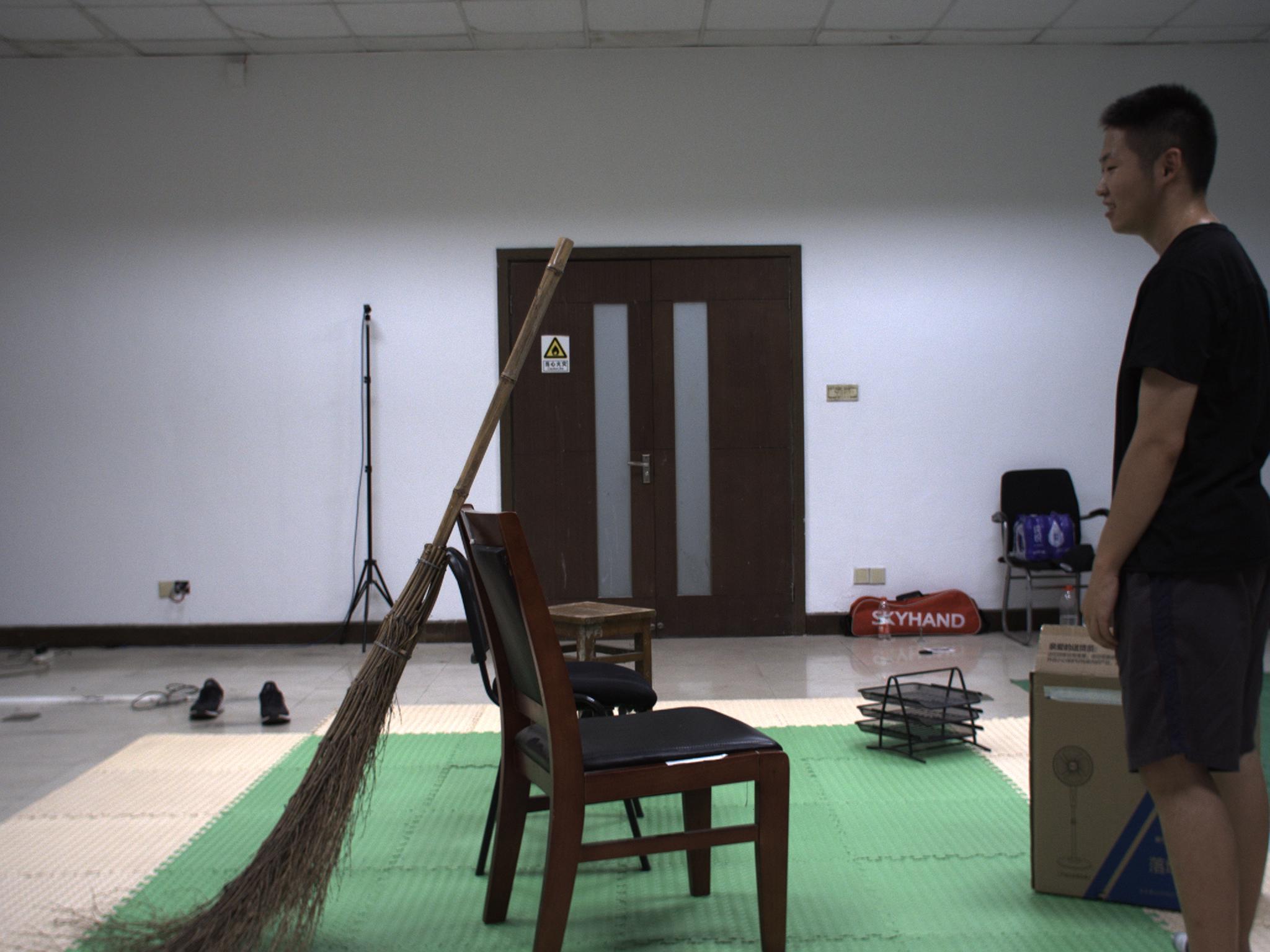}
    \end{subfigure}
    \\
    \begin{subfigure}{0.11\linewidth}
    \centering
        \includegraphics[width=\linewidth]{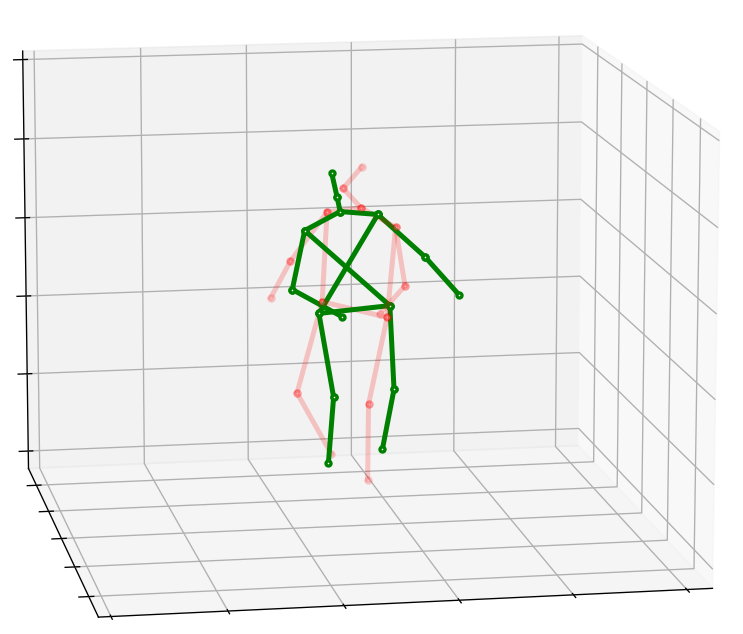}
    \end{subfigure}\hfill%
    \begin{subfigure}{0.11\linewidth}
    \centering
        \includegraphics[width=\linewidth]{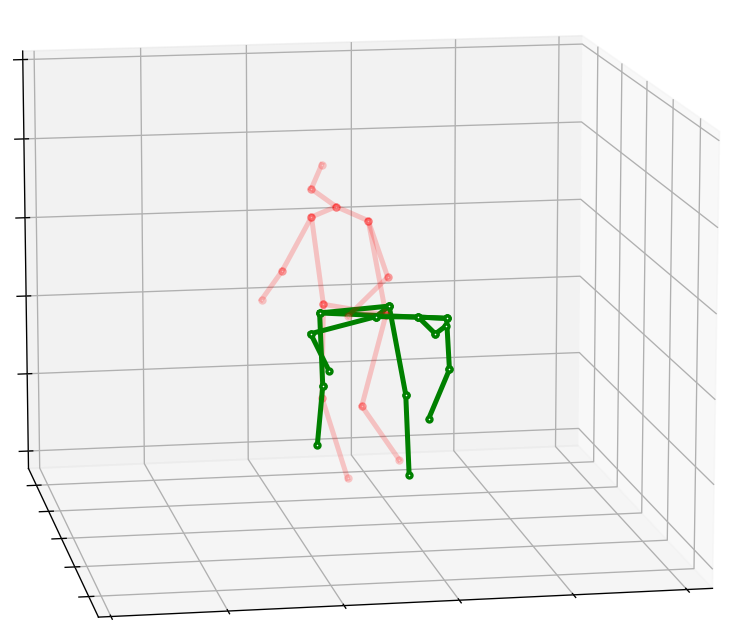}
    \end{subfigure}\hfill%
    \begin{subfigure}{0.11\linewidth}
    \centering
        \includegraphics[width=\linewidth]{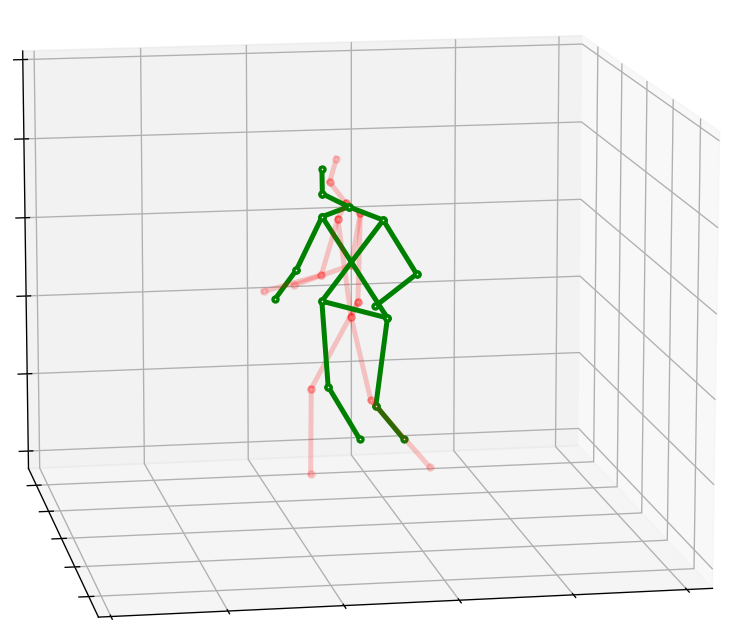}
    \end{subfigure}\hfill%
    \begin{subfigure}{0.11\linewidth}
     \centering
       \includegraphics[width=\linewidth]{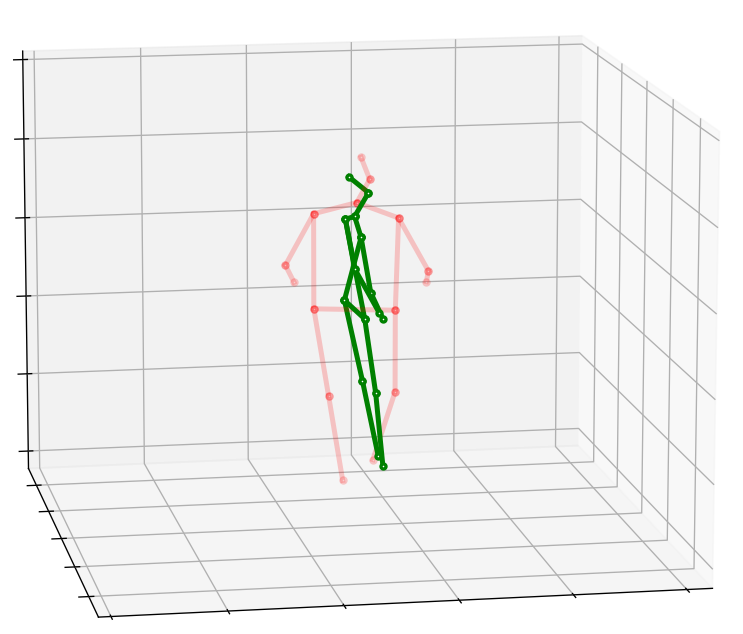}
    \end{subfigure}\hfill%
    \begin{subfigure}{0.11\linewidth}
    \centering
        \includegraphics[width=\linewidth]{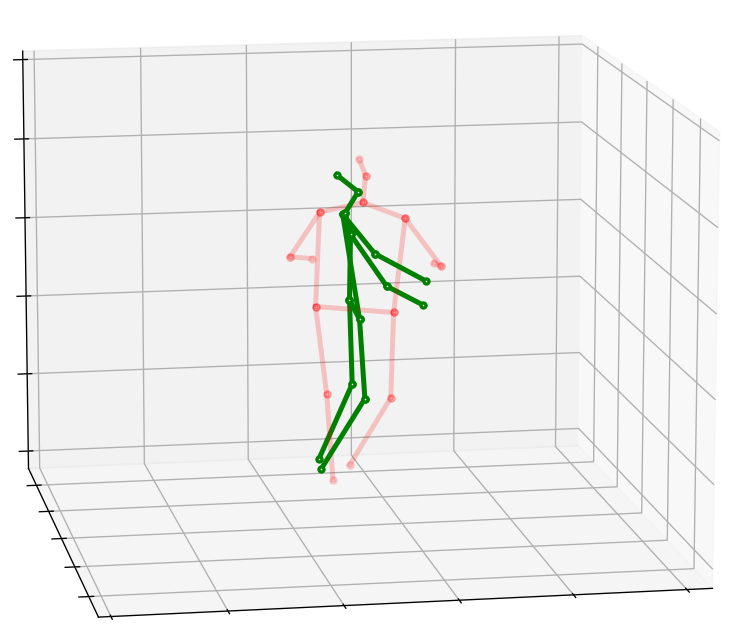}
    \end{subfigure}\hfill%
    \begin{subfigure}{0.11\linewidth}
\centering
    \includegraphics[width=\linewidth]{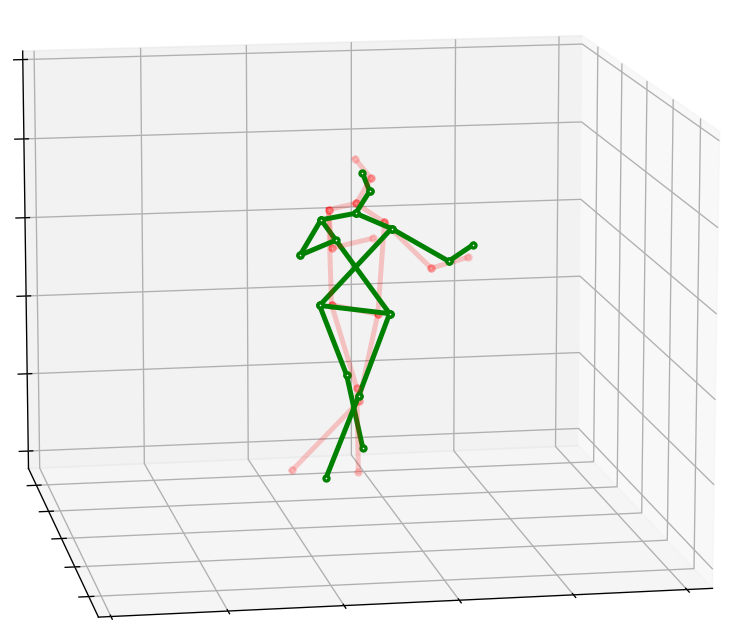}
\end{subfigure}\hfill%
    \begin{subfigure}{0.11\linewidth}
    \centering
        \includegraphics[width=\linewidth]{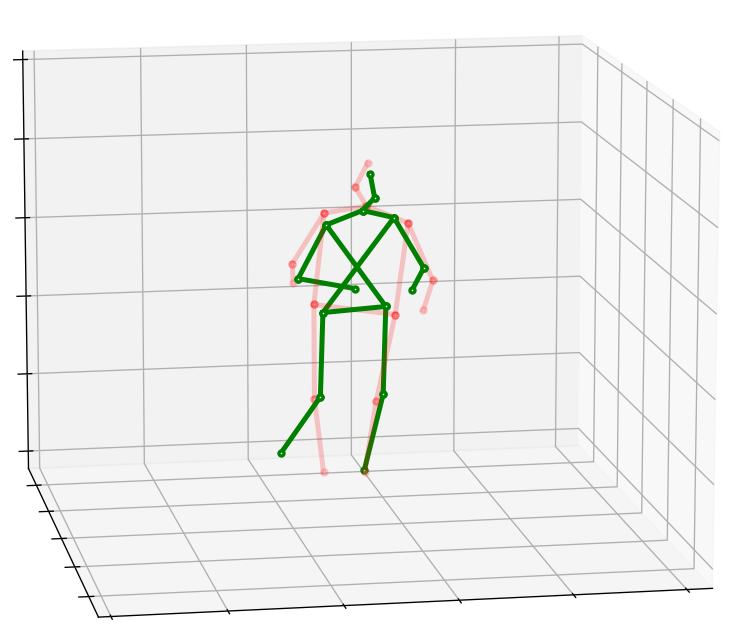}
    \end{subfigure}\hfill%
    \begin{subfigure}{0.11\linewidth}
    \centering
        \includegraphics[width=\linewidth]{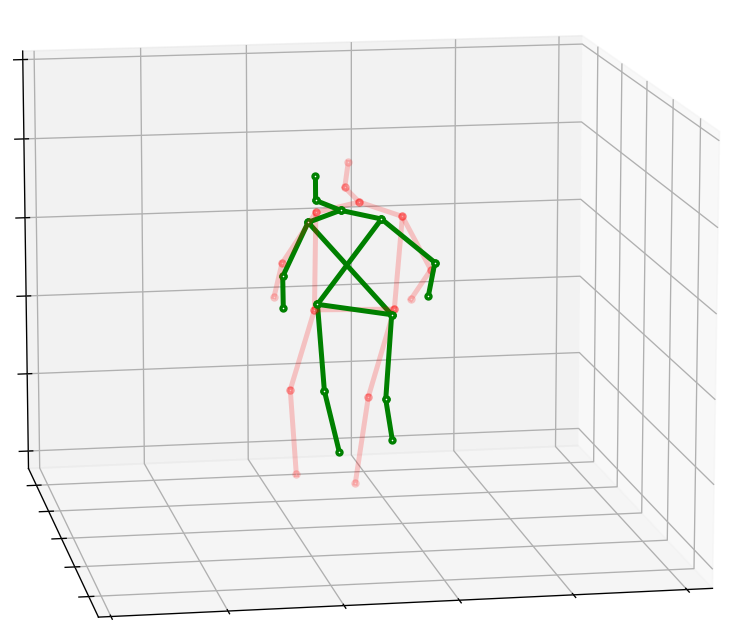}
    \end{subfigure}\hfill%
    \begin{subfigure}{0.11\linewidth}
        \centering
        \includegraphics[width=\linewidth]{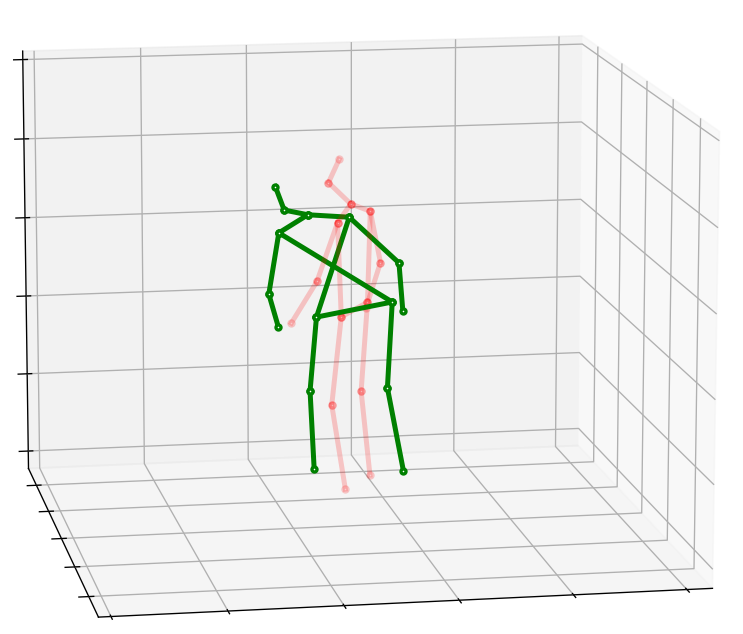}
    \end{subfigure}
    \\
    \begin{subfigure}{0.11\linewidth}
    \centering
        \includegraphics[width=\linewidth]{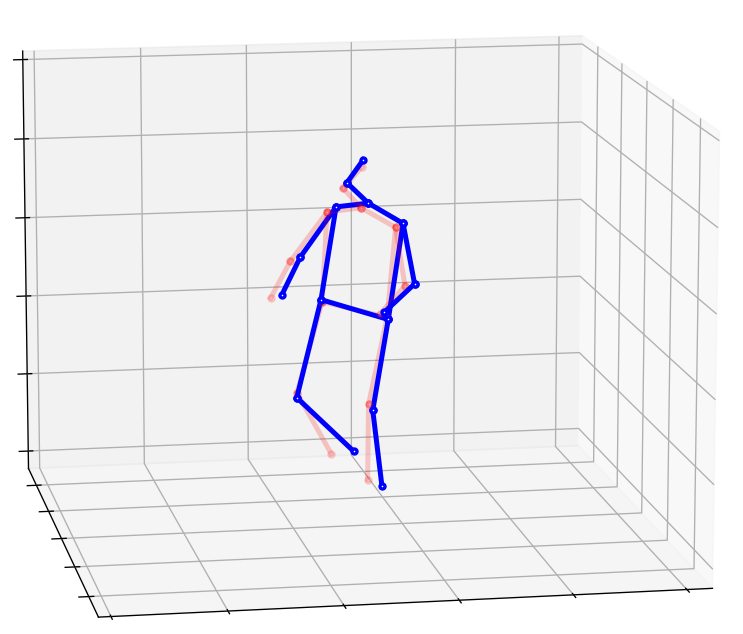}
    \end{subfigure}\hfill%
    \begin{subfigure}{0.11\linewidth}
    \centering
        \includegraphics[width=\linewidth]{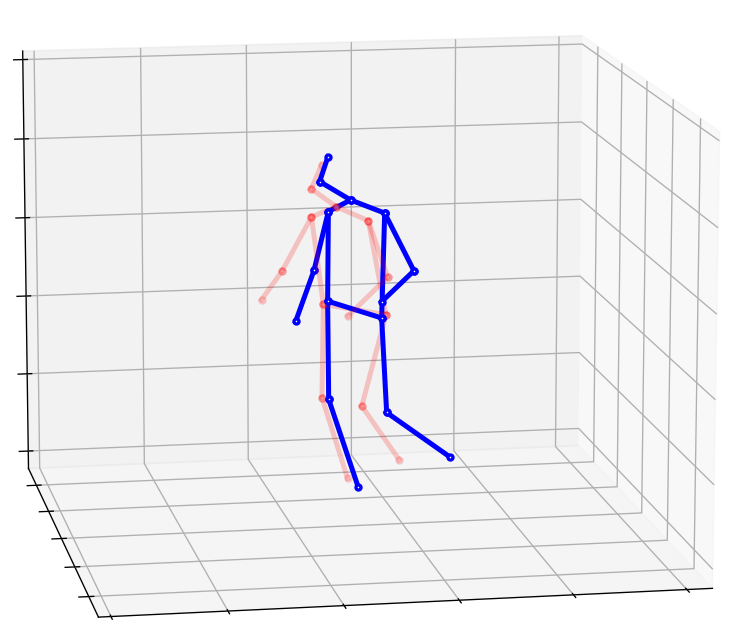}
    \end{subfigure}\hfill%
    \begin{subfigure}{0.11\linewidth}
    \centering
        \includegraphics[width=\linewidth]{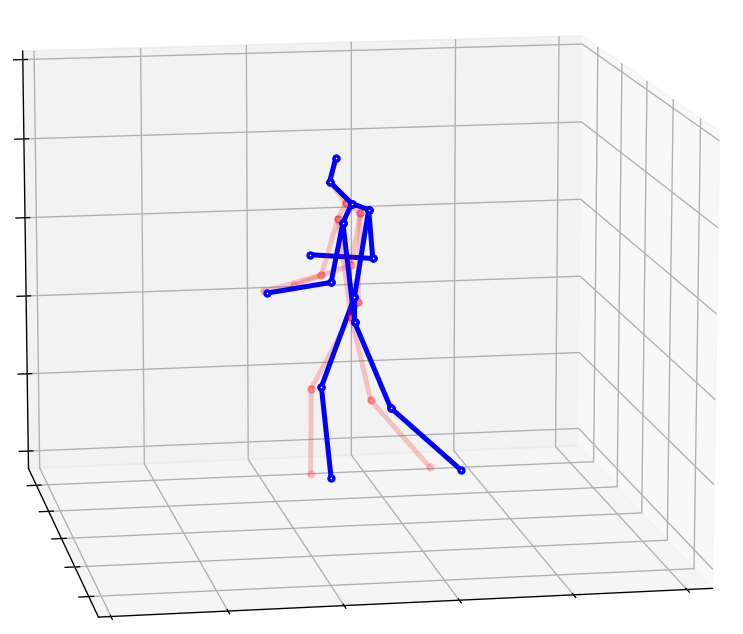}
    \end{subfigure}\hfill%
    \begin{subfigure}{0.11\linewidth}
     \centering
       \includegraphics[width=\linewidth]{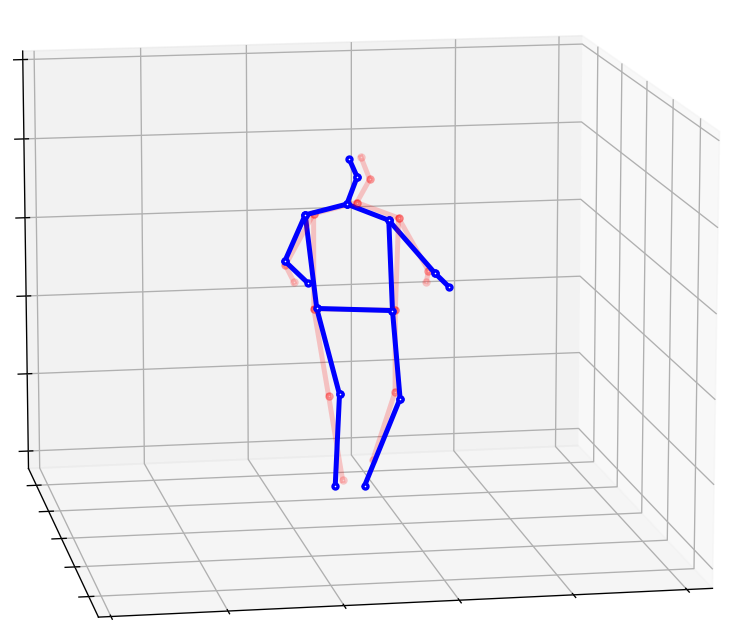}
    \end{subfigure}\hfill%
    \begin{subfigure}{0.11\linewidth}
    \centering
        \includegraphics[width=\linewidth]{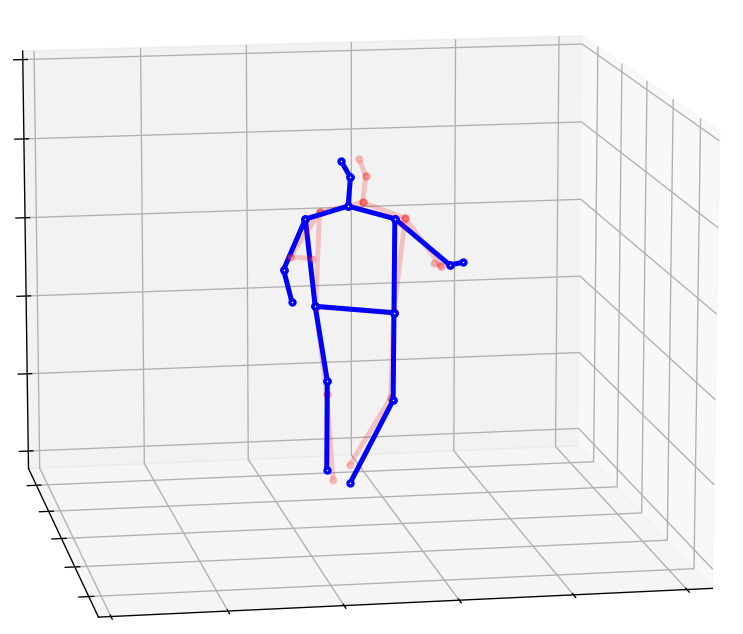}
    \end{subfigure}\hfill%
    \begin{subfigure}{0.11\linewidth}
\centering
    \includegraphics[width=\linewidth]{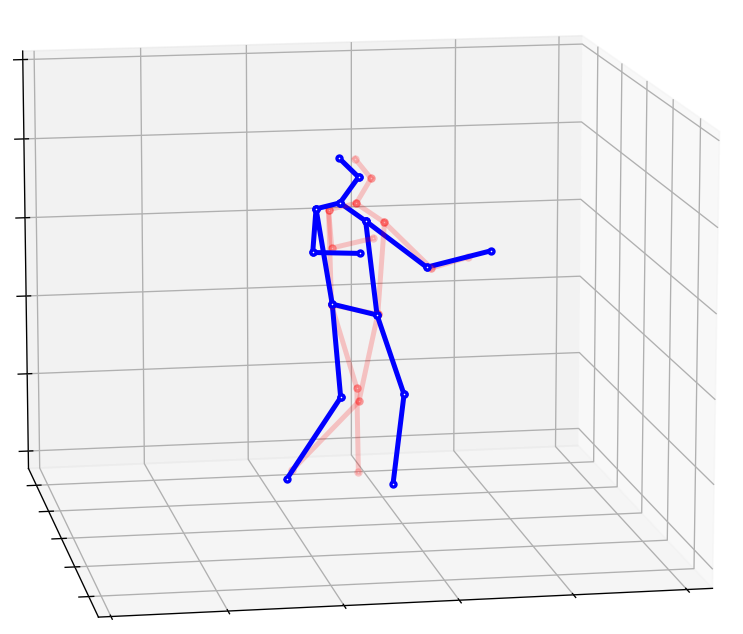}
\end{subfigure}\hfill%
    \begin{subfigure}{0.11\linewidth}
    \centering
        \includegraphics[width=\linewidth]{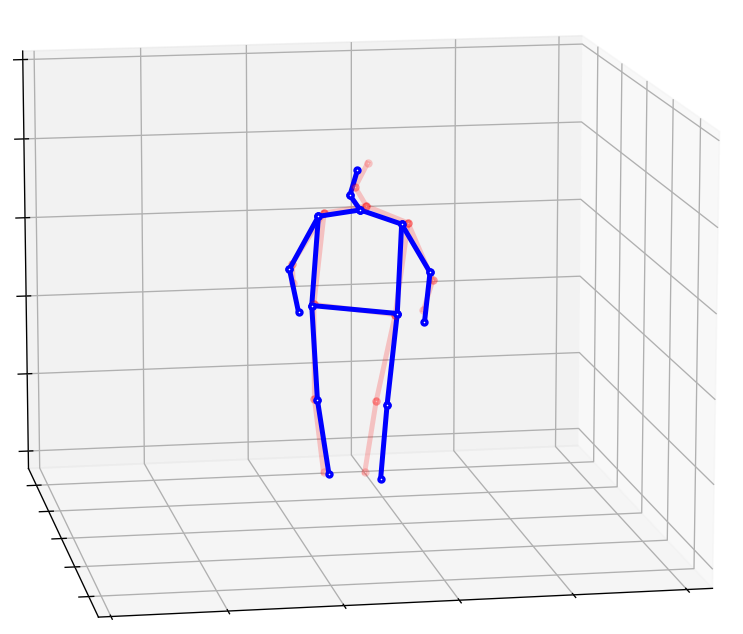}
    \end{subfigure}\hfill%
    \begin{subfigure}{0.11\linewidth}
    \centering
        \includegraphics[width=\linewidth]{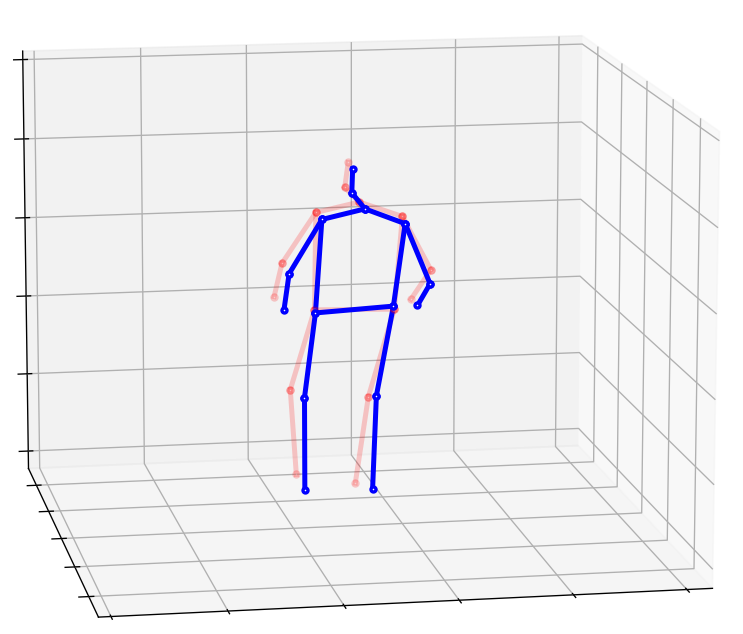}
    \end{subfigure}\hfill%
    \begin{subfigure}{0.11\linewidth}
        \centering
        \includegraphics[width=\linewidth]{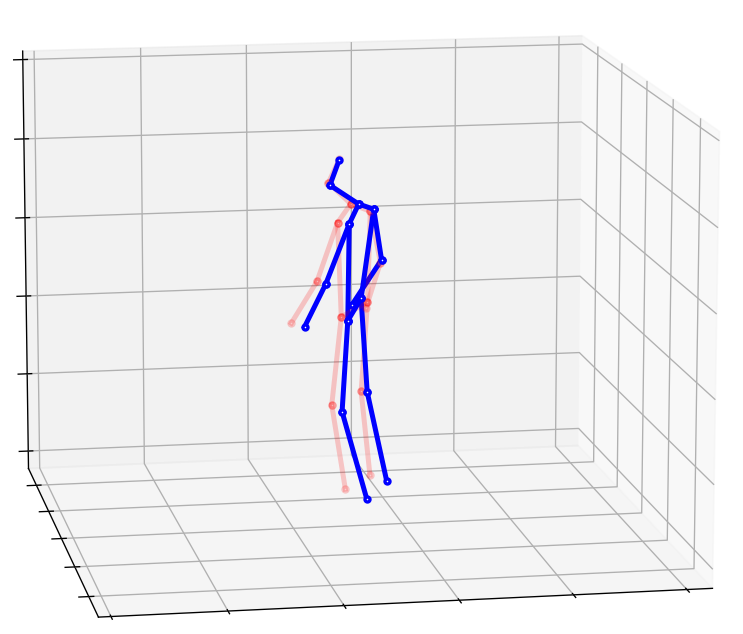}
    \end{subfigure}
    \caption{\textbf{3D pose estimation results on OCMotion (\textit{0013, Camera01})}. This figure demonstrates how our method incorporates temporal continuity into video sequences under occlusion. The \textit{second} row represents 3D poses predicted by CycleAdapt \cite{cycleadapt}. The \textit{third} row represents 3D poses predicted by \name. Note: The 3D poses shown in translucent red color in the \textit{second} and \textit{third} row represent the ground truths.
      }
    \label{fig:OCMotion_qualitative_res}
\end{figure*}

\noindent \textbf{Inference speed:} Table~\ref{tab:speed} compares the inference times of various 3D pose estimation methods on a 243-frame OCMotion video using an RTX 3090 GPU. HuMor and GLAMR are notably slower, exceeding 10 minutes due to their intensive pose optimization phase. In contrast, PoseFormerV2 and CycleAdapt show efficiency improvements with inference times of 129 and 126 seconds, respectively. \name~ outperforms these, achieving a significant reduction to 68 seconds, making it 46\% faster and highlighting its suitability for real-time applications without sacrificing accuracy. We have ensured that the testing is consistent across all algorithms, and results are reported \emph{including the TTT phase with 30 epochs}. Our method takes less time because we train only on a single video at test time, which contains significantly fewer frames compared to the entire dataset. 

\begin{table}[h]
\centering
\resizebox{0.3\textwidth}{!}{
    \begin{tabular}{lc}
    \toprule
    Method & Time (sec) \\
    \midrule
    HuMor~\cite{rempe2021humor} & $>600$ \\
    GLAMR~\cite{yuan2022glamr} & $>600$ \\
    PoseFormerV2~\cite{zhao2023poseformerv2} & $129$ \\
    CycleAdapt~\cite{cycleadapt} & $126$ \\
    \midrule
    {\name~(ours)} & \textbf{68} \scriptsize{{(46\%$\downarrow$)}} \\
    \bottomrule
    \end{tabular}}
\caption{Inference time for various 3D pose estimation methods.}
\label{tab:speed}
\end{table}
\subsection{Qualitative Results}

To provide a comprehensive analysis and comparison of \name~ against other methods, we have compiled and shared several qualitative video results in the supplementary material.
Our evaluation juxtaposes \name~ against leading state-of-the-art techniques like CycleAdapt \cite{cycleadapt}. Key insights from our comparison include: \\
\noindent \textbf{Occluded Human3.6M}: In this proposed dataset, traditional approaches often fall short in accurately predicting missing 3D poses, struggling with high levels of occlusion. In contrast, our method utilizes the dynamics of human motion to precisely infill missing poses, leading to a 57\% error improvement compared to the former methods. \\
\noindent \textbf{BRIAR} \cite{cornett2023expanding}: The videos within the BRIAR dataset present a substantial domain shift, a scenario not previously encountered by existing methodologies. Our algorithm distinguishes itself by mitigating these distribution shifts, resulting in markedly superior performance. While other techniques yield almost random predictions under these conditions, our method dynamically adapts to this domain shift during test time. Although direct quantitative comparisons are impossible due to the absence of ground truth 3D pose data on BRIAR, the visual comparisons provided through our videos convincingly demonstrate our method's enhanced adaptability and efficacy. \\
\noindent \textbf{OCMotion} \cite{huang2022occluded}: In Fig.~\ref{fig:OCMotion_qualitative_res}, we compare our method against an existing state-of-the-art pose estimation method CycleAdapt~\cite{cycleadapt}. In Frame 5, we can observe that CycleAdapt fails to perform well in cases when there is self-occlusion. We observe that \name's predictions are best aligned with the ground truth poses, even under significant occlusions or when the person goes out of the frame. 

Please refer to the Section 4 supplementary material for additional qualitative 3D pose estimation results and Section 6 of supplementary material for details on mesh generation in videos and mesh recovery results.

\subsection{Ablation Study}
\begin{table}[]
    \centering
    \setlength{\tabcolsep}{3pt} 
    \resizebox{0.95\linewidth}{!}{
    \begin{tabular}{ccccccc}
    \toprule
    $\mathcal{M}$ &$\mathcal{L}_{mpjp}$ & $\mathcal{L}_{vel}$ & $\mathcal{L}_{lim}$ & $\mathcal{L}_{nmpjp}$ & MPJPE & PA-MPJPE \\
    \midrule
    \xmark & \xmark & \xmark & \xmark & \xmark & 179.5 & 98.9 \\
    \cmark & \xmark & \xmark & \xmark & \xmark & 106.5 & 80.2 \\
    \cmark & \cmark & \xmark & \xmark & \xmark & 82.1 & 60.4 \\
    \cmark & \cmark & \cmark & \xmark & \xmark & 81.4 & 59.6 \\
    \cmark & \cmark & \cmark & \cmark & \xmark & 81.1 & 59.6 \\
    \cmark & \cmark & \cmark & \cmark & \cmark & \textbf{80.7} & \textbf{59.0} \\
    \bottomrule
    \end{tabular}
    }
    \caption{\textbf{Ablation study.} This table illustrates how integrating a pre-trained motion prior and various losses collectively contribute to \name's final accuracy on the Occluded Human3.6M dataset.}
    \label{tab:ablation}
\end{table}
An ablation study conducted in Table~\ref{tab:ablation} provides insights into the significance of each component in \name.  Starting from a baseline with substantial errors, introducing a motion prior alone drastically improves performance, underscoring its effectiveness in driving the model toward realistic human pose dynamics. Adding $L_{mpjp}$ enhances spatial accuracy, further lowering MPJPE to 82.1 and PA-MPJPE to 60.4. Improvement with $L_{vel}$ suggests its role in smoothing motion. The best results are observed when $L_{nmpjp}$ is also included, indicating its critical function in accounting for scale variations.
Thus, the ablation study reveals that each component contributes to improving the accuracy and temporal consistency of the pose estimations, with the full combination of components yielding state-of-the-art results. This shows that while the motion prior sets a strong foundation for plausible poses, the various loss functions refine and stabilize the pose predictions to align closely with natural human movement dynamics and unseen poses. In Section 5 of supplementary material, we show how varying off-the-shelf pose estimation methods within the backbone of \name~affects its performance. We find that using any off-the-shelf pose estimation method yields similar improvements, thereby making \name~agnostic to any specific 3D pose estimation method.
\section{Conclusion}
\label{sec:conclusion}
We introduce \name, a novel algorithm for self-supervised test-time training aimed at improving 3D human pose estimation in individual video frames. \name~utilizes extensive self-supervised pre-training to develop a robust model of human motion priors. It integrates self-supervised optimization with temporal regularization, achieving state-of-the-art performance in terms of both pose accuracy and computational efficiency across diverse challenging datasets, even those with significant occlusions. A limitation of \name~is its ability to extract temporally continuous 3D poses only in scenarios where there are no human-to-human occlusions. Future efforts can  concentrate on adapting \name~ for situations involving multi-person occlusions. Handling such situations requires modelling of complex human-to-human interaction alongside external identification and tracking framework. In conclusion, \name~sets a new benchmark for 3D human pose estimation in occluded environments and introduces promising directions for enhancing related downstream applications.

\section*{Acknowledgment} 
\noindent The work was partially supported by NSF grants 2326309 and 2312395, DURIP grant N000141812252, USDA grant 2021-67022-33453, and the Office of the Director of National Intelligence (ODNI), specifically through the Intelligence Advanced Research Projects Activity (IARPA), under contract number [2022- 21102100007]. The views and conclusions in this research reflect those of the authors and should not be construed as officially representing the policies, whether explicitly or implicitly, of ODNI, IARPA, or the U.S. Government. Nevertheless, the U.S. Government retains the authorization to reproduce and distribute reprints for official government purposes, regardless of any copyright notices included.


{\small
\bibliographystyle{ieee_fullname}
\bibliography{egbib}
}

\end{document}


\maketitle



\section{Network Architecture}
\label{suppl:sec:network_arch}
$\mathcal{M}$ contains two key components: 1) a \emph{spatial block} to capture the orientation of joints, and 2) a \emph{temporal block} to model the temporal dynamics of a joint. The spatial block refines poses in each frame, while the temporal block smooths the transitions between frames. We describe these components below:\\
\noindent \textbf{Spatial block.}
This block utilizes \emph{Spatial Multi-Head Self-Attention} (S-MHSA) to model relationships among joints within each pose in the input sequence. Mathematically, the S-MHSA operation is defined as:
\begin{align*}\small
\begin{split}
\text{S-MHSA}(\mathbf{Q}_{\text{S}}, \mathbf{K}_{\text{S}}, \mathbf{V}_{\text{S}})&= [\texttt{head}_{1};...;\texttt{head}_{h}]\mathbf{W}_\text{S}^{P} ;\;
\\
\text{head}_{i}&=\text{softmax}( \frac{\mathbf{Q}_\text{S}^{i}(\mathbf{K}_\text{S}^{i})^{\texttt{T}}}{\sqrt{d_K}})\mathbf{V}_\text{S}^{i}
\end{split}
\end{align*}
Here, $\mathbf{Q}_{\text{S}}^{i}, \mathbf{K}_{\text{S}}^{i}, \mathbf{V}_{\text{S}}^{i}$ denote the query, key, and value projections for the $i^{th}$ attention head, $d_k$ is the key dimension, and $\mathbf{W}_{\text{S}}^{P}$ is the projection parameter matrix. 
We apply S-MHSA to features of different time steps in parallel. The output undergoes further processing, including residual connection and layer normalization (LayerNorm), followed by a multi-layer perceptron (MLP). \\
\noindent\textbf{Temporal block.}
This block utilizes \emph{Temporal Multi-Head Self-Attention} (T-MHSA) to model the relationships between poses across time steps, thereby enabling the smoothing of the pose trajectories over the sequence.  It operates similarly to S-MHSA but is applied to per-joint temporal features parallelized over the spatial dimension:
\begin{align*}\small
\begin{split}
\text{T-MHSA}(\mathbf{Q}_{\text{T}}, \mathbf{K}_{\text{T}}, \mathbf{V}_{\text{T}}) &= [\texttt{head}_{1};...;\texttt{head}_{h}]\mathbf{W}_\text{T}^{P};\;
\\
\texttt{head}_{i} &= \text{softmax}(\frac{\mathbf{Q}_{\text{T}}^{i}{(\mathbf{K}_{\text{T}}^{i}})^{\texttt{T}}}{\sqrt{d_K}})\mathbf{V}_{\text{T}}^{i}
\end{split}
\end{align*}
By attending to temporal relationships, T-MHSA produces smooth pose transitions over time.

\noindent \textbf{Dual-Stream Spatio-temporal Transformer.} We then use the dual-stream architecture which employs spatial and temporal Multi-Head Self-Attention mechanisms. These mechanisms capture intra-frame and inter-frame body joint interactions, necessitating careful consideration of three key assumptions: both streams model comprehensive spatio-temporal contexts, each stream specializes in distinct spatio-temporal aspects, and the fusion dynamically balances weights based on input characteristics.

\begin{figure*}[t]
  \centering
  \begin{minipage}{0.32\textwidth}
    \centering
    \includegraphics[width=\linewidth]{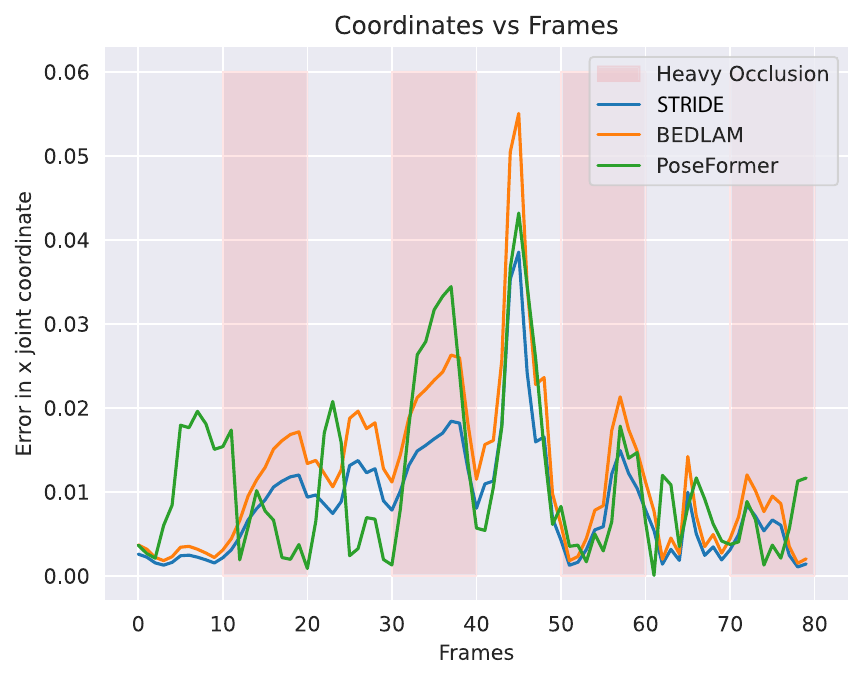}
  \end{minipage}
  \hfill
  \begin{minipage}{0.32\textwidth}
    \centering
    \includegraphics[width=\linewidth]{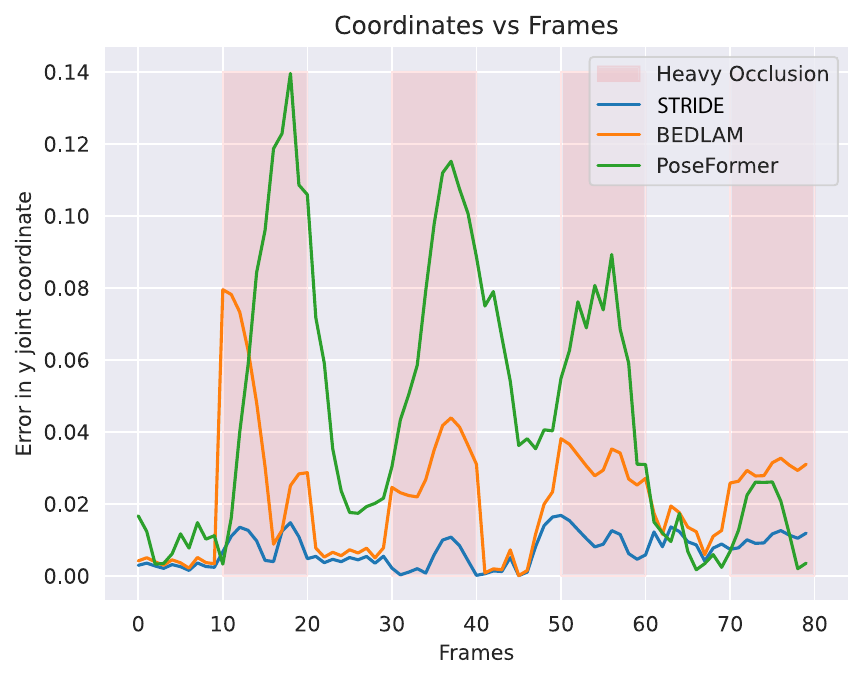}
  \end{minipage}
  \hfill
  \begin{minipage}{0.32\textwidth}
    \centering
    \includegraphics[width=\linewidth]{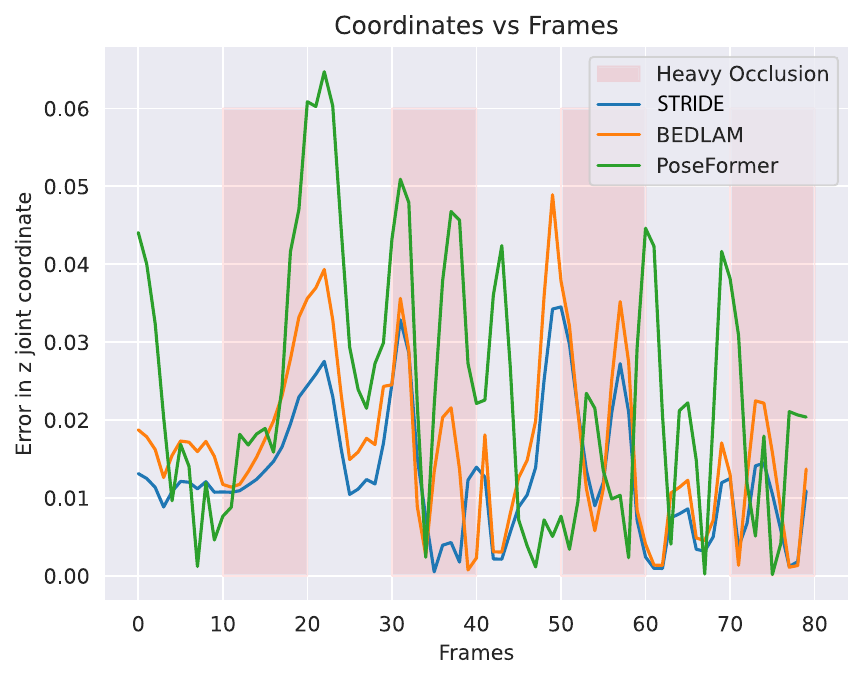}
  \end{minipage}
\caption{The figure above, from left to right, illustrates the variation in error values across the x, y, and z coordinates within a single video. Notably, \name~ exhibits relatively lower error, particularly in scenarios involving occlusion. Furthermore, for y-coordinate, it is evident that the error demonstrates a remarkable level of smoothness.}
\label{fig:error}
\end{figure*}

\section{Implementation Details}
We implement the proposed motion encoder DSTformer with depth N = 5, number of heads h = 8, feature size = 512, embedding size = 512. For pretraining, we use sequence length T = 243. The pretrained model could handle different input lengths thanks to the transformer-based backbone. During finetuning, we set the backbone learning rate to be 0.1× of the new layer learning rate. 

\textbf{Setup.} We have implemented the proposed model using PyTorch. For our experiments, we utilized a CentOS machine equipped with 4 NVIDIA 3090 GPUs, specifically designed for accelerating pretraining tasks. It's worth noting that for finetuning and inference processes, a single GPU typically proves to be more than adequate.

\textbf{Pretraining.} We do large scale pertaining using AMASS and Training split of Human3.6M. For the implementation of AMASS~\cite{mahmood2019amass}, we initiate the process by rendering the parameterized human model SMPL+H. Subsequently, we extract 3D keypoints using a predefined regression matrix. The extraction of 3D keypoints from the Human3.6M dataset is accomplished through camera projection. Motion clips with a length of $T=243$ are sampled for the 3D mocap data. The input channels are set to $C_\text{in} = 3$, representing the $(x,y,z)$ coordinate. Data augmentation is applied through random horizontal flipping.

The entire network undergoes training for a total of $90$ epochs, employing a learning rate of $0.0005$ and a batch size of $64$, facilitated by the Adam optimizer. The weights assigned to the loss terms are parameterized by $\lambda_\text{O}=20$. Additionally, we set the 3D skeleton masking ratio to $15\%$, aligning with BERT's configuration. This involves using $10\%$ frame-level masks and $5\%$ joint-level masks. Despite variations in the proportion of these mask types, only marginal differences are observed.

To ensure the smoothness of the noise and prevent severe jittering, we initially sample noise $\bm{z}\in\mathbb{R}^{T_{K} \times J}$ for $T_{K}=27$ keyframes. Subsequently, we upsample it to $\bm{z}'\in\mathbb{R}^{T \times J}$ and introduce a small Gaussian noise $\mathcal{N}(0, 0.002^2)$.

\textbf{3D Pose Estimation.} We conduct training during the inference stage for a duration of 30 epochs, employing the following hyperparameters:

\begin{itemize}
    \item Batch size: 1
    \item Learning rate: 0.0002
    \item Weight decay: 0.01
    \item Learning rate decay: 0.99
\end{itemize}

The total loss, denoted as
\[
\mathcal{L}_{\text{total}} = \lambda_1\mathcal{L}_{\text{mpjp}} + \lambda_2\mathcal{L}_{\text{vel}} + \lambda_3\mathcal{L}_{\text{lim}} + \lambda_4\mathcal{L}_{\text{nmpjp}},
\]
is comprised of multiple components, each weighted by specific coefficients. For this configuration, we set the weights as follows: $\lambda_1 = 1, \quad \lambda_2 = 20, \quad \lambda_3 = 200, \quad \lambda_4 = 0.5$.  These weightings contribute to the overall optimization objective, allowing for a fine-tuned balance during the training process.

\begin{figure}
    \centering
    \includegraphics[width=\linewidth]{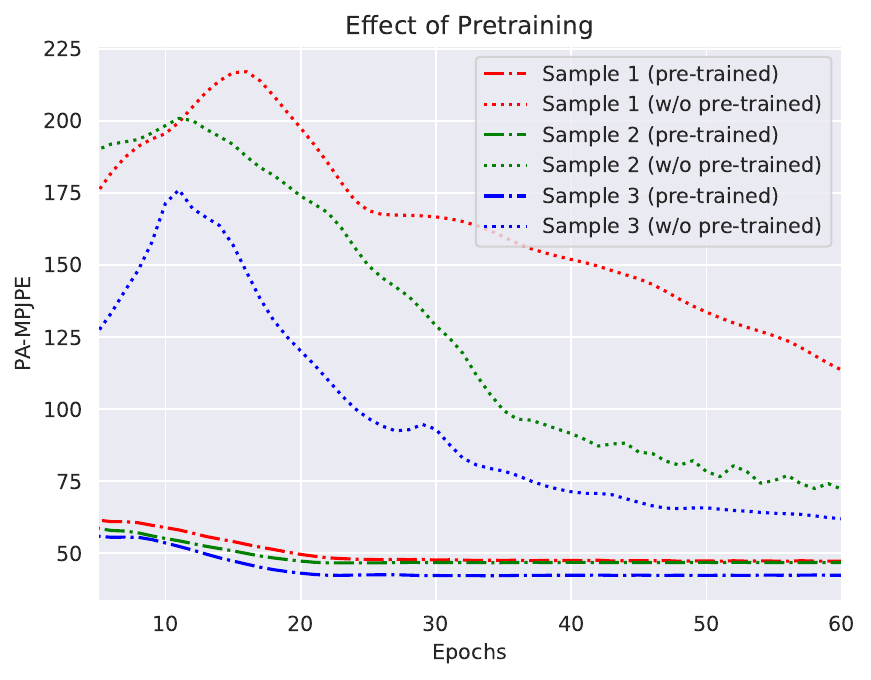}
    \caption{Effect of large-scale pre-training. We take 5 random samples from Occluded Human3.6M and try to align DSTFormer architecture. We find that when DSTFormner is initialised with motion-prior weights it converges faster.}
    \label{fig:effect_of_pretrain}
\end{figure}

\section{Temporal Smoothness}
The existing metric falls short in capturing temporal smoothness or assessing errors during occlusion. Additionally, there's a likelihood that a model excelling in occluded scenarios might not significantly impact overall performance if non-occluded cases dominate the results. This becomes particularly apparent in cases of sporadic temporal occlusion.

To address this issue and gain deeper insights into predictions during occlusions, we visualize various errors in Fig. \ref{fig:error}. This plot illustrates how the error in the x, y, and z coordinates evolves in a video featuring occlusions. Notably, other methods demonstrate subpar performance during occlusions, with the error in the x and z coordinates being relatively minimal, exerting less influence on the final error. In contrast, the y-coordinate error predominantly contributes to the overall error, where \name~stands out by consistently having the least amount of error. The noteworthy aspect is the sustained and consistent performance throughout the occluded duration.

\begin{figure*}[t]
    \centering
    \begin{subfigure}{0.23\textwidth}
        \centering
        \includegraphics[width=0.9\linewidth]{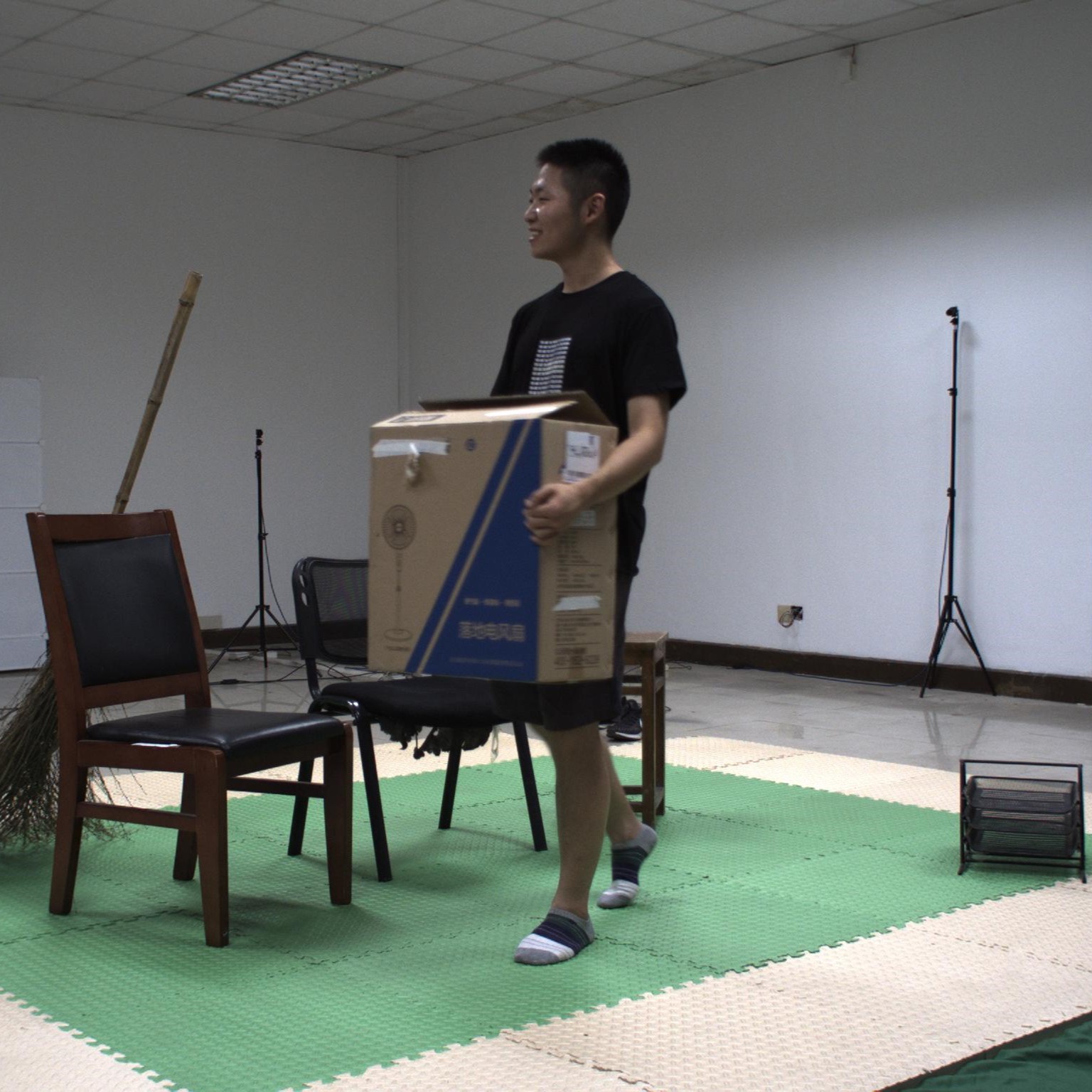}
    \end{subfigure}%
    \begin{subfigure}{0.23\textwidth}
        \centering
        \includegraphics[width=0.9\linewidth]{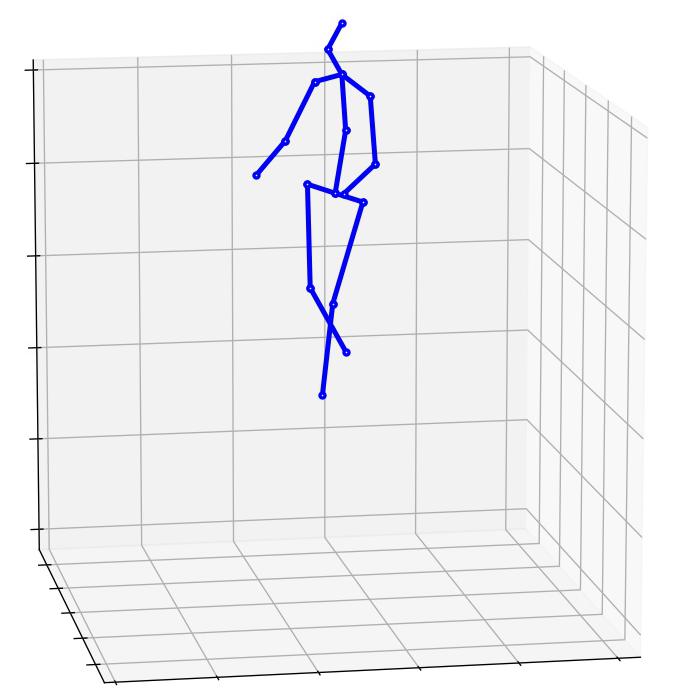}
    \end{subfigure}%
    \begin{subfigure}{0.23\textwidth}
        \centering

        \includegraphics[width=0.9\linewidth]{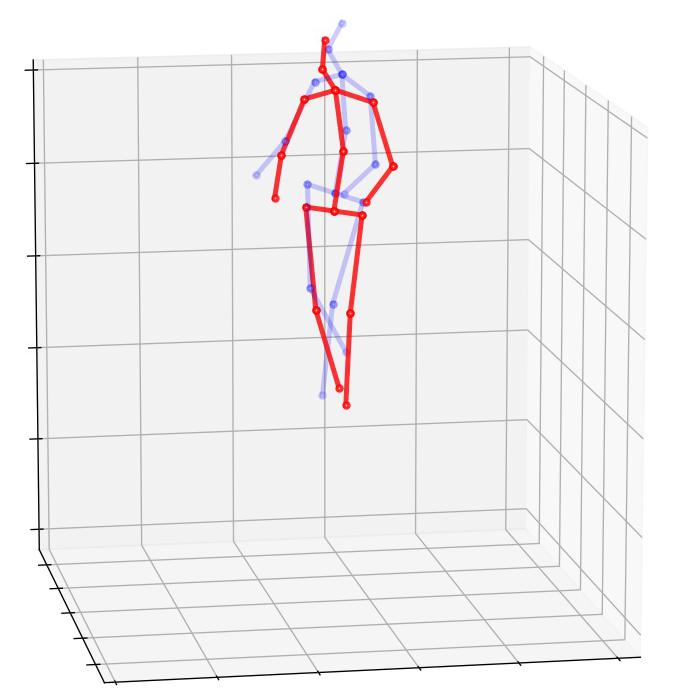}
    \end{subfigure}%
    \begin{subfigure}{0.23\textwidth}
        \centering

        \includegraphics[width=0.9\linewidth]{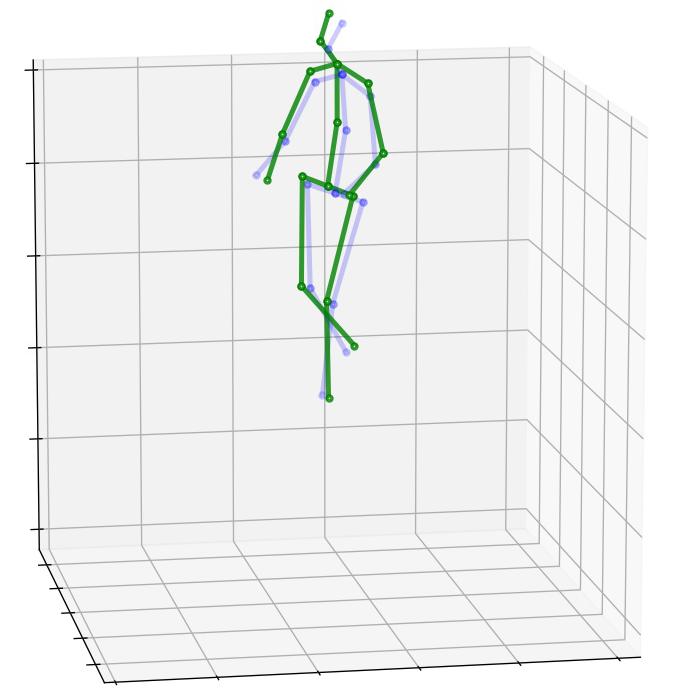}
    \end{subfigure}
    \\
    \begin{subfigure}{0.23\textwidth}
        \centering
        \includegraphics[width=0.9\linewidth]{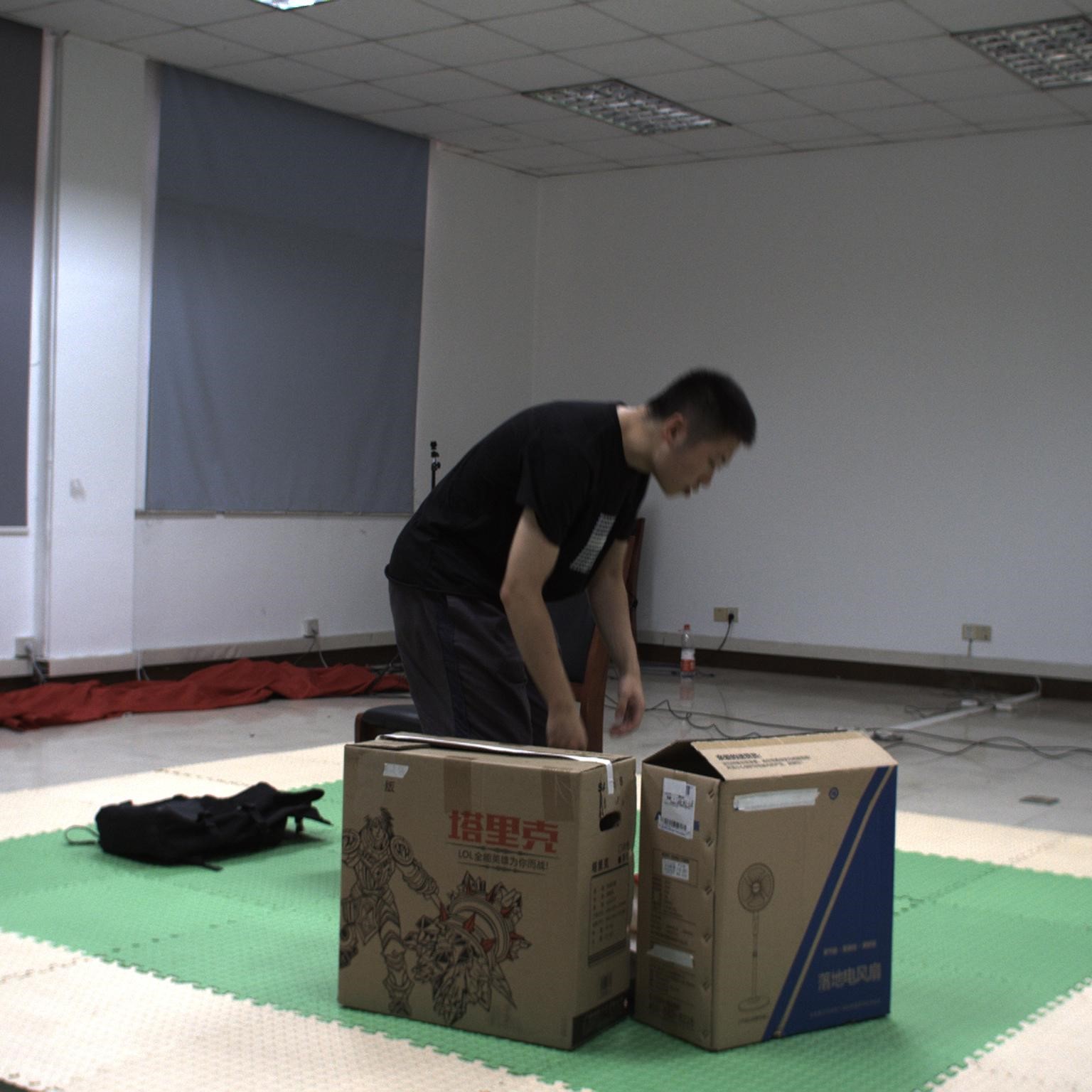}
    \end{subfigure}%
    \begin{subfigure}{0.23\textwidth}
        \centering
        \includegraphics[width=0.9\linewidth]{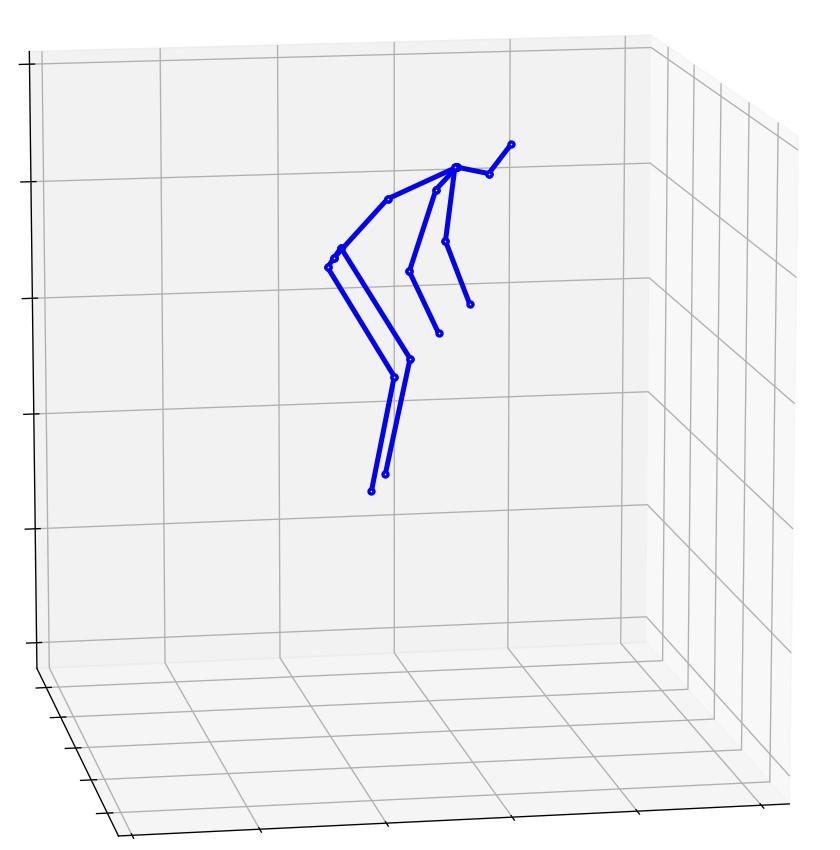}
    \end{subfigure}%
    \begin{subfigure}{0.23\textwidth}
        \centering
        \includegraphics[width=0.9\linewidth]{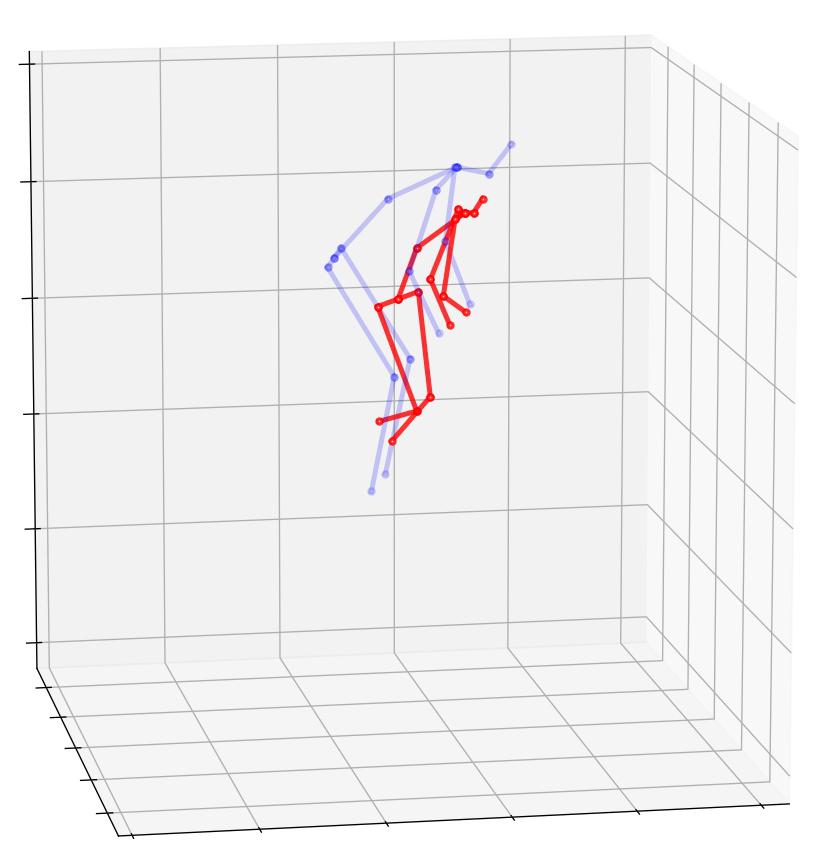}
    \end{subfigure}%
    \begin{subfigure}{0.23\textwidth}
        \centering
        \includegraphics[width=0.9\linewidth]{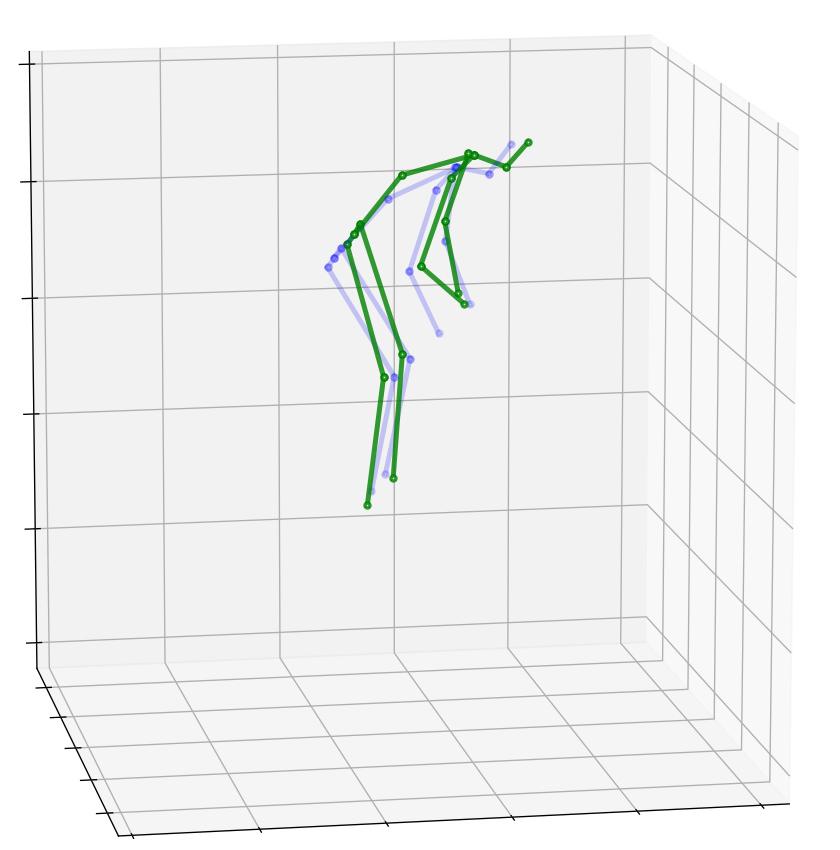}
    \end{subfigure}
    \\
    \begin{subfigure}{0.23\textwidth}
        \centering
        \includegraphics[width=0.9\linewidth]{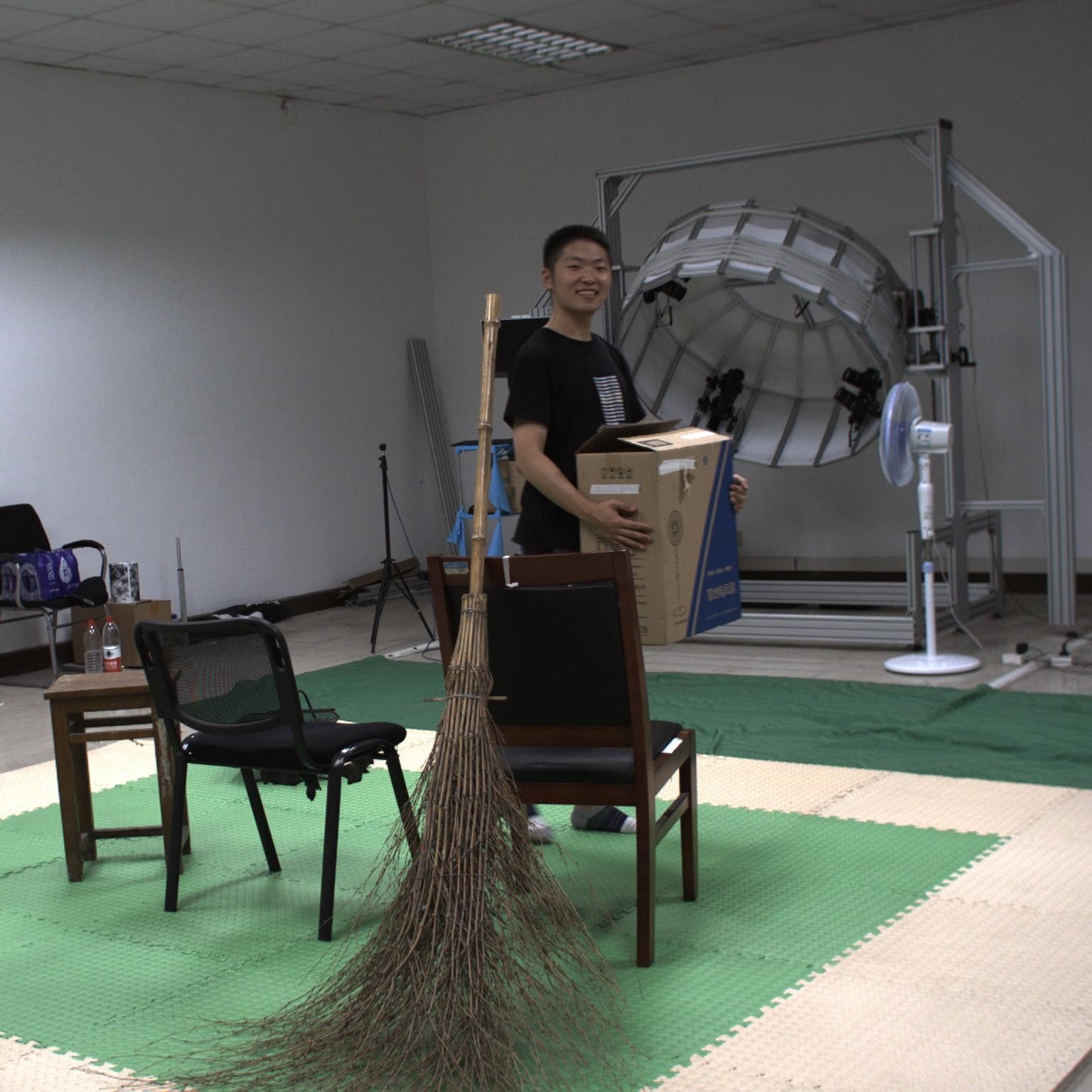}
        \caption{OCMotion Image}

    \end{subfigure}%
    \begin{subfigure}{0.23\textwidth}
        \centering
        \includegraphics[width=0.9\linewidth]{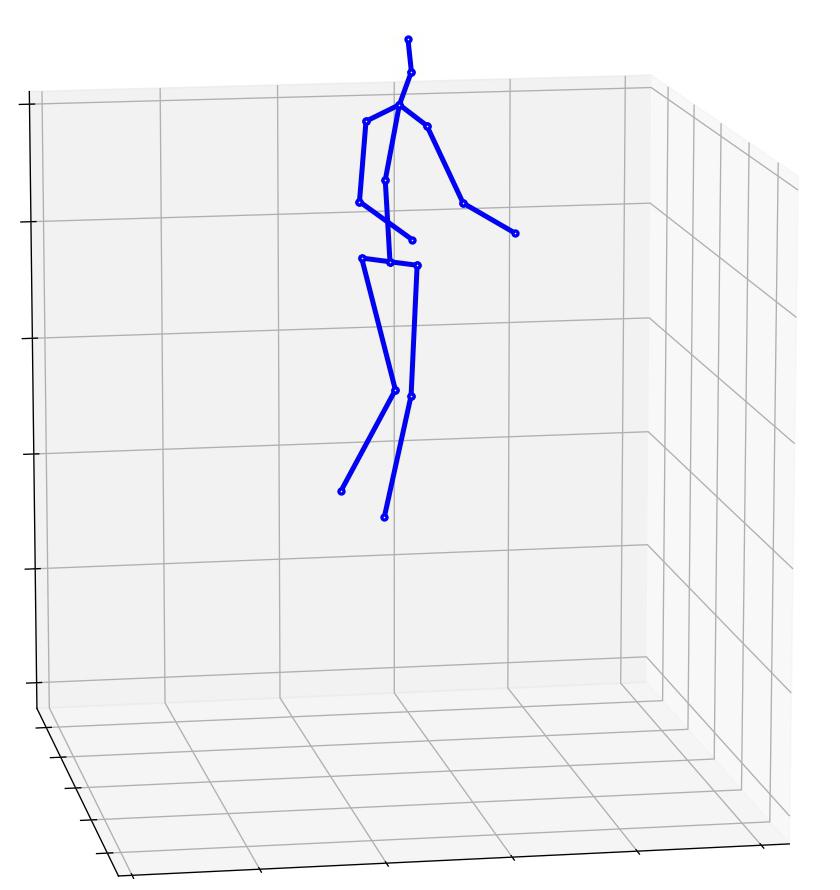}
        \caption{Ground Truth}

    \end{subfigure}%
    \begin{subfigure}{0.23\textwidth}
        \includegraphics[width=0.9\linewidth]{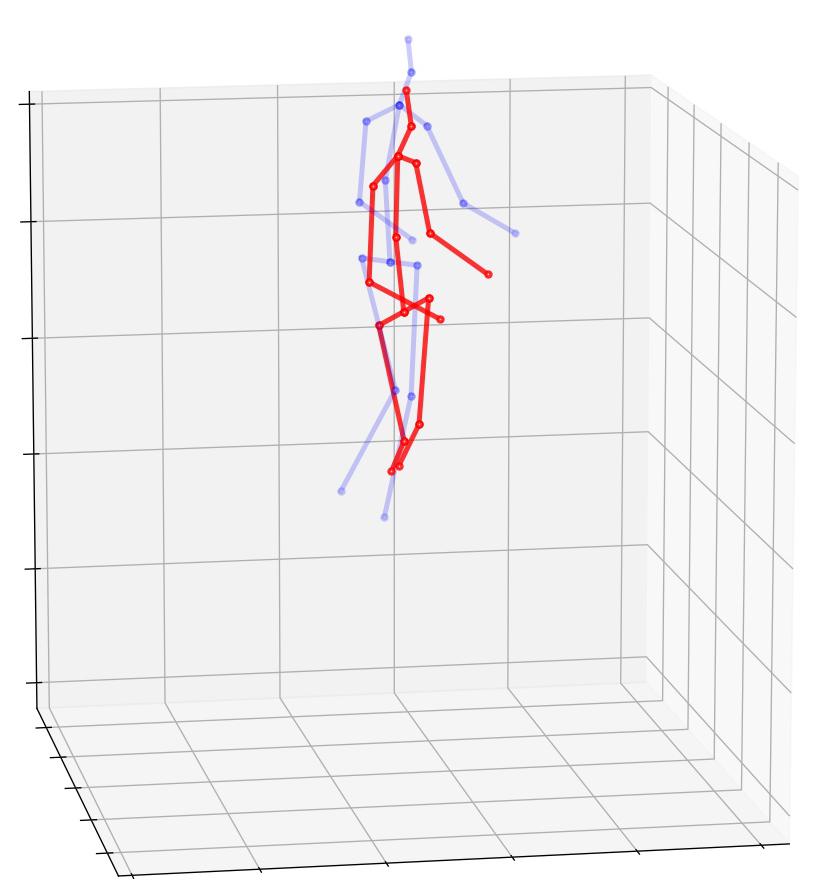}
        \caption{PoseFormerV2}

    \end{subfigure}%
    \begin{subfigure}{0.23\textwidth}
        \includegraphics[width=0.9\linewidth]{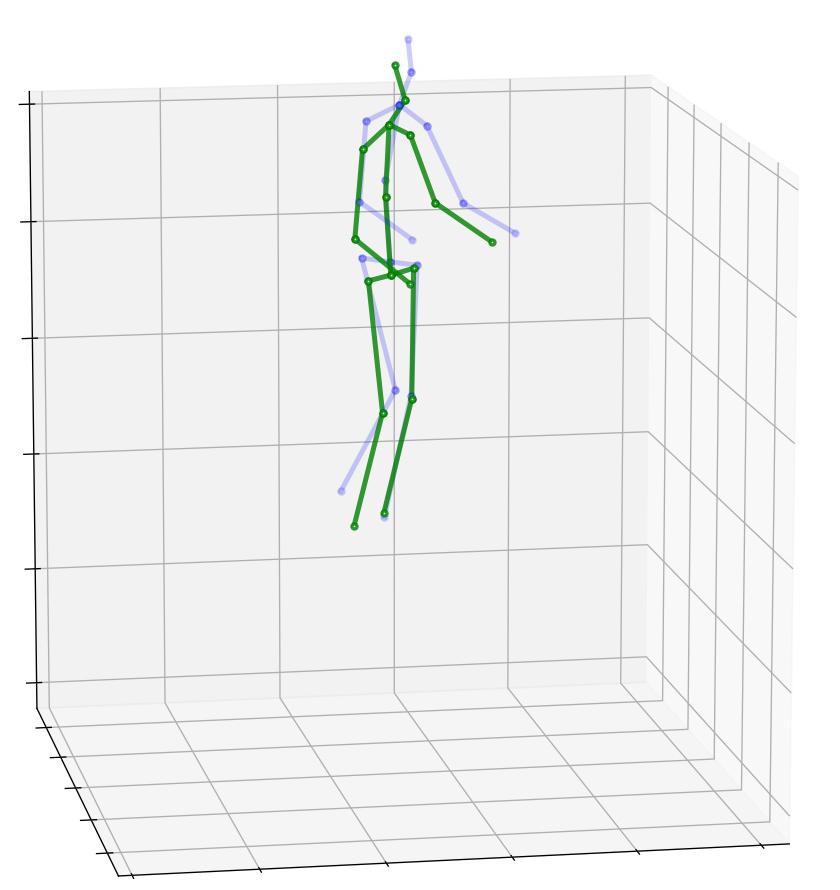}
        \caption{\name~(ours)} 
    \end{subfigure}
    \caption{This figure shows how our method works when tested in natural occlusion cases. The translucent blue color in the \textit{second column}, \textit{third column}, and \textit{fourth column} represents the ground truth. Blue, red, and green similarly represent Ground Truth, PoseformerV2 and \name~results, respectively.}
    \label{fig:OCMotion_qualitative_res}
\end{figure*}

\begin{figure*}[]
    \centering
    \begin{subfigure}{0.11\linewidth}
    \centering
        \includegraphics[width=\linewidth]{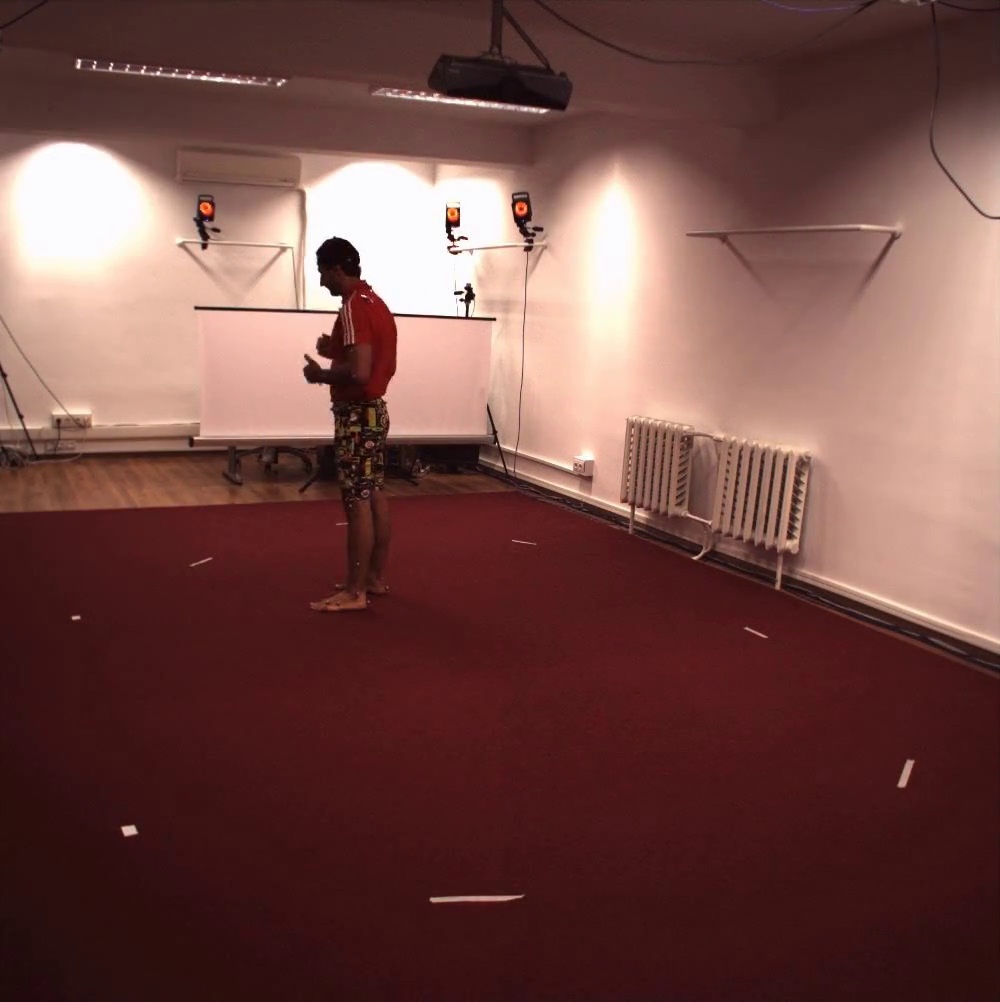}
    \end{subfigure}\hfill%
    \begin{subfigure}{0.11\linewidth}
    \centering
        \includegraphics[width=\linewidth]{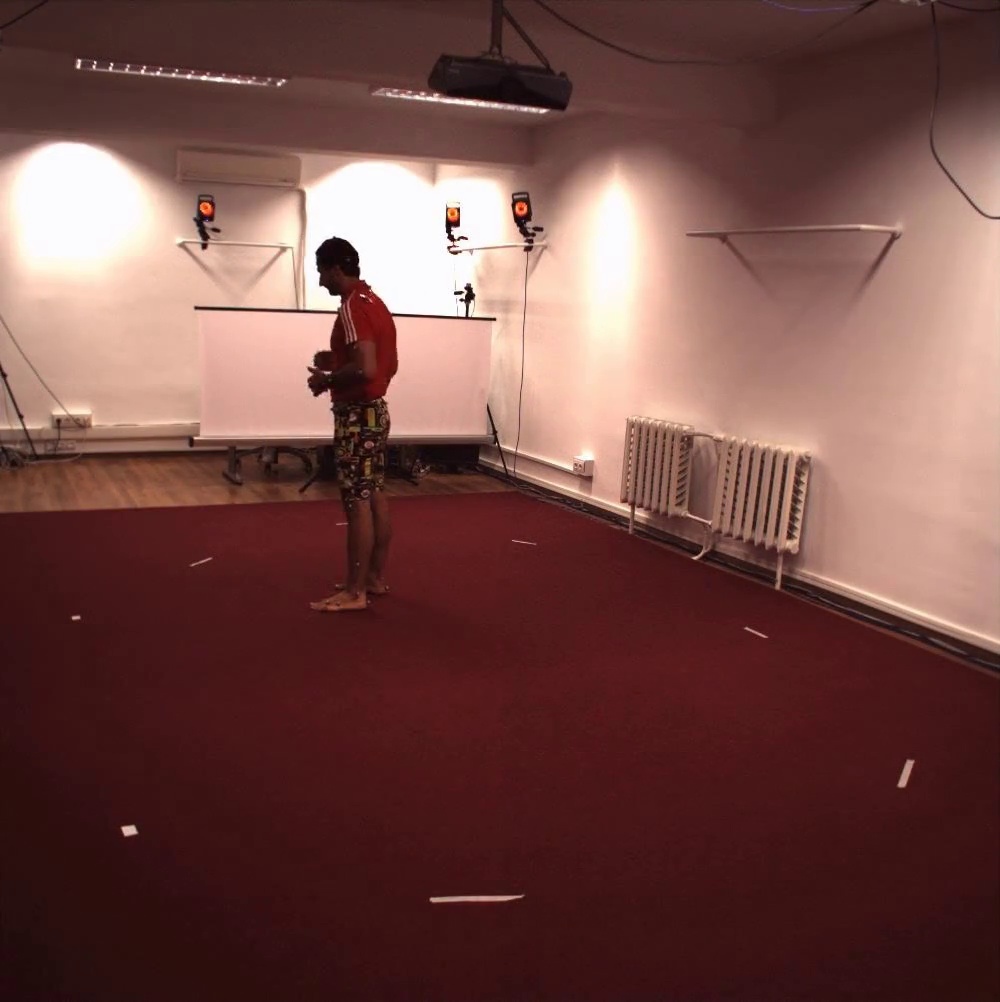}
    \end{subfigure}\hfill%
    \begin{subfigure}{0.11\linewidth}
    \centering
        \includegraphics[width=\linewidth]{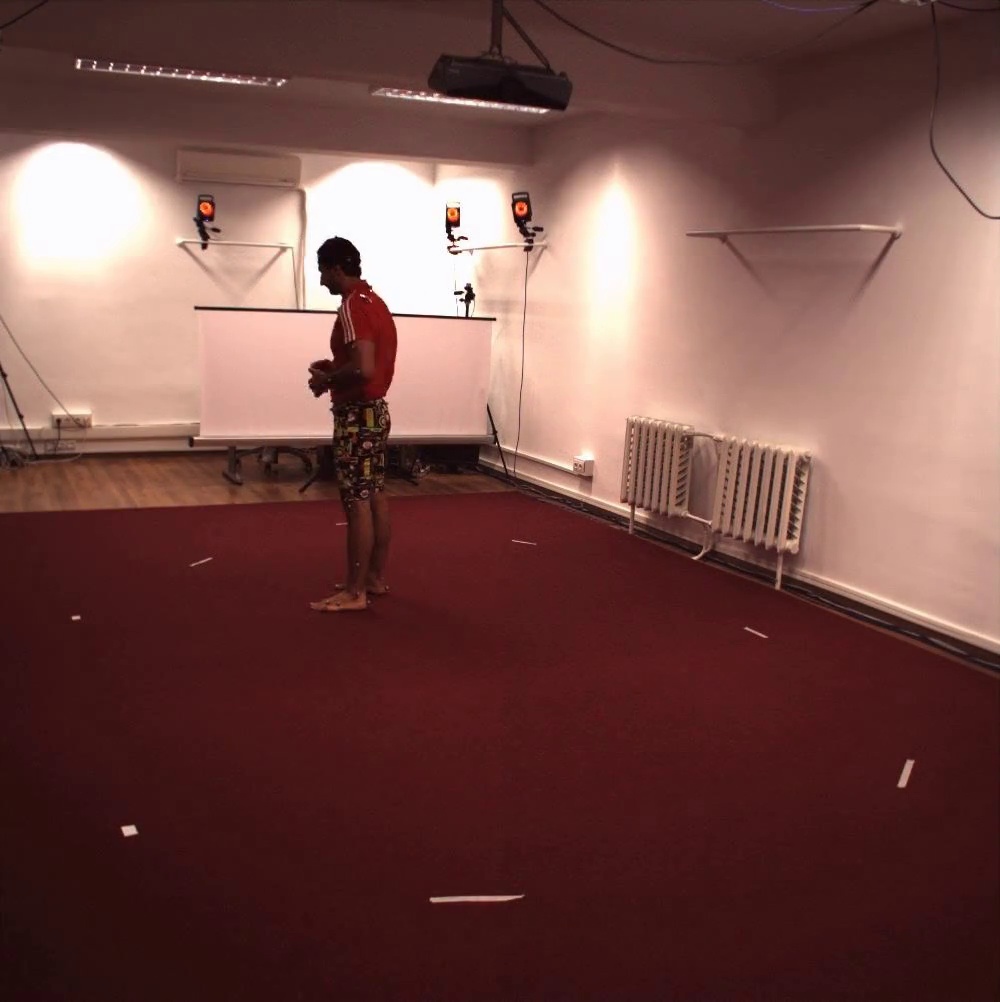}
    \end{subfigure}\hfill%
    \begin{subfigure}{0.11\linewidth}
     \centering
       \includegraphics[width=\linewidth]{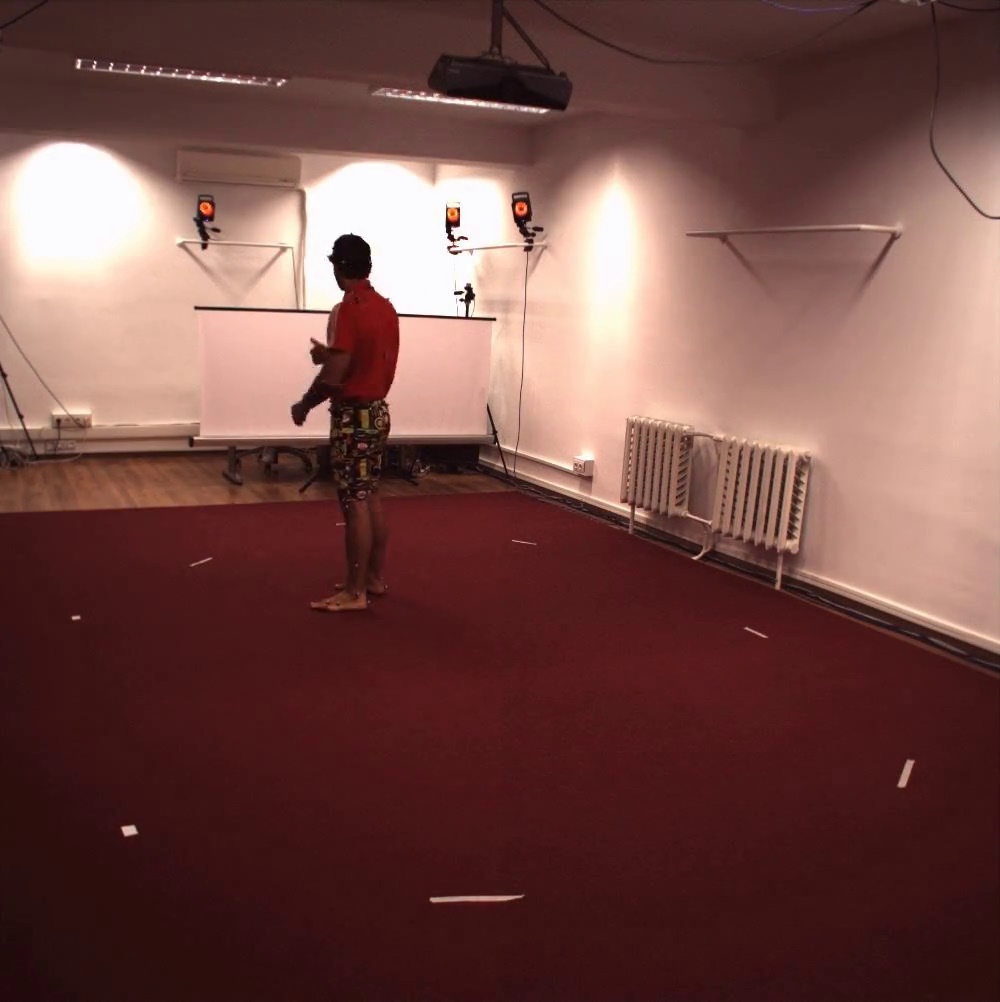}
    \end{subfigure}\hfill%
    \begin{subfigure}{0.11\linewidth}
    \centering
        \includegraphics[width=\linewidth]{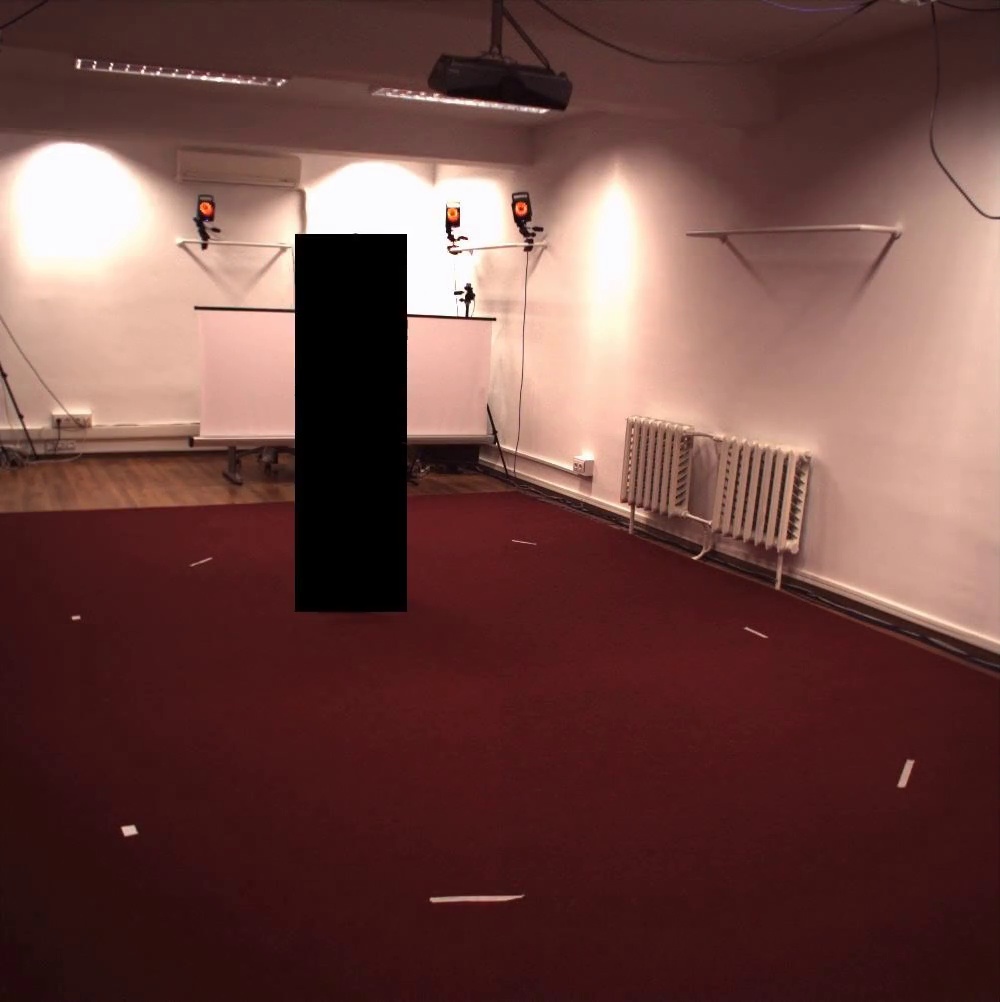}
    \end{subfigure}\hfill%
    \begin{subfigure}{0.11\linewidth}
\centering
    \includegraphics[width=\linewidth]{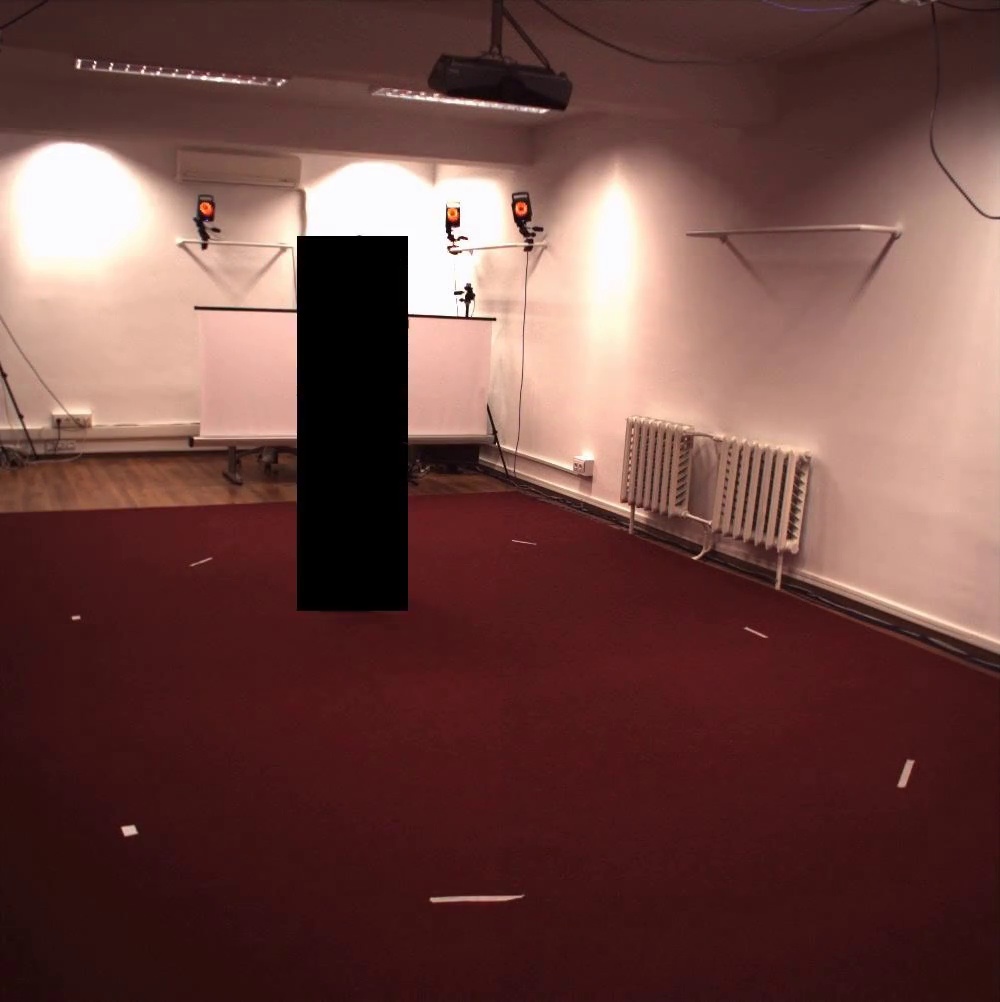}
\end{subfigure}\hfill%
    \begin{subfigure}{0.11\linewidth}
    \centering
        \includegraphics[width=\linewidth]{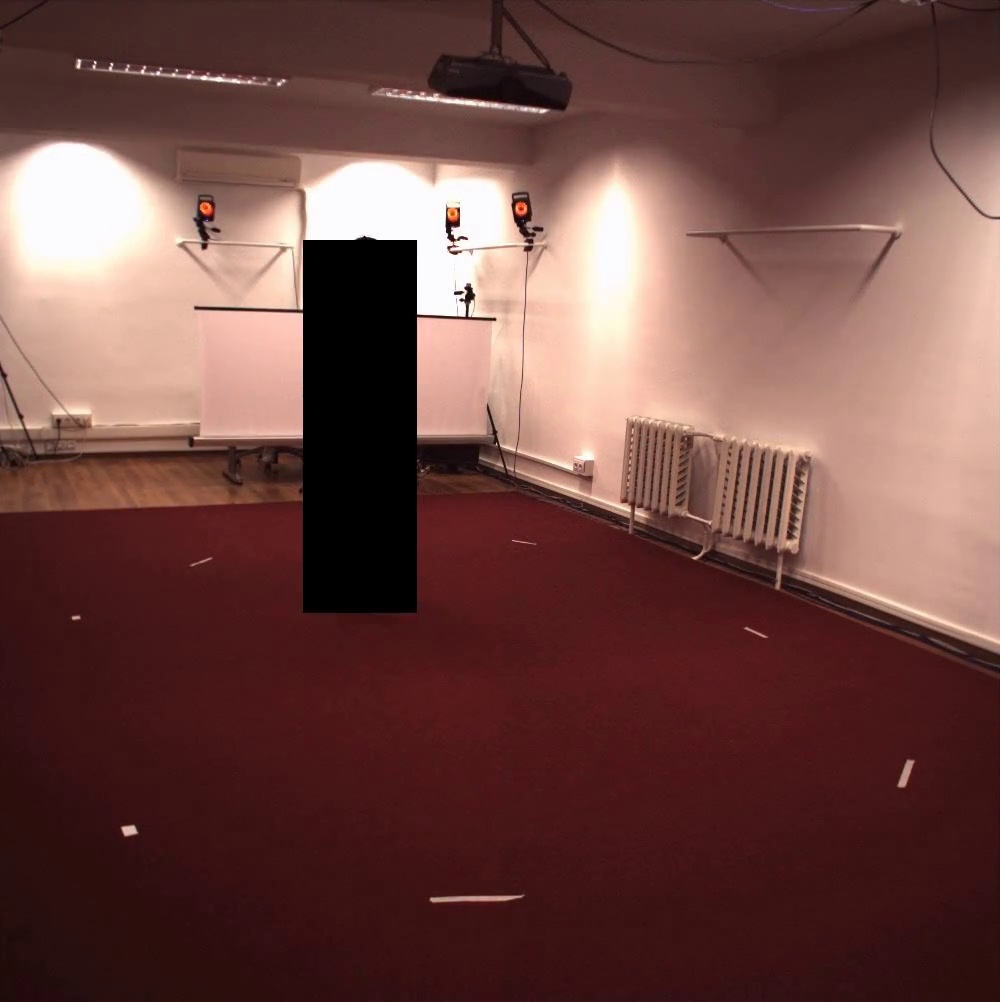}
    \end{subfigure}\hfill%
    \begin{subfigure}{0.11\linewidth}
    \centering
        \includegraphics[width=\linewidth]{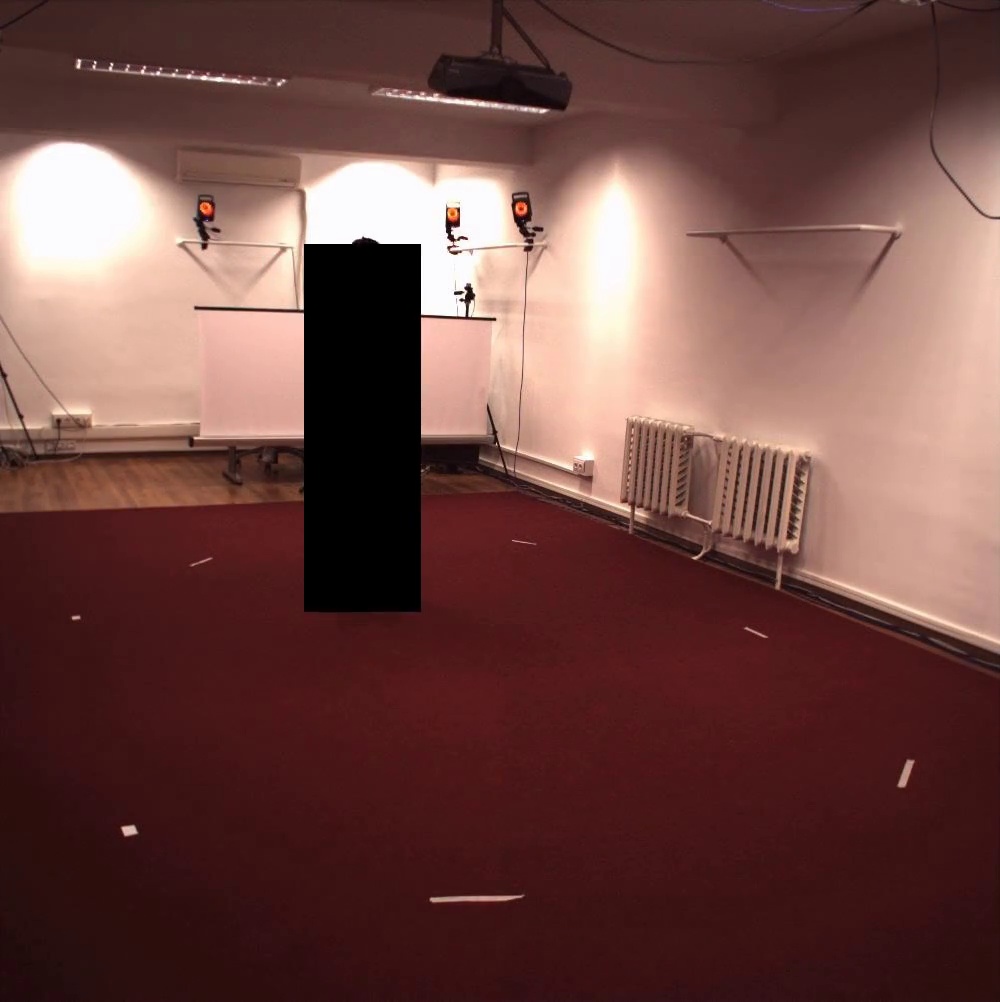}
    \end{subfigure}\hfill%
    \begin{subfigure}{0.11\linewidth}
        \centering
        \includegraphics[width=\linewidth]{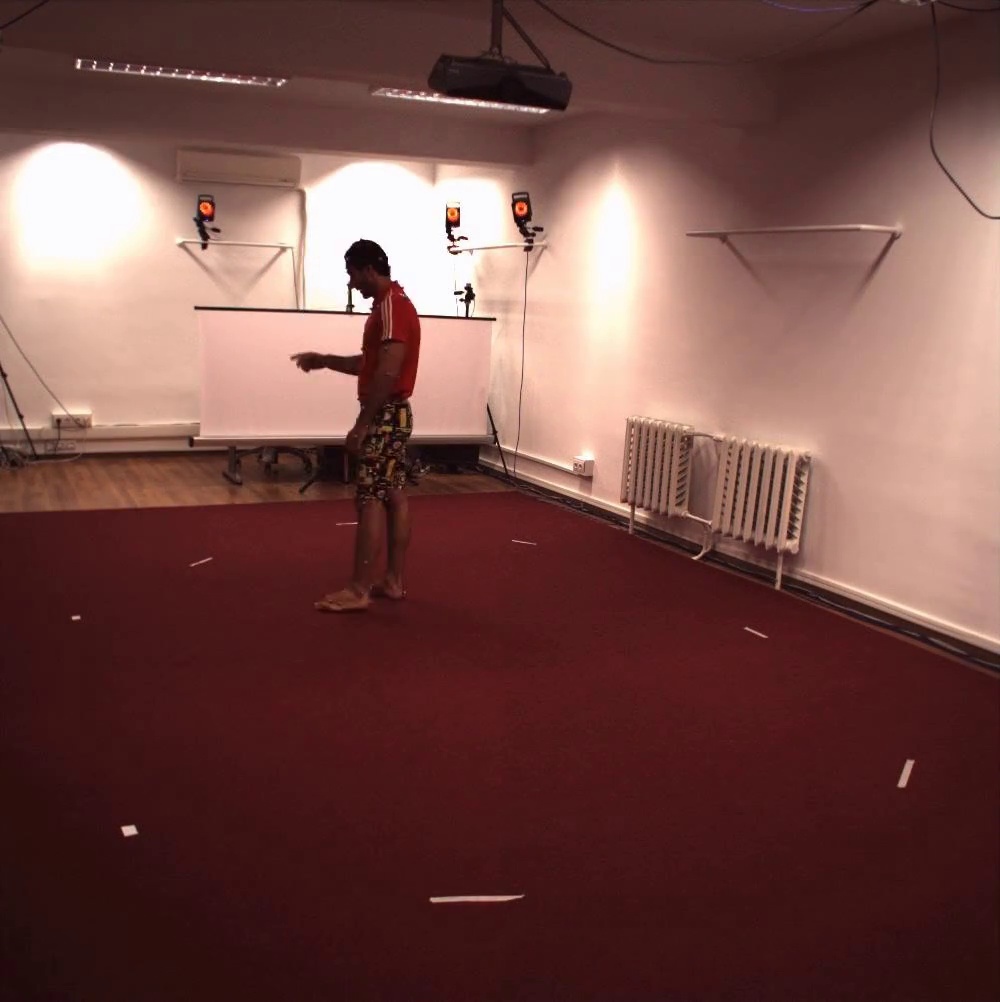}
    \end{subfigure}
    \\
    \begin{subfigure}{0.11\linewidth}
    \centering
        \includegraphics[width=\linewidth]{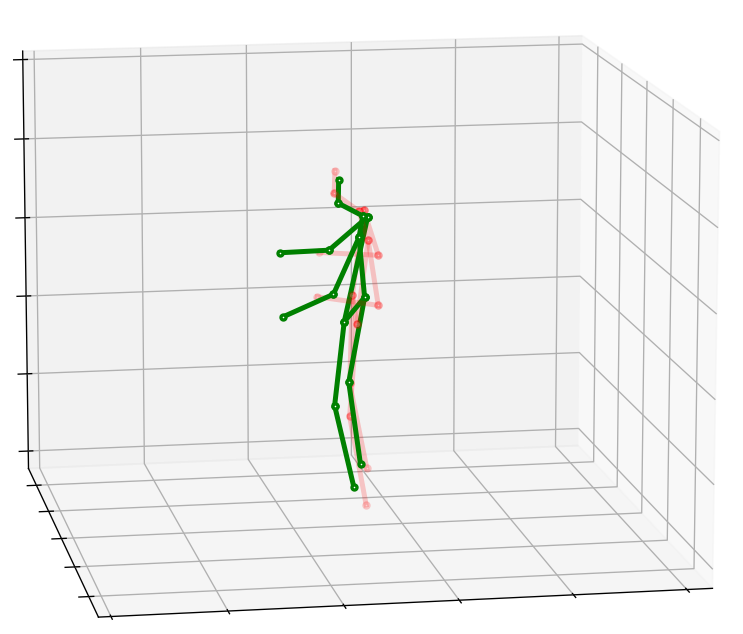}
    \end{subfigure}\hfill%
    \begin{subfigure}{0.11\linewidth}
    \centering
        \includegraphics[width=\linewidth]{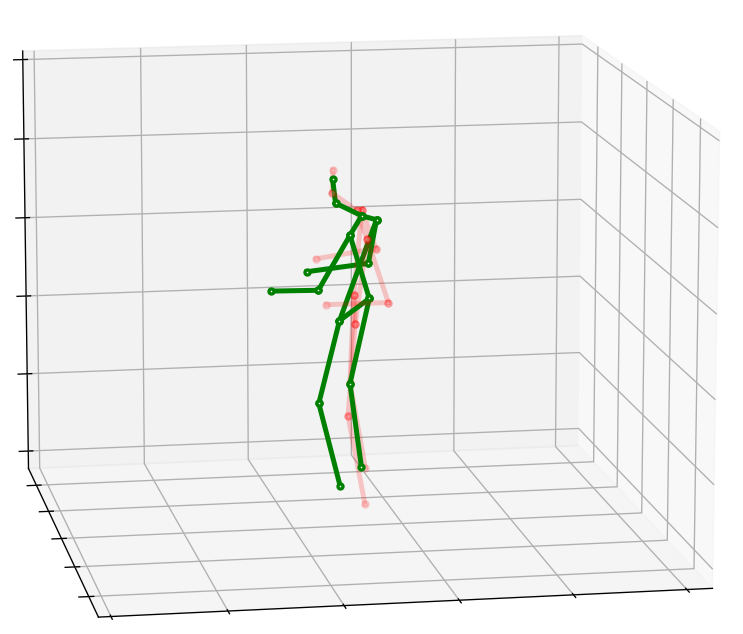}
    \end{subfigure}\hfill%
    \begin{subfigure}{0.11\linewidth}
    \centering
        \includegraphics[width=\linewidth]{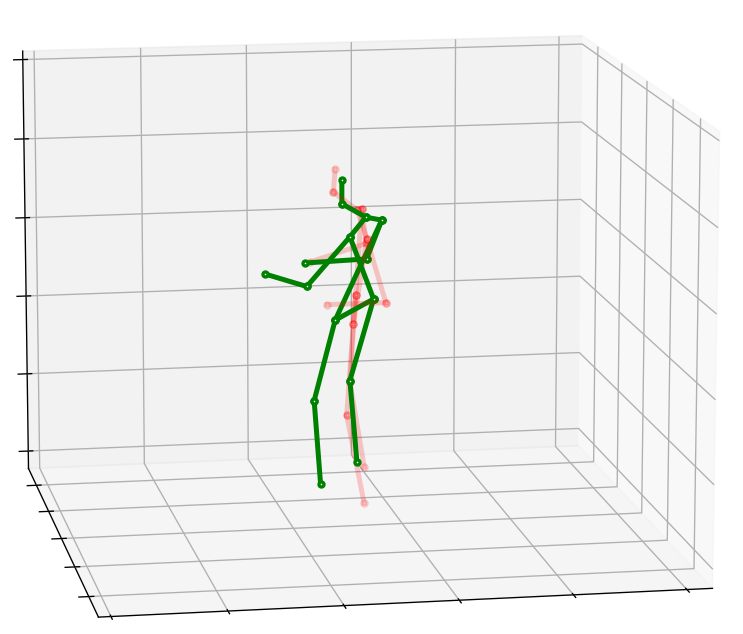}
    \end{subfigure}\hfill%
    \begin{subfigure}{0.11\linewidth}
     \centering
       \includegraphics[width=\linewidth]{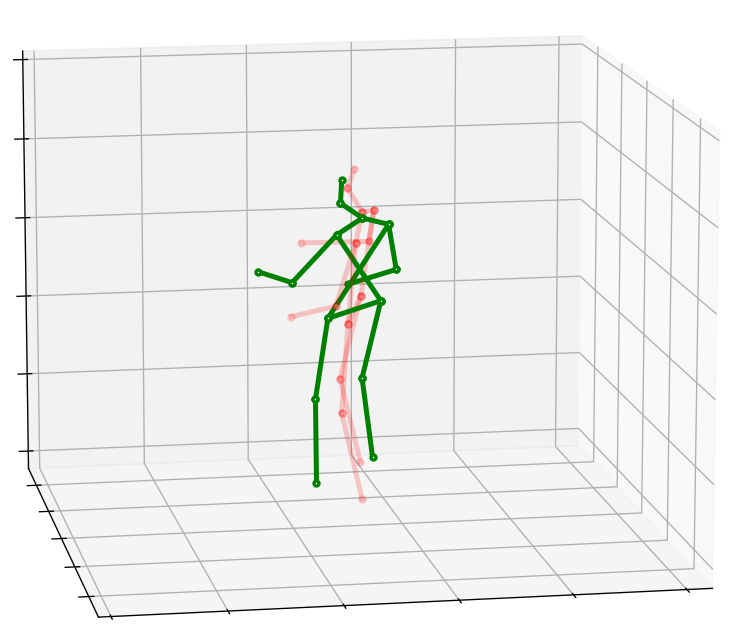}
    \end{subfigure}\hfill%
    \begin{subfigure}{0.11\linewidth}
    \centering
        \includegraphics[width=\linewidth]{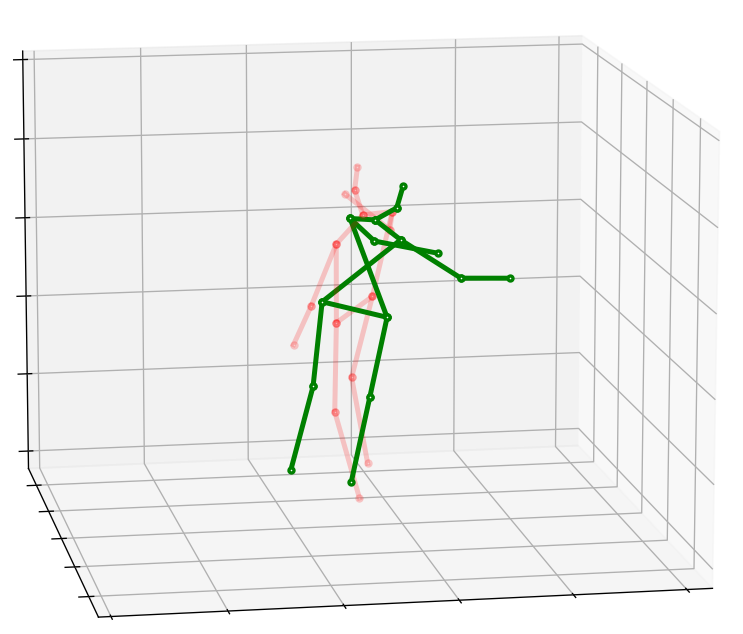}
    \end{subfigure}\hfill%
    \begin{subfigure}{0.11\linewidth}
\centering
    \includegraphics[width=\linewidth]{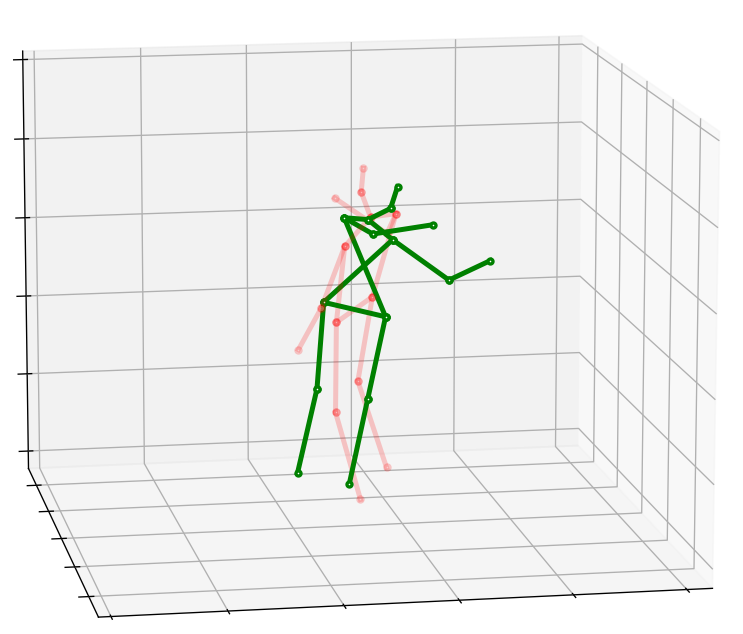}
\end{subfigure}\hfill%
    \begin{subfigure}{0.11\linewidth}
    \centering
        \includegraphics[width=\linewidth]{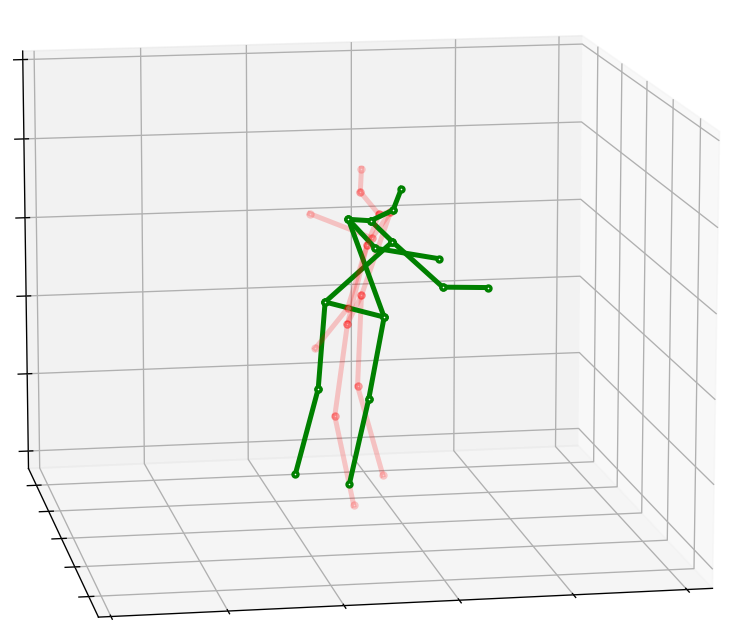}
    \end{subfigure}\hfill%
    \begin{subfigure}{0.11\linewidth}
    \centering
        \includegraphics[width=\linewidth]{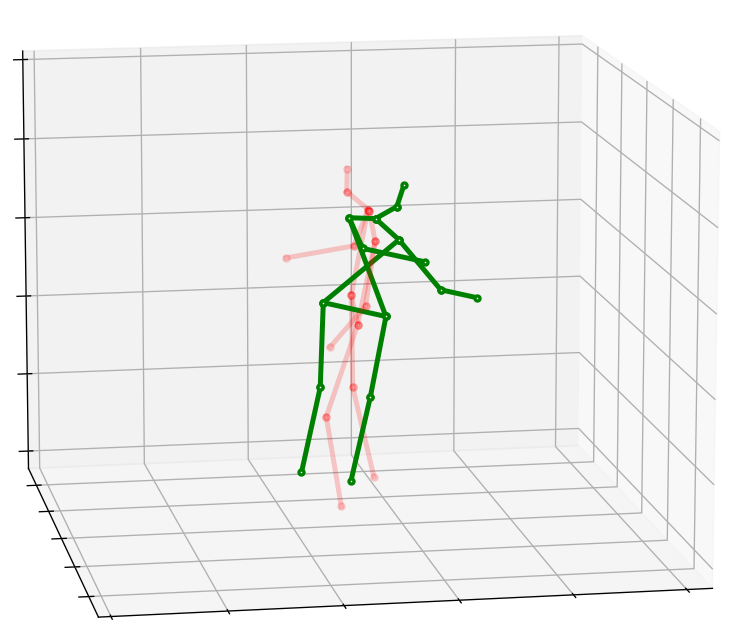}
    \end{subfigure}\hfill%
    \begin{subfigure}{0.11\linewidth}
        \centering
        \includegraphics[width=\linewidth]{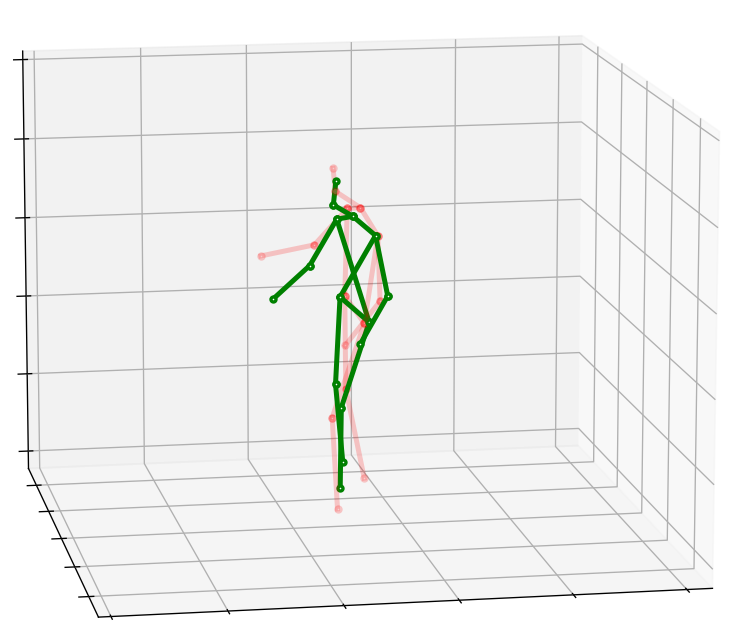}
    \end{subfigure}
    \\
    \begin{subfigure}{0.11\linewidth}
    \centering
        \includegraphics[width=\linewidth]{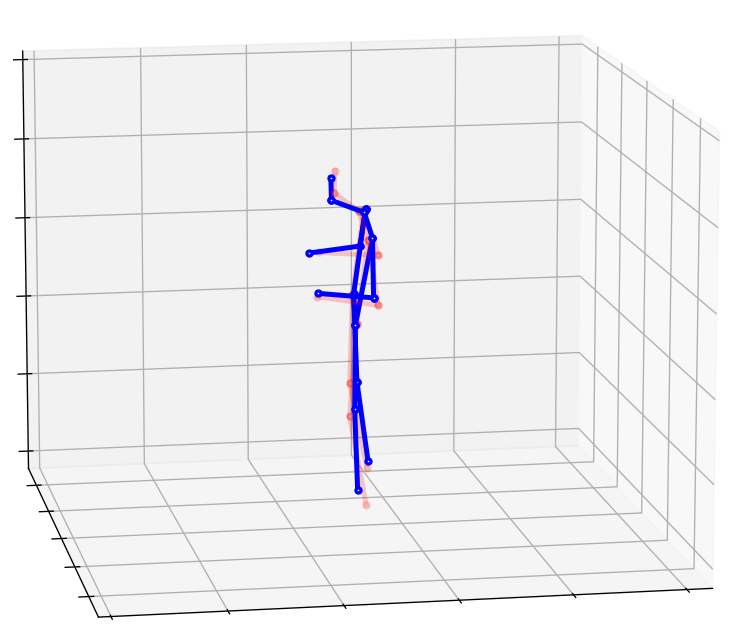}
    \end{subfigure}\hfill%
    \begin{subfigure}{0.11\linewidth}
    \centering
        \includegraphics[width=\linewidth]{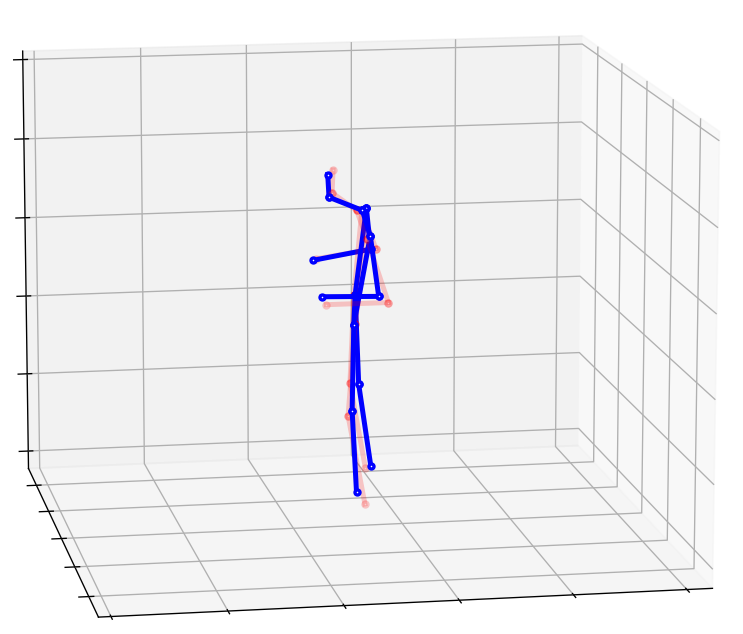}
    \end{subfigure}\hfill%
    \begin{subfigure}{0.11\linewidth}
    \centering
        \includegraphics[width=\linewidth]{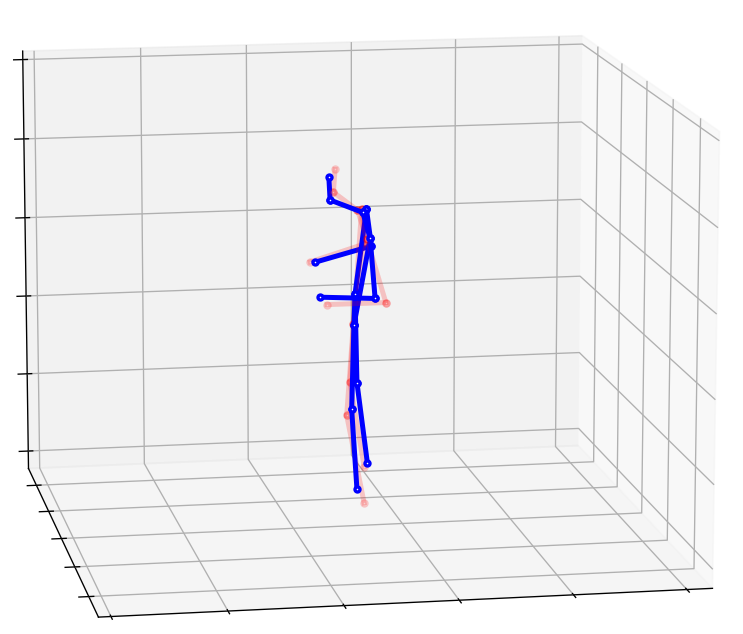}
    \end{subfigure}\hfill%
    \begin{subfigure}{0.11\linewidth}
     \centering
       \includegraphics[width=\linewidth]{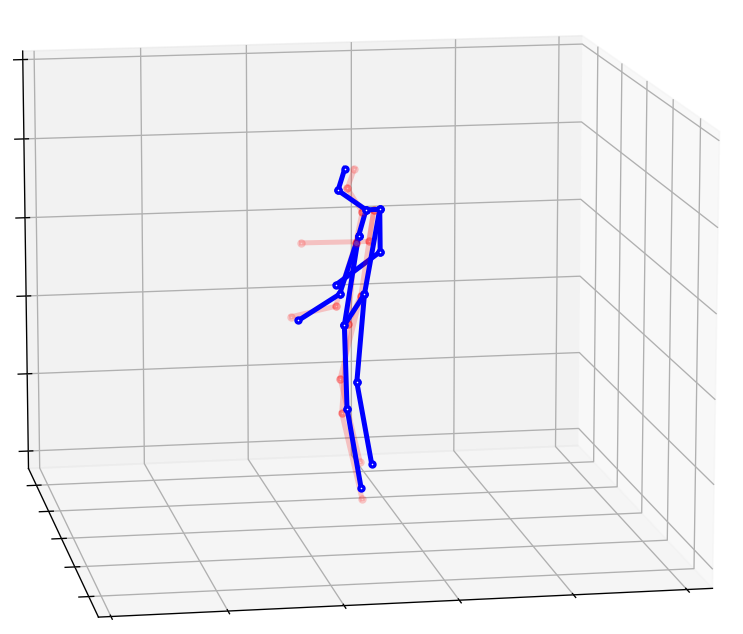}
    \end{subfigure}\hfill%
    \begin{subfigure}{0.11\linewidth}
    \centering
        \includegraphics[width=\linewidth]{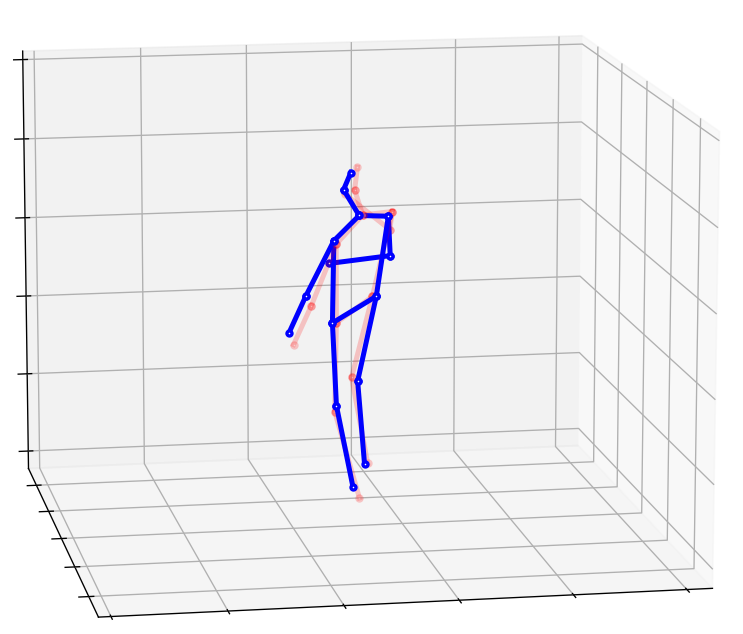}
    \end{subfigure}\hfill%
    \begin{subfigure}{0.11\linewidth}
\centering
    \includegraphics[width=\linewidth]{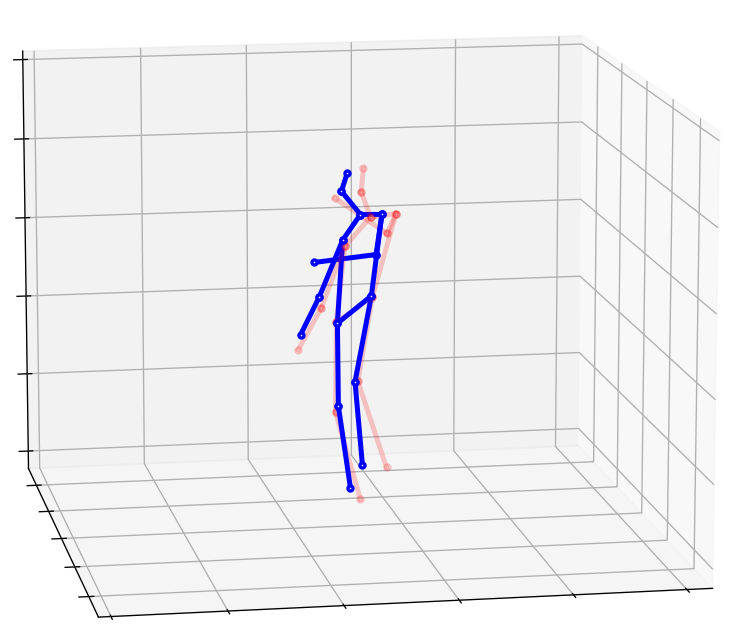}
\end{subfigure}\hfill%
    \begin{subfigure}{0.11\linewidth}
    \centering
        \includegraphics[width=\linewidth]{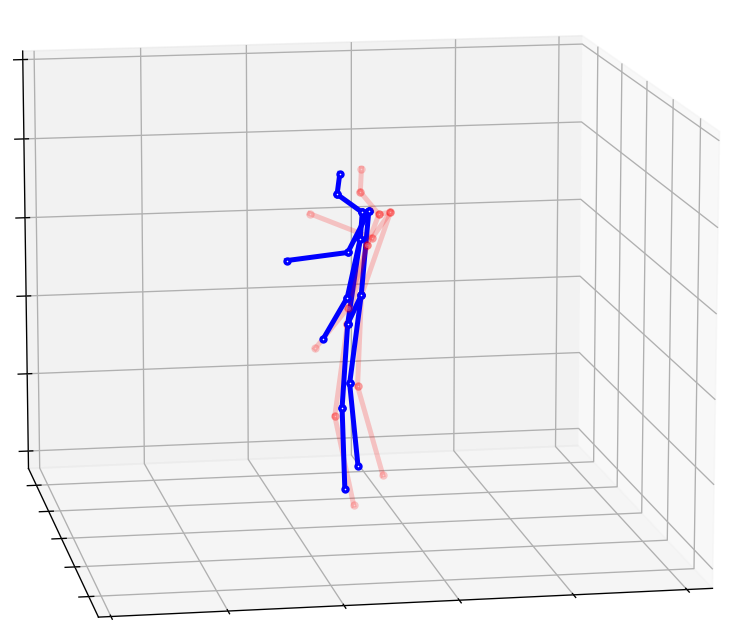}
    \end{subfigure}\hfill%
    \begin{subfigure}{0.11\linewidth}
    \centering
        \includegraphics[width=\linewidth]{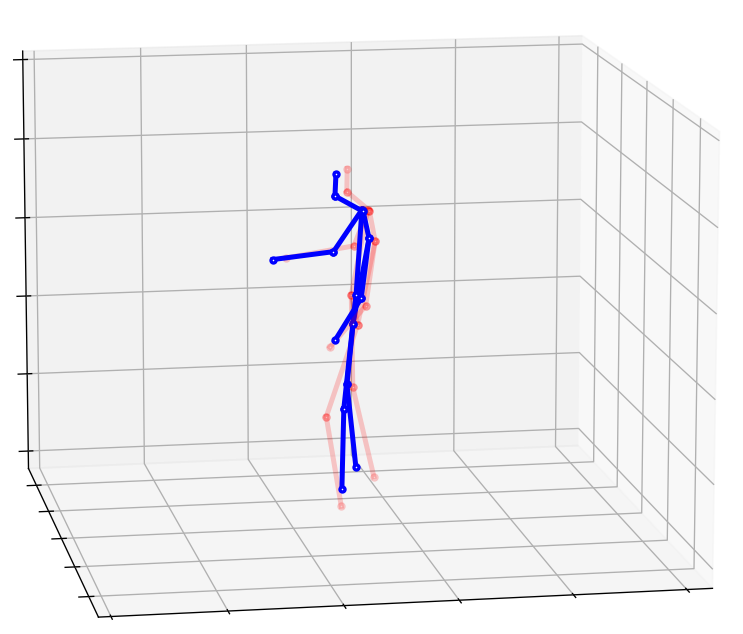}
    \end{subfigure}\hfill%
    \begin{subfigure}{0.11\linewidth}
        \centering
        \includegraphics[width=\linewidth]{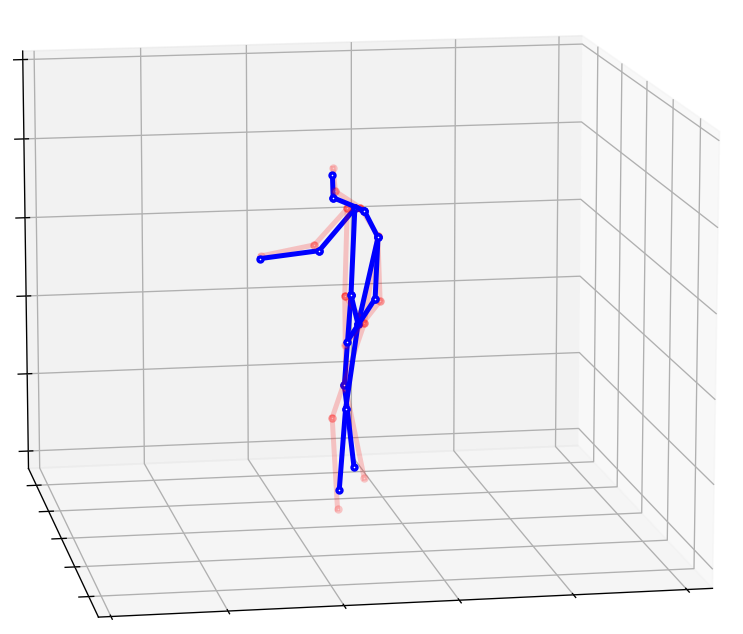}
    \end{subfigure}
    \caption{\textbf{3D pose estimation results on Occluded Human3.6M}. CycleAdapt (\textit{second} row) fails to generalize in cases when there is complete occlusion. \name~ (\textit{third} row) produces temporally coherent pose infilling due to test time training. Note that the translucent red color represents the ground truth poses.
      }
    \label{fig:OccH36M_qualitative_res}
\end{figure*}

\section{Additional Qualitative Results}
In Fig.~\ref{fig:OCMotion_qualitative_res} we compare our method against a different state-of-the-art 3D pose estimation method named PoseFormerV2~\cite{zhao2023poseformerv2}. We observe that \name's skeleton is best aligned with the actual ground truth pose, even when there is significant occlusion.

One trivial way to improve the results of BEDLAM is by linear interpolation between frames. However, we qualitatively found that the interpolation was very smooth and misses to capture the intricate motion. Our loss optimization during inference helps to achieve the best results. Interpolation results are shown in Table \ref{tab:occludedH36M_supp}. 

\section{Additional Quantitative Results}
\vspace{-2mm}
In our study, we conduct a comparison with the WHAM model \cite{shin2023wham}, a method that has only recently been introduced. WHAM employs a fully supervised training approach, benefiting from extensive datasets for its development. Notably, our method surpasses WHAM's performance on the Occluded Human3.6M dataset (refer to Table~\ref{tab:occludedH36M_supp}), demonstrating \name's enhanced capability in managing significant occlusions. Additionally, we achieve results comparable to those of WHAM on the OCMotion dataset (refer to Table~\ref{tab:OCMotion_supp}). It is important to highlight that the OCMotion dataset lacks substantial occlusions, which limits the opportunity to showcase \name's strengths fully. \name~is particularly effective in scenarios with heavy occlusion, as evidenced by its performance on Occluded Human3.6M. Further, we also compare with 3DNBF \cite{zhang2023nbf}, which was specially introduced to tackle occlusions while estimating the pose. It is evident from the results that \name~surpasses 3DNBF's performance (refer to Table~\ref{tab:occludedH36M_supp}) and also highlights the point that \name~performs the best when there are severe occlusions in the scene.

\name~is originally proposed to improve the temporal continuity of any existing pose estimation method. This makes \name~agnostic to any existing pose estimation method. Note that in Table~\ref{tab:clf} even if we use CLIFF, \name~outperforms existing SOTA methods on Occluded Human3.6.


\begin{table}[!htp]
\centering
\setlength{\tabcolsep}{3.5pt}
\def\arraystretch{1.35}  
\resizebox{0.8\linewidth}{!}{
\begin{tabular}{ccccc}
\toprule
& Method & PA-MPJPE & Accel & Avg \\ \hline
\multirow{3}{*}{\rotatebox{90}{Image}}& 
OOH~\cite{zhang2020object} & 55.0 & 48.6 & 51.8\\
& PARE~\cite{kocabas2021pare} & 52.0 & 43.6 & 47.8\\ 
& BEDLAM~\cite{black2023bedlam} & 47.1 & 49.0 & 48.0\\
\midrule   
\multirow{5}{*}{\rotatebox{90}{Video}} 
& PoseFormerV2~\cite{zhao2023poseformerv2} & 126.3 & 28.5 & 77.4 \\
& GLAMR~\cite{yuan2022glamr} & 89.9 & 51.3 & 70.6 \\
& CycleAdapt~\cite{cycleadapt} & 74.6 & 57.5 & 66.0 \\
& ROMP~\cite{sun2021monocular} & 48.1 & 57.2 & 52.6 \\
& SPIN$^\dagger$~\cite{kolotouros2019learning} & 56.7 & 47.0 & 51.8\\
& VIBE$^\dagger$~\cite{kocabas2020vibe} & 58.6 & 44.5 & 51.5 \\
& WHAM~\cite{shin2023wham} & \textbf{42.4} & \textbf{27.0} & \textbf{34.7} \\
\midrule
& {\name~(ours)} & 46.2 & 47.8 & 47.0 \\
\bottomrule
\end{tabular}}
\caption{\textbf{3D pose estimation results on OCMotion \cite{huang2022object}}. WHAM performs better than other methods because it is a supervised method and has been trained on large amounts of data compared to \name's backbone. Hence, it is able to generalize well on the OCMotion dataset.}
\label{tab:OCMotion_supp}
\end{table}
\begin{table}[ht]
\centering
\setlength{\tabcolsep}{3.5pt}
\def\arraystretch{1.35}  
\resizebox{0.8\linewidth}{!}{
\begin{tabular}{ccccc}
\toprule
& Method & PA-MPJPE & MPJPE & Accel \\
\midrule
\multirow{4}{*}{\rotatebox{90}{Image}} 
& CLIFF~\cite{li2022cliff}  & 183.5 &  100.5 & 38.4\\
& BEDLAM~\cite{black2023bedlam} & 179.5  & 98.9  & 39.1 \\
& BEDLAM Interpolation~\cite{black2023bedlam} &  64.1 & 83.3 & -\\
& 3DNBF~\cite{zhang2023nbf} & 204.3 & 260.4 & 39.3\\
\midrule
\multirow{4}{*}{\rotatebox{90}{Video}} 
& GLAMR~\cite{yuan2022glamr} & 213.9 & 380.3 & 42.3 \\
& PoseFormerV2~\cite{zhao2023poseformerv2} & 193.9 & 260.2  & 38.7\\
& CycleAdapt~\cite{cycleadapt} & 77.6 & 132.6 & 48.7 \\
& MotionBERT~\cite{zhu2023motionbert} & 76.1 & 112.8 & 28.7 \\
& WHAM~\cite{shin2023wham} & 119.5 & 237.7 & 46.8 \\
\midrule
& {\name~(ours)} & \textbf{59.0} & \textbf{80.7} & \textbf{26.6} \\
\bottomrule
\end{tabular}}
\caption{3D Pose estimation results on \textbf{Occluded Human3.6M}. This dataset is crucial as it is the only dataset that has significant occlusion. The results underscore that \name~ surpasses all state-of-the-art including WHAM with substantial percentage improvements, affirming its robustness in handling occlusions.}
\label{tab:occludedH36M_supp}
\end{table}

\begin{table}[!htp]
    \centering
    \resizebox{\linewidth}{!}{
    \setlength{\tabcolsep}{5pt}
    \begin{tabular}{ccccccc}
        \toprule
        & \multicolumn{2}{c}{Occluded H36M}  & \multicolumn{2}{c}{Human3.6M}  & \multicolumn{2}{c}{OCMotion}       \\
        \cmidrule{2-7}
         Method & PA-MPJPE & MPJPE & PA-MPJPE & MPJPE & PA-MPJPE & Accel  \\
         \midrule
         CLIFF & 183.5 & 100.5 &  \textbf{39.4} & \textbf{62.9}   & 54.2  & 56.8 \\
         \name~ (CLIFF) &  \textbf{64.8} & \textbf{82.1} & 40.1 & 63.2 & \textbf{52.1}  & \textbf{54.2}\\
         \midrule
         BEDLAM & 179.5 & 98.9 & 50.9 & 70.9 & 47.1 & 49.0 \\
         \name~ (BEDLAM) & \textbf{59.0} & \textbf{80.7} & \textbf{50.4} &  \textbf{69.7} & \textbf{46.2} & \textbf{47.8}\\
         \bottomrule
    \end{tabular}
    }
    \caption{Effect of various 3D Pose estimation method on \name. We observe that we gain similar performance improvement if we have use any other backbone as well.}
    \label{tab:clf} 
\end{table}

\begin{figure*}[!ht]
    \centering
    \begin{subfigure}{0.11\linewidth}
    \centering
        \includegraphics[width=\linewidth]{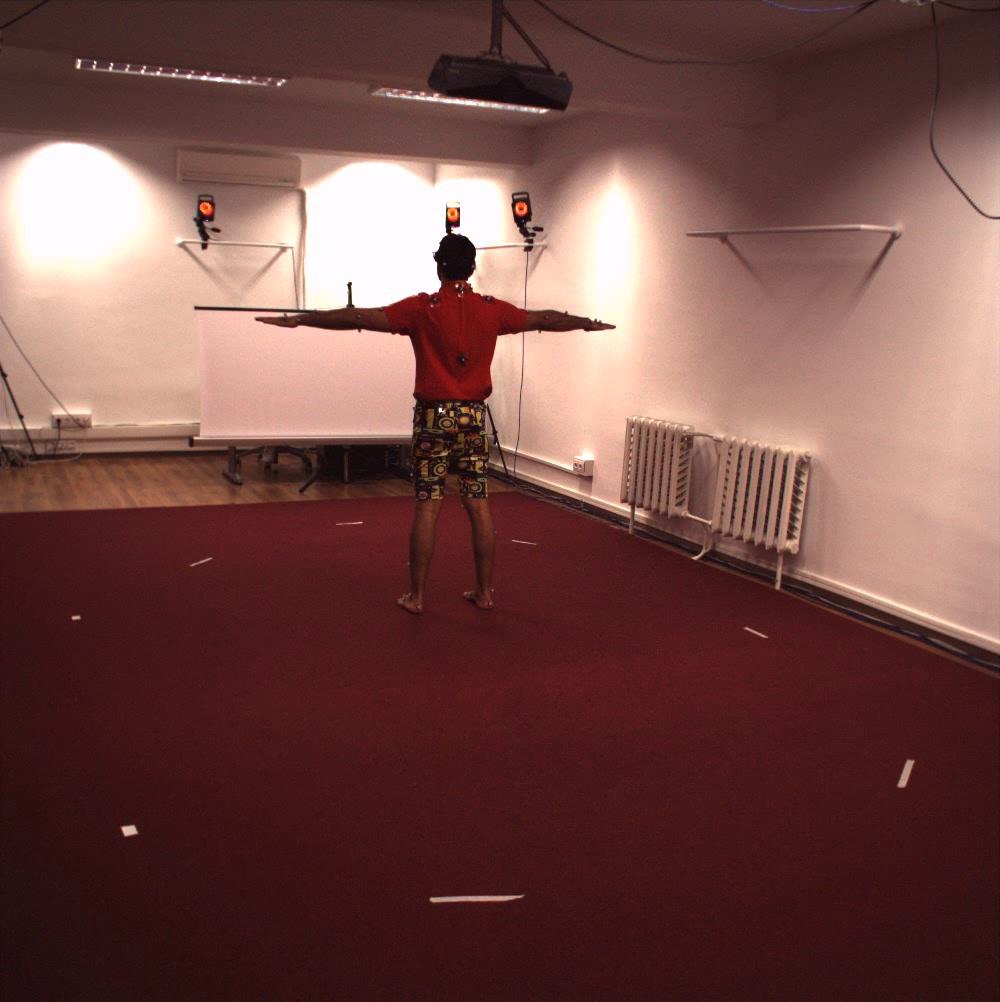}
        \caption*{Frame 1}
    \end{subfigure}\hfill%
    \begin{subfigure}{0.11\linewidth}
    \centering
        \includegraphics[width=\linewidth]{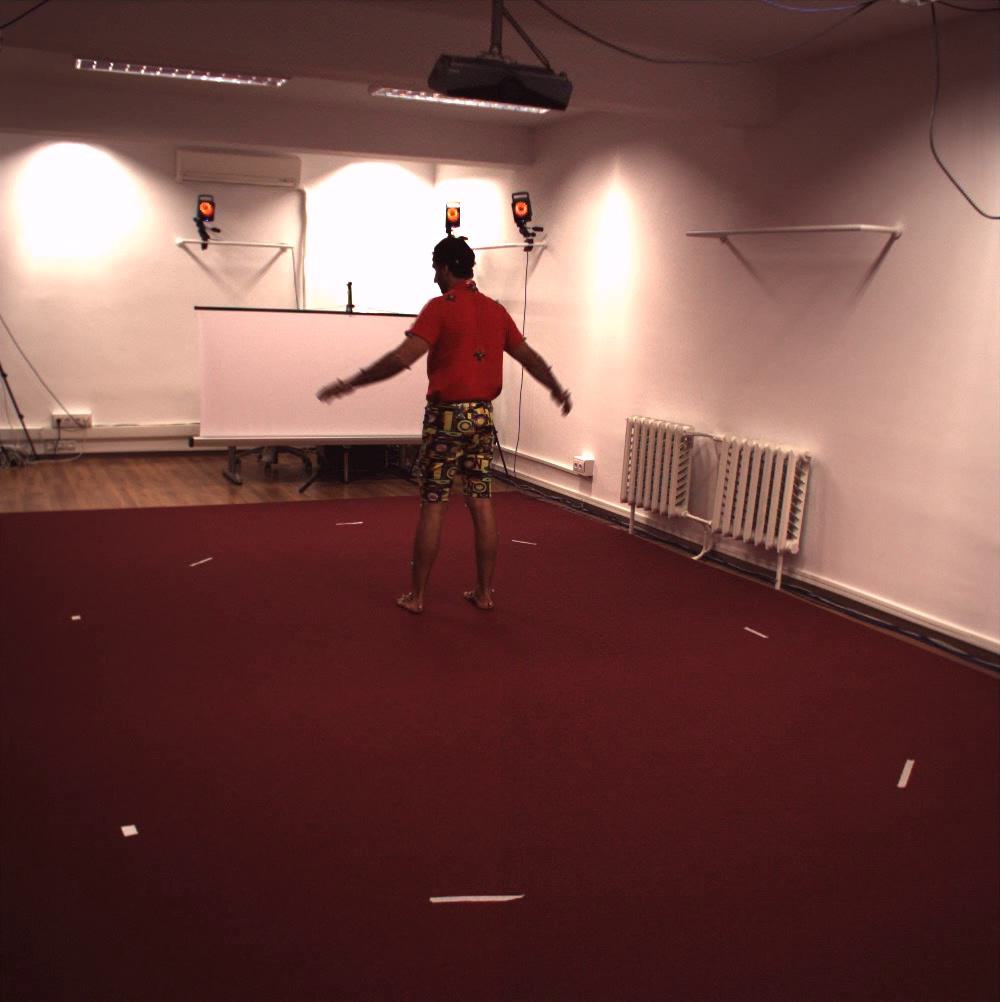}
        \caption*{Frame 6}
    \end{subfigure}\hfill%
    \begin{subfigure}{0.11\linewidth}
    \centering
        \includegraphics[width=\linewidth]{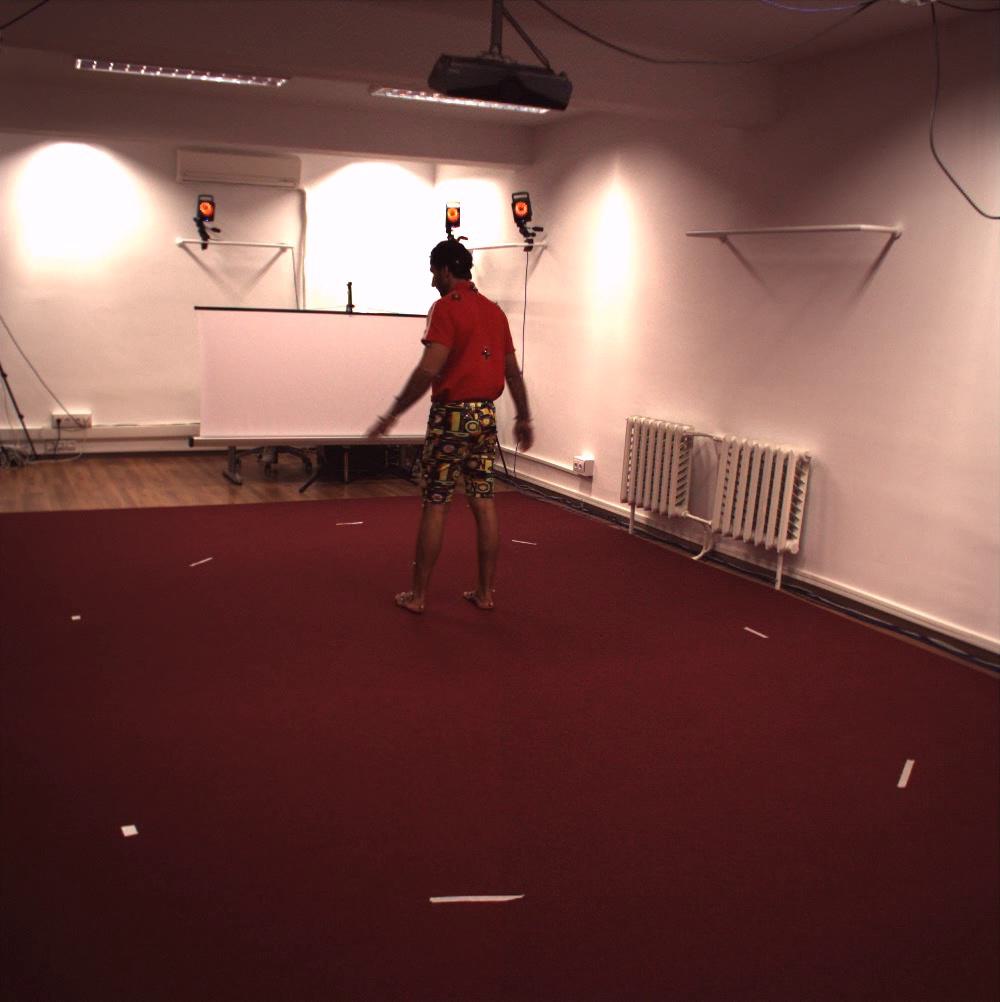}
        \caption*{Frame 9}
    \end{subfigure}\hfill%
    \begin{subfigure}{0.11\linewidth}
     \centering
       \includegraphics[width=\linewidth]{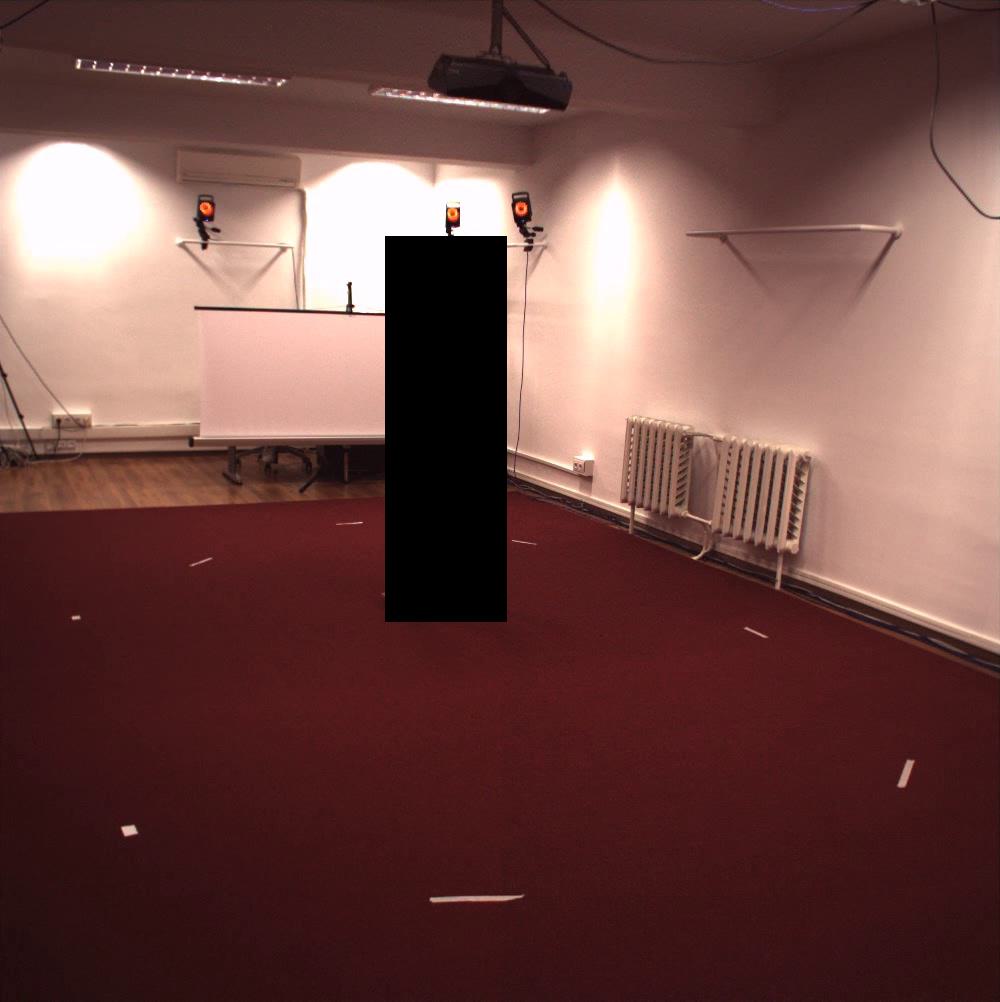}
        \caption*{Frame 10}
    \end{subfigure}\hfill%
    \begin{subfigure}{0.11\linewidth}
    \centering
        \includegraphics[width=\linewidth]{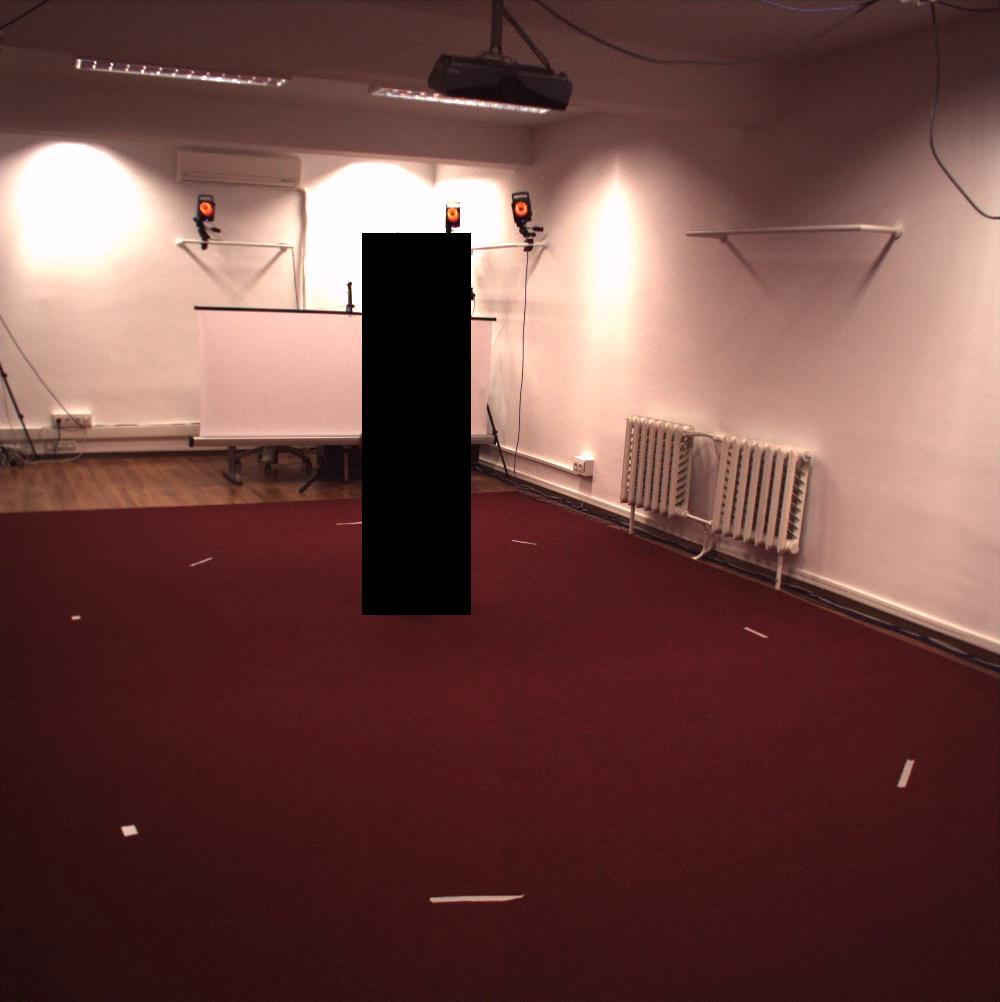}
        \caption*{Frame 16}
    \end{subfigure}\hfill%
    \begin{subfigure}{0.11\linewidth}
        \centering
            \includegraphics[width=\linewidth]{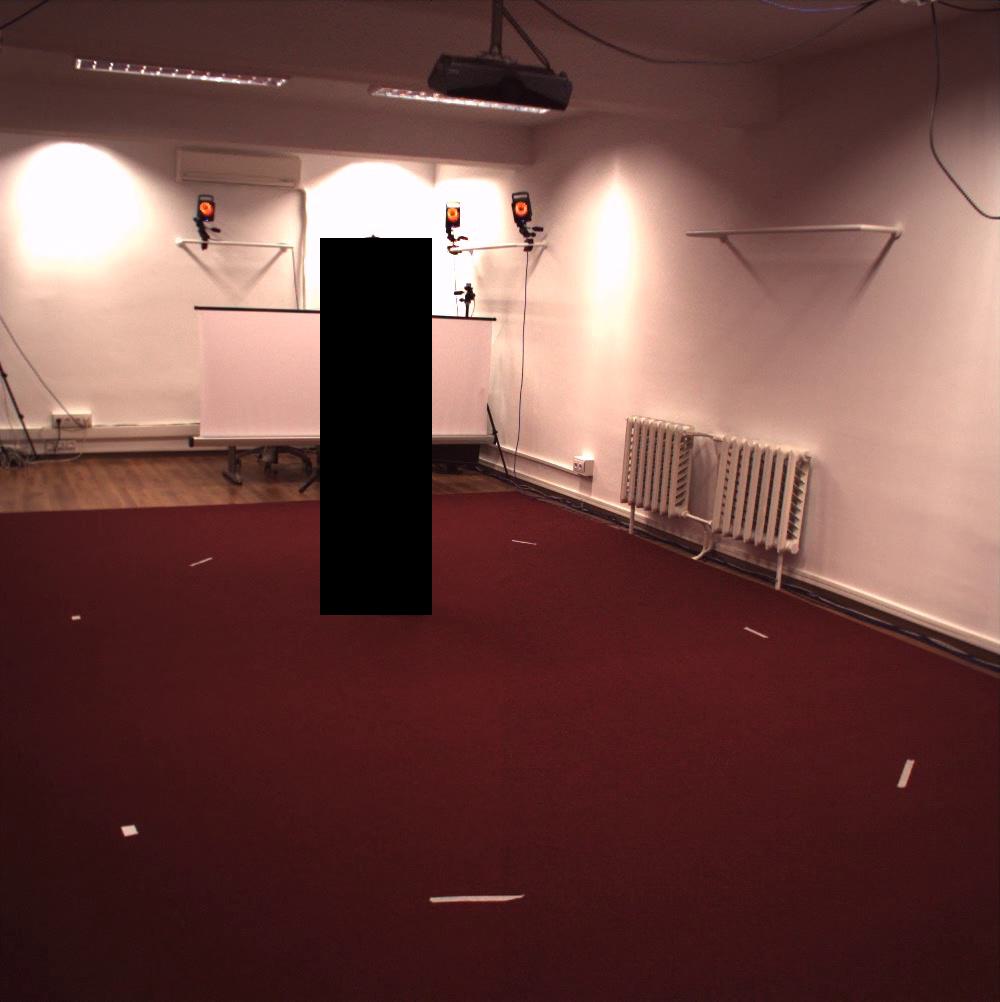}
            \caption*{Frame 19}
    \end{subfigure}\hfill%
    \begin{subfigure}{0.11\linewidth}
\centering
    \includegraphics[width=\linewidth]{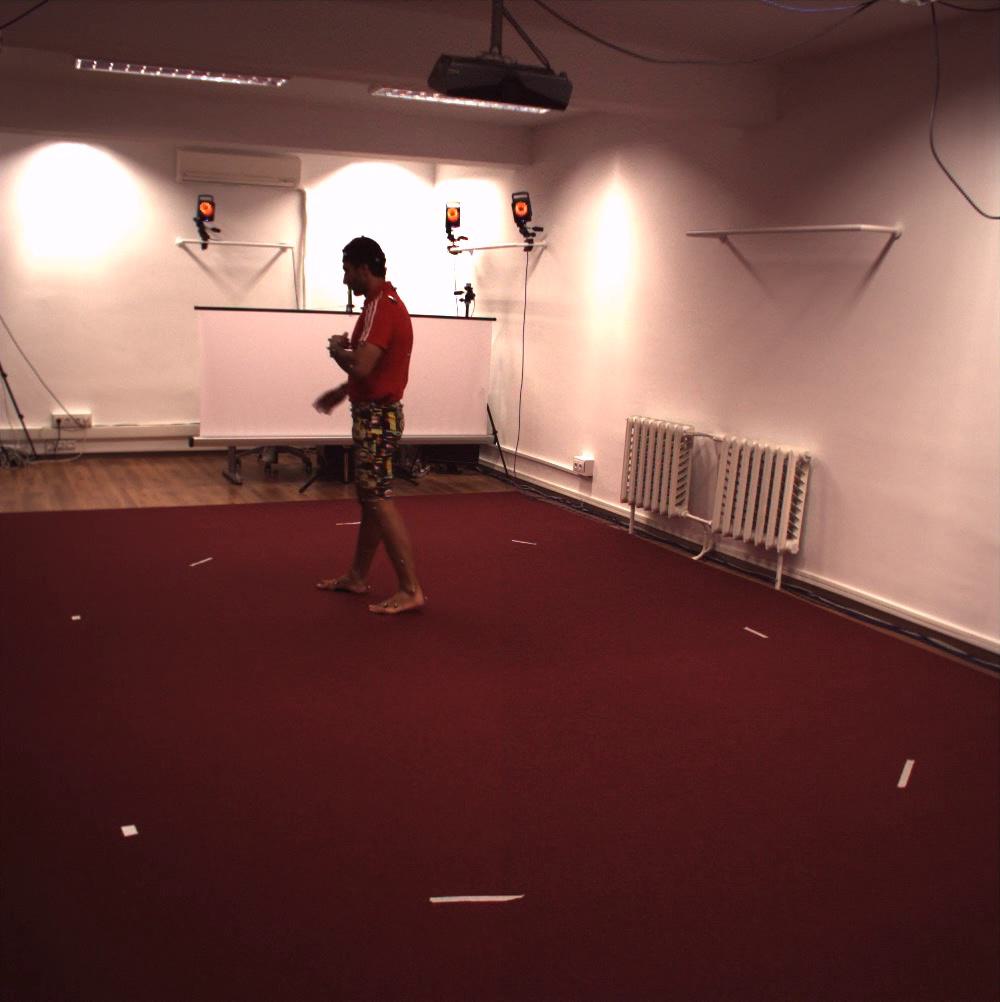}
    \caption*{Frame 20}
\end{subfigure}\hfill%
    \begin{subfigure}{0.11\linewidth}
        \centering
        \includegraphics[width=\linewidth]{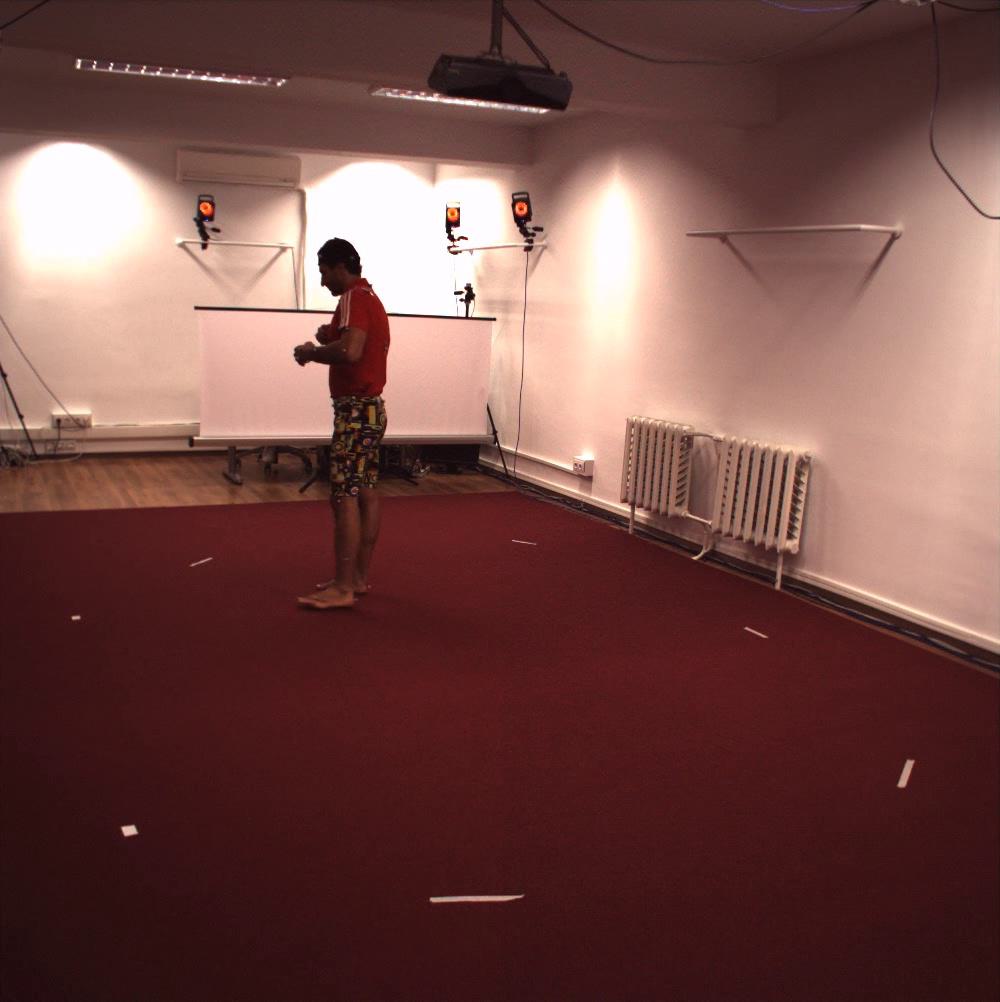}
        \caption*{Frame 26}
    \end{subfigure}\hfill%
    \begin{subfigure}{0.11\linewidth}
    \centering
        \includegraphics[width=\linewidth]{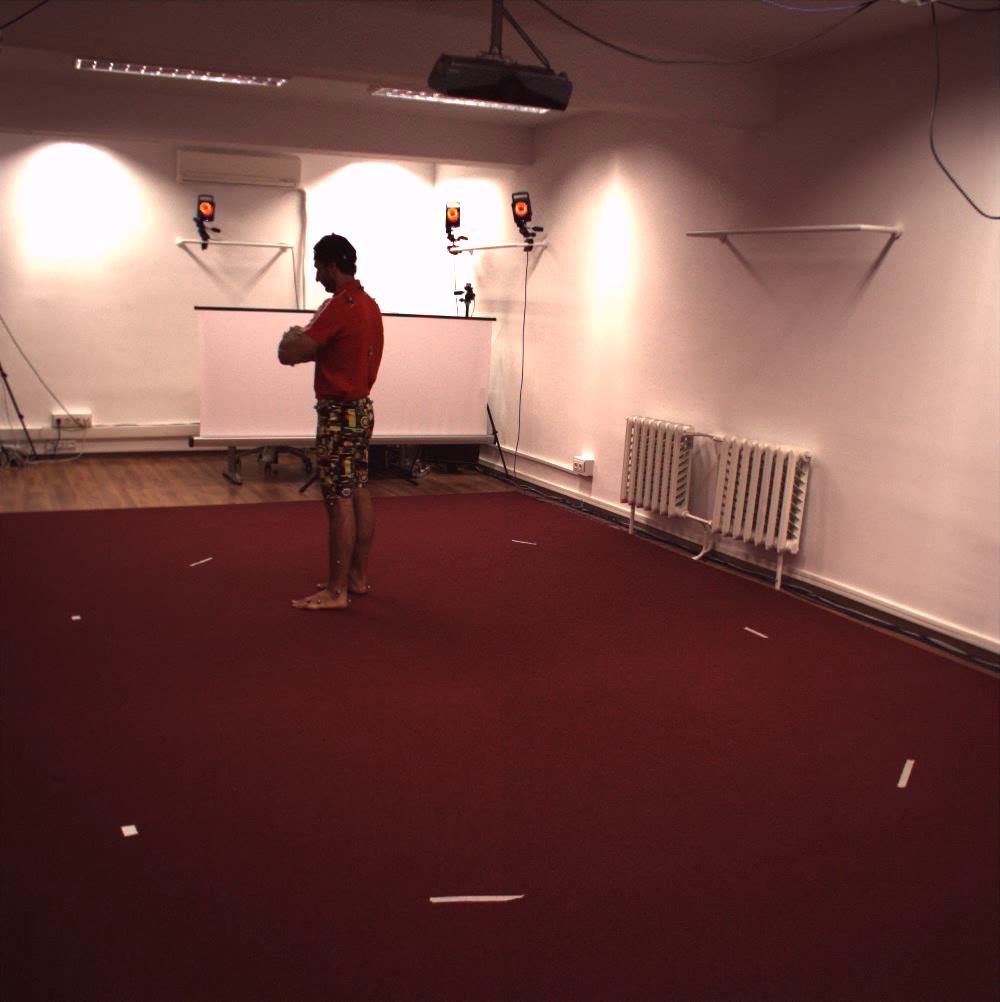}
        \caption*{Frame 29}
    \end{subfigure}
    \caption{\textbf{Samples of Occluded Humans3.6M dataset}. We add artificial occlusions on Human3.6M dataset that persist both spatially and temporally, covering the subject up to 100\%. Here the person between the Frame 10 to Frame 19 remains occluded.}
    \label{fig:dataset_occluded_humans36m}
\end{figure*}

\section{Extending \name~ for Mesh Generation and Recovery}
\name~ is originally proposed to extract the 3D pose estimation of the Human which is $(T,17,3)$ dimensional vector. where $T$ is the number of frames. Extension of \name~ to mesh recovery is trivial as we can simply train the DSTformer backbone to produce a sequence of temporally clean $(T,24,3)$ output from a sequence of noisy input SMPL parameters. Here $(24,3)$ represents the $\theta$ SMPL \cite{10.1145/3596711.3596800}
parameter shape. We can then extract human mesh using the SMPL parameters using an SMPL head. For the re-projection of the human mesh onto the image, we simply use the $\beta$ and camera parameters predicted by BEDLAM-HMR/BEDLAM-CLIFF \cite{black2023bedlam} and do linear interpolation/extrapolation to incorporate missing camera predictions.
\section{Additional Dataset Details}
  
\noindent\textbf{Human3.6M} \cite{h36m_pami}. An indoor-scene dataset, Human3.6M is a pivotal benchmark for 3D human pose estimation from 2D images. Following \cite{black2023bedlam}, we retain every 1 in 5 frames in the test split comprising the S9 and S11 sequence. We perform experiments on the original publically available Human3.6M dataset to show that our method achieves comparable performance with other state-of-the-art methods. \\
\noindent\textbf{OCMotion} \cite{huang2022object}. OCMotion is a video dataset that extends the 3DOH50K image dataset \cite{zhang2020object}, incorporating natural occlusions. The dataset comprises 300K images captured at 10 FPS, featuring 43 sequences observed from 6 viewpoints. Its annotations for 3D motion include SMPL, 2D poses, and camera parameters. The sequences \{0013, 0015, 0017, 0019\} are designated for testing. Our method does not require supervised training, so we have only used the test split when performing all experiments. \\
\noindent\textbf{Occluded Human3.6M}. We curate the Occluded Human3.6M dataset to evaluate our method, specifically designed for assessing human pose estimation under significant occlusion, unlike existing datasets such as Human3.6M, MPI-INF-3DHP\cite{mono-3dhp2017}, and 3DPW\cite{vonMarcard2018}. To accomplish this, we use random erase occlusions on Human3.6M videos, completely covering a person up to 100\%. These occlusions persist spatially and temporally for 1.6 seconds within 3.2 seconds of the video. Some samples are shown in Figure \ref{fig:dataset_occluded_humans36m}.

\noindent\textbf{BRIAR \cite{cornett2023expanding}}. BRIAR is a large-scale biometric dataset featuring videos of human subjects captured in extremely challenging conditions. These videos are recorded at varying distances $i.e$ close range, 100m, 200m, 400m, 500m, and unmanned aerial vehicles (UAV),
with each video lasting around 90 seconds. Most of the pose estimation methods fail on this dataset due to the extreme amount of domain shifts. Additionally, BRIAR lacks ground truth data for poses, which means evaluations of pose estimation methods on this dataset can only be qualitative, relying on visual assessments rather than quantitative metrics.

\section{Additional Related Works}
\noindent \textbf{2D-3D human pose lifting.} Modern 3D human pose estimation encounters significant challenges in generalization due to limited labeled data for real-world applications. \cite{martinez2017simple} addressed this issue by breaking down the problem into 2D pose estimation and 2D to 3D lifting. Subsequently, \cite{chen2019unsupervised} improved on this by including self-supervised geometric regularization, by synthetic data usage \cite{zhu2022decanus}, spatio-temporal transformers \cite{zheng20213d}, and frequency domain analysis \cite{zhao2023poseformerv2}. \cite{zhu2023motionbert} achieved state-of-the-art results by modelling motion priors from a sequence of 2D poses. Although these works perform well up to a certain degree, they suffer from two problems: depth ambiguity of 2D human poses, inaccurate 3D human poses if the initial 2D human poses are noisy. In contrast, we focus on 3D pose estimation in a video-based setting and does not involve any 2D-3D pose lifting. 

{\small
\bibliographystyle{ieee_fullname}
\bibliography{egbib}
}